\documentclass[5p,twocolumn,number,a4paper,11pt]{elsarticle}
\makeatletter
\let\proof\@undefined
\let\@beginproof\@undefined
\let\@endproof\@undefined
\makeatother
\usepackage{url}
\usepackage{cvpr-abbriv}
\usepackage[numbers]{natbib}
\usepackage{epsfig}
\usepackage{array}
\usepackage{multirow}
\usepackage{arydshln}
\usepackage{graphicx}
\usepackage{amsmath}
\usepackage{placeins}
\usepackage{verbatim}
\usepackage{comment}
\usepackage{tocloft}
\usepackage{amssymb}
\usepackage{mathabx}
\usepackage{tocloft}
\usepackage{amssymb}
\usepackage{appendix}
\usepackage{mathabx}
\usepackage{tikz}
\usepackage{color}
\usepackage{amsthm}
\usepackage{aliascnt}
\usepackage{mathrsfs}
\usepackage[absolute]{textpos}
\usepackage{textcomp}
\usepackage{bbm}
\usepackage[draft=false, pagebackref=false,breaklinks=true,colorlinks,bookmarks=false]{hyperref}
\hypersetup{hidelinks,colorlinks=false,linkbordercolor={1 1 1}}
\usepackage[ruled,vlined,linesnumbered]{algorithm2e}
\DeclareMathAlphabet{\mathcalligra}{T1}{calligra}{m}{n}
\DeclareMathAlphabet{\mathantt}{OT1}{antt}{li}{it}
\DeclareMathAlphabet{\mathpzc}{OT1}{pzc}{m}{it}

\newcommand{\argmin}{\mathop{\rm argmin}}

\DeclareMathOperator{\dom}{dom}

\DeclareMathOperator{\conv}{conv}
\DeclareMathOperator{\coni}{coni}
\DeclareMathOperator{\Null}{null}

\let\emptyset\varnothing

\newcommand{\<}{\langle}
\renewcommand{\>}{\rangle}
\newcommand{\Mid}{\:\big|\,}
\renewcommand{\mid}{\:|\,}

\newcommand{\Real}{\mathbb{R}}

\newcommand{\WIx}[1]{\ensuremath{\mathbb{W}_{#1}}\xspace}
\newcommand{\SIx}[1]{\ensuremath{\mathbb{S}_{#1}}\xspace}
\newcommand{\WI}{\WIx{f}}
\newcommand{\SI}{\SIx{f}}

\def\bbbone{{\mathchoice {\rm 1\mskip-4mu l} {\rm 1\mskip-4mu l}{\rm 1\mskip-4.5mu l} {\rm 1\mskip-5mu l}}}

\newcommand{\tab}{{\hphantom{bla}}}
\newcommand{\A}{\mathcal{A}}
\newcommand{\Bool}{\mathbb{B}}

\newcommand{\B}{\mathcal{B}}

\def\O{\mathcal{O}}

\def\G{\mathcal{G}}
\def\H{\mathcal{H}}

\newcommand{\V}{\mathcal{V}}
\newcommand{\I}{\mathcal{I}}

\renewcommand{\P}{\mathcal{P}}
\newcommand{\E}{\mathcal{E}}

\newcommand{\N}{\mathcal{N}}
\newcommand{\X}{\mathcal{X}}

\newcommand{\M}{\mathcal{M}}
\newcommand{\LL}{\mathcal{X}}
\newcommand{\Labels}{\X}

\def\T{^{\mathsf T}}

\newcounter{myRomanCounter}

\providecommand{\Zeta}{\mathcal{Z}}

\def\etalb{ et al.}
\newcommand{\f}{{E_f}}
\newcommand{\g}{{E_g}}
\newcommand{\h}{{E_h}}
\newcommand{\x}{{x}}
\newcommand{\y}{{y}}

\def\c{\textsc{c}}
\def\d{\textsc{d}}
\def\h{\textsc{h}}
\def\a{\textsc{a}}
\def\b{\textsc{b}}

\newcommand{\Esubset}{\stackrel{.}{\subsetneq}}

\newcommand{\Esupset}{\stackrel{.}{\supsetneq}}

\newcommand{\energy}[1]{{E_{#1}}}

\newcommand{\K}{\mathcal{K}}

\newcommand{\AS}{Alexander Shekhovtsov}

\newlength{\figwidth}
\newlength{\figwidtha}

\renewcommand*{\paragraph}[1]{\par{\normalsize\bf #1}\ }

\def\mathrlap{\mathpalette\mathrlapinternal} 
\def\mathllap{\mathpalette\mathllapinternal}
\def\mathllapinternal#1#2{\llap{$\mathsurround=0pt#1{#2}$}}
\def\mathrlapinternal#1#2{\rlap{$\mathsurround=0pt#1{#2}$}}

\DeclareRobustCommand{\ind}[1]{\vphantom{f}_{#1}}
\DeclareRobustCommand{\ind}[1]{(#1)}

\def\leftbb{\mathrlap{[}\hskip1.3pt[}
\def\rightbb{]\hskip1.36pt\mathllap{]}}

\def\Wedge{\bigwedge}
\def\phi{\delta}
\def\epsilon{\varepsilon}

\newcommand{\IF}{\mbox{ \rm if }}

\newcommand{\AND}{\mbox{ \rm and }}
\newcommand{\OTHERWISE}{\mbox{ \rm otherwise}}
\newcommand{\Algorithm}[1]{{Algorithm\,\ref{#1}}}

\newcommand{\Section}[1]{\S\ref{#1}}
\newcommand{\Figure}[1]{{Figure\,\ref{#1}}}
\newcommand{\Theorem}[1]{{Theorem\,\ref{#1}}}
\newcommand{\Lemma}[1]{{Lemma\,\ref{#1}}}

\newcommand{\Statement}[1]{{Statement\,\ref{#1}}}

\newcommand{\lsub}[1]{\limits_{\begin{subarray}{l}\setlength{\arraycolsep}{0.2em}#1\end{subarray}}}
\newcommand{\csub}[1]{\limits_{\begin{subarray}{c}\setlength{\arraycolsep}{0.2em}#1\end{subarray}}}
\newcommand{\clim}[1]{\limits_{\begin{subarray}{c}\setlength{\arraycolsep}{0.2em}#1\end{subarray}}}


\def\maxwi{\eqref{best L-improving}\xspace}
\def\maxsi{{(\mbox{\sc max-si})}\xspace}

\def\anchor [#1]#2{%
\phantomsection{}(#1)\xspace\label{#2}%
\def\arga{#2}%
\global\expandafter\def\csname#2\endcsname{%
\hyperref[#2]{(#1)}\xspace%
}%
}

\def\codefunction [#1]#2{%
\phantomsection{}\label{#2}{\ttfamily #1\xspace}%
\def\arga{#2}%
\global\expandafter\def\csname#2\endcsname{%
\hyperref[#2]{\ttfamily #1}\xspace%
}%
}

\newlength{\myskip}
\renewenvironment{enumerate}%
  {\begin{list}{(\alph{enumi})}%
     {\topsep=0in\itemsep=0in\parsep=0pt\partopsep=0in\usecounter{enumi}}%
   }{\end{list}}

\renewenvironment{itemize}%
  {\begin{list}{$\bullet$}%
     {\topsep=0in\itemsep=0pt\parsep=0pt\partopsep=0in\usecounter{itemi}}%
   }{\end{list}\addvspace{0pt}}

\SetAlgoSkip{skipalgomyskip}
\SetAlgoInsideSkip{}

\makeatletter
\let\corollary\@undefined
\let\c@corollary\@undefined
\let\endcorollary\@undefined
\let\definition\@undefined
\let\c@definition\@undefined
\let\enddefinition\@undefined
\let\theorem\@undefined
\let\c@theorem\@undefined
\let\endtheorem\@undefined
\let\lemma\@undefined
\let\c@lemma\@undefined
\let\endlemma\@undefined
\makeatother

\newtheoremstyle{tightItalic}
  {0.5\myskip}
  {0.5\myskip}
  {}
  {}
  {\itshape}
  {.}
  { }
  {}

\newtheoremstyle{tightBf}
  {0.5\myskip}
  {0.5\myskip}
  {}
  {}
  {\bf}
  {.}
  {.5em}
  {}

\theoremstyle{tightBf}
\newtheorem{theorem}{Theorem}[section]

\newtheorem*{theorem*}{Theorem}
\newaliascnt{corollary}{theorem}%
\newtheorem{corollary}[corollary]{Corollary}
\aliascntresetthe{corollary}

\newaliascnt{definition}{theorem}%
\newtheorem{definition}[definition]{Definition}
\aliascntresetthe{definition}

\newaliascnt{statement}{theorem}%
\newtheorem{statement}[statement]{Statement}
\aliascntresetthe{statement}

\newaliascnt{lemma}{theorem}%
\newtheorem{lemma}[lemma]{Lemma}
\aliascntresetthe{lemma}

\newaliascnt{example}{theorem}%

\aliascntresetthe{example}

\newaliascnt{remark}{theorem}%
\newtheorem*{remark*}{Remark}
\aliascntresetthe{remark}

\newaliascnt{proposition}{theorem}%

\aliascntresetthe{proposition}

\newaliascnt{property}{theorem}%

\aliascntresetthe{property}

\theoremstyle{tightItalic}


\usepackage{thmtools, thm-restate}
\usepackage{cleveref}
\makeatletter
\g@addto@macro\normalsize{%
  \setlength\abovedisplayskip{1ex}
  \setlength\belowdisplayskip{1ex}
  \setlength\abovedisplayshortskip{1ex}
  \setlength\belowdisplayshortskip{1ex}
}
\makeatother

\def\checkmark{\tikz\fill[scale=0.4](0,.35) -- (.25,0) -- (1,.7) -- (.25,.15) -- cycle;}

\let\cite=\citep
\newif\ifblacknwhite
\blacknwhitetrue
\def\AS{Alexander Shekhovtsov}

\newcommand{\mytitle}{Higher Order Maximum Persistency and Comparison Theorems}
\def\mythanks{This work was supported by the Austrian Science Fund (FWF) under the START project BIVISION, No. Y729.}
\begin{document}

\begin{textblock}{14}(1,0.5)
\noindent
\textcopyright\ 2015, Elsevier. Licensed under the Creative Commons Attribution-NonCommercial-NoDerivatives 4.0 International \url{http://creativecommons.org/licenses/by-nc-nd/4.0/}.
\end{textblock}

\def\qqed{}
\begin{frontmatter}

\title{\mytitle}

\author{\AS\fnref{myfootnote}}
\address{Graz University of Technology \\[5pt] {\tt\small shekhovtsov@icg.tugraz.at}}
\fntext[myfootnote]{\mythanks}


\begin{abstract}
We address combinatorial problems that can be formulated as minimization of a partially separable function of discrete variables (energy minimization in graphical models, weighted constraint satisfaction, pseudo-Boolean optimization, 0-1 polynomial programming).
For polyhedral relaxations of such problems it is generally not true that variables integer in the relaxed solution will retain the same values in the optimal discrete solution. Those which do are called persistent. Such persistent variables define a part of a globally optimal solution. Once identified, they can be excluded from the problem, reducing its size.

To any polyhedral relaxation we associate a sufficient condition proving persistency of a subset of variables. We set up a specially constructed linear program which determines the set of persistent variables maximal with respect to the relaxation. The condition improves as the relaxation is tightened and possesses all its invariances.
The proposed framework explains a variety of existing methods originating from different areas of research and based on different principles. A theoretical comparison is established that relates these methods to the standard linear relaxation and proves that the proposed technique identifies same or larger set of persistent variables.

\end{abstract}


\end{frontmatter}

%
\setlength\cftparskip{0pt}
\setlength\cftbeforesecskip{2pt}
\setlength\cftbeforesubsecskip{1pt}
\tableofcontents
\setlength{\figwidth}{\linewidth}
\setlength{\figwidtha}{\linewidth}
\section{Introduction}
Optimization models in the general form of minimizing a partially separable function of discrete variables, known as energy minimization, weighted/valued constraint satisfaction or max-sum labeling, proved useful in many areas. 
The function has the form $\f(x) = \sum_{\c\in\E}f_\c(x_c)$. In computer vision and machine learning such models are largely motivated by maximum a posteriori inference in graphical models~\cite{Wainwright-08} used to model a variety of structured statistical recognition problems.
%
In case variables take only two values $0$ or $1$, the problem is known as {\em pseudo-Boolean optimization} or $0$-$1$ polynomial programming. 
Problems where terms (summands) involve at most two $0$-$1$ variables at a time are called {\em quadratic}.
We consider the general case, where terms may couple more that two variables at a time (higher order) and variables can take more then two values (multilabel).
\par
One major trend for performing inference in graphical models is represented by graph cut methods. 
The basic capability is essentially to solve a binary pairwise submodular problem, \eg, image segmentation~\cite{Greig89}, by reduction to a minimum cut / maximum flow problem. For the latter, many efficient algorithms exist and their running time is experimentally  near linear for typical vision problems~\cite{Boykov04}.
This basic method was extended to submodular multilabel problems~\cite{Ishikawa03,DSchlesinger-K2}, to general multilabel problems by solving for an optimized crossover between two candidate solutions at a time~\cite{Boykov99}, to higher-order $0$-$1$ models reducible to a graph cut~\cite{Kolmogorov02:regular-pami,FreedmanD05} and to combinations of higher order and multilabel~\cite{Ladicky2010,DelongOIB12}.
\par
Another technique that can be considered nowadays as a basic graph cut method is the roof dual relaxation~\cite{BorosHammer02} known in computer vision as quadratic pseudo-Boolean optimization (QPBO)~\cite{Kolmogorov-Rother-07-QBPO-pami}. It allows to find a partial optimal solution to a non-submodular binary problem and reduces to finding a minimum cut in a specially constructed network~\cite{Boros:TR91-maxflow}. It can be interpreted~\cite{Kolmogorov12-bisub} as solving a submodular relaxation of the initial problem.
This basic method is again extended to multilabel problems by solving crossover problems~\cite{Lempitsky08fusionflow} and to general higher order $0$-$1$ problems by reduction (quadratization) techniques expressing the function as a quadratic function with auxiliary variables~\cite{Ishikawa-11,Fix-11,Boros-12}.
\par
Another direction of extending graph cuts to higher order models relies on minimization of more general submodular functions. 
Several efficient max-flow based algorithms have been proposed~\cite{Arora-12,Kolmogorov-10} for minimization of a sum of submodular functions (SoS). A natural extension of QPBO is represented by submodular and bisubmodular relaxations~\cite{Kolmogorov12-bisub,KahlS-12}.
\par
Arguably, linear programming (LP) is a much more costly tool than computing a minimum cut. Yet, it provides theoretical insight to many methods~\cite{Komodakis-07,Komodakis-2008} and there has been solvers developed that can address (sometimes approximately) large scale problems. Dual decomposition methods~\cite{Schles-subgradient,Komodakis-11} or dual block-descent methods, in particular TRW-S~\cite{Kolmogorov-06-convergent-pami}, are competitive with graph cut based methods in terms of speed and quality. 
There are extensions of these specialized LP methods to higher order models~\cite{Kolmogorov-12-GTRWS,Kolmogorov-13-SRMP}. 
Smoothing~\cite{Savchynskyy-11} and proximal~\cite{RavikumarAW10} methods are scalable and offer a theoretically guaranteed convergence speed.
Cutting plane approaches~\cite{Sontag-12,Werner-08-higher-arity} are used to tighten the relaxation adaptively to the problem.
\par 
One drawback of relaxation based methods is that the final discrete solution is obtained by so-called rounding schemes and often appears inferior to solutions by graph cut methods as they stay feasible to the discrete space.
Even in the case when many of the relaxed variables take integer values in the optimal relaxed solution, a fundamental problem remains that they may not take the same integer values in the optimal discrete solution. Therefore, unless the relaxation is tight, a local rounding technique cannot provide any guarantees for general models. The situation is dramatically different when we consider quadratic pseudo-Boolean functions. There, all variables that are integer in the relaxation correspond to at least one globally optimal discrete solution~\cite{Nemhauser-75,Hammer-84-roof-duality}. This property of the relaxation is called {\em persistency}. 
For general $0$-$1$ polynomial problems persistency was studied by~\cite{Lu-Williams-1987,Adams:1998}. In their terminology persistency is associated with relaxations and is a property of the relaxed solution as a whole.
In this work we call any partial assignment of a subset of discrete variables {\em persistent} if it can be provably extended to a globally optimal solution based on the properties of the relaxation or any other sufficient condition. 
%
%
Success of relaxation based exact methods such as~\cite{SavchynskyyNIPS2013} on computer vision and machine learning problems suggests that often a large part of the relaxed solution is integral. In this case we are interested in determining the largest subset of such variables that is persistent. 
%
\paragraph{Related Work}
The present work was to a large extent inspired by simple local sufficient conditions proposed in bioinformatics under the name {\em dead end elimination} (DEE)~\cite{Desmet-92-dee,Goldstein-94-dee}. 
In computer vision, several methods were proposed that identify a persistent assignment directly in the multi-label setting. These are methods by~\citet{Kovtun03,Kovtun-10} and \citet{Swoboda-13,Swoboda-14}. 
Method~\cite{Swoboda-14} is applicable with a general polyhedral relaxation and it maximizes the subset of peristent variables for their sufficient condition (discussed in \Section{sec:swoboda}).
\par
In the case of $0$-$1$ variables there are several different techniques. \citet{Adams:1998} proposed a sufficient condition on dual multipliers to prove persistency of the integral part of the relaxation. Quadratization techniques by~\citet{Ishikawa-11,Fix-11,Boros-12} introduce auxiliary variables in order to reduce the function to a quadratic form and infer persistency from the QPBO method. \citet{Lu-Williams-1987} generalized the roof duality approach to higher order by using a higher order linear relaxation. 
\citet{Kolmogorov12-bisub} generalized both QPBO and the construction by \citet{Lu-Williams-1987} by proposing discrete submodular and bisubmodular relaxations. He argues that the key property of QPBO that needs to be generalized is the existence of a totally half integral optimal solution to the relaxation, \ie, with values in~$\{0,\frac{1}{2},1\}$. He characterized all totally half-integral relaxations as bisubmodular relaxations. Finding a good (bi)submodular relaxation appears to be a challenging problem. 
To our knowledge it was only resolved for the special case of mincut-reducible relaxations~\cite{KahlS-12,Strandmark2012PhD}, and even in this case it requires solving a series of linear programs. Even though the relaxed problem itself can be efficiently optimized (in particular when it is a sum of submodular functions), having a sound persistency result at a comparable computation cost is an open problem. No theoretical comparison seems to be possible between (bi)submodular relaxations and quadratization techniques~\cite{Kolmogorov12-bisub}.
\par
\citet{kohli:icml08} reduce multilabel pairwise problems to $0$-$1$ quadratic and \citet{Windheuser-et-al-eccv12} reduce multilabel higher order problems to submodular relaxations of~\cite{Kolmogorov12-bisub}.
\par
%
\paragraph{Contribution}
In this work we settle the persistency capabilities achievable with a general polyhedral relaxation. The previously known results are in a certain sense unique, relying on a specific sufficient condition or on a specific type of the relaxation. 
We show that persistency guarantees are not that rare. To any polyhedral relaxation we associate clear sufficient conditions for persistency. We propose a polynomial time method to determine the largest strongly persistent subset of variables according to the sufficient condition. 
The method sets up a linear program connected to the given relaxation polytope and maximizes the number of strongly persistent variables. In comparison to QPBO-based or submodularity-based techniques, we employ a more costly optimization tool, but gain the following advantages:
\begin{itemize}
\item the new sufficient condition generalizes a wide variety of existing methods that span across different fields of research and apply different techniques;
\item it is possible to pose formally and solve (under certain restrictions) the problem of determining the largest subset of persistent variables;
\item the maximum \wrt to the proposed general sufficient condition is guaranteed to be at least as good as any of the individual methods or their combinations;
\item the method is invariant to the permutation of labels and reparametrization of the problem as long as the relaxation is invariant;
\item persistent assignments form a hierarchy when tightening the relaxation. 
\end{itemize} 
%
The author's previous work~\cite{shekhovtsov-phd, shekhovtsov-14-TR} considered only pairwise models and the standard LP relaxation. This paper generalizes to higher order and arbitrary polyhedral relaxations, gives more complete proofs of some properties and establishes comparisons with a novel multilabel method~\cite{Swoboda-14} and higher-order $0$-$1$ methods~\cite{Kolmogorov12-bisub,Ishikawa-11,Fix-11,Adams:1998}.
\paragraph{Outline}
In \Section{sec:max-persistency} we propose a general approach to persistency with a general polyhedral relaxation. This includes the proposed linear program formulation of maximum persistency and general properties of the problem. 
In \Section{sec:Simplex-LP} we consider standard LP relaxations and specialize the construction for this case, many properties are simplified. In \Section{sec:comparison} we propose a theoretical comparison between the proposed framework and other approaches. In \Section{sec:experiments} we validate our theoretical findings experimentally and compare performance on small random problems. In \Section{sec:conclusion} there is a conclusion and discussion and \Section{sec:A} contains proofs. 

\subsection{Notation} When generalizing to higher order models many statements and proofs simplify if we use a properly defined notion of the empty sum, the empty product and the empty Cartesian product. They are respectively: $\sum_{i\in\emptyset} x_i = 0$, $\prod_{i\in\emptyset} x_i = 1$ and $\prod_{i\in \emptyset} X_i = \{\emptyset \}$. The inclusion $\subset$ is non-strict and $\subsetneq$ is strict. LHS refers to the left hand side of an equation. $\leftbb\cdot\rightbb$ denotes the Iverson bracket. $\argmin$ is the set of minimizers. Sets $\Real$, $\Real_+$ denote real and non-negative real numbers and $\Bool = \{0,1\}$ is the Boolean domain. A composition of functions is denoted as $(f\circ g)(x) = f(g(x))$. Finally, a {\em polytope} is assumed to be convex but may be unbounded and a {\em polyhedron} means the same as a polytope. 
\subsection{Energy Minimization}\label{sec:energy}
\begin{figure}[t]
\centering
\includegraphics[width=0.8\linewidth]{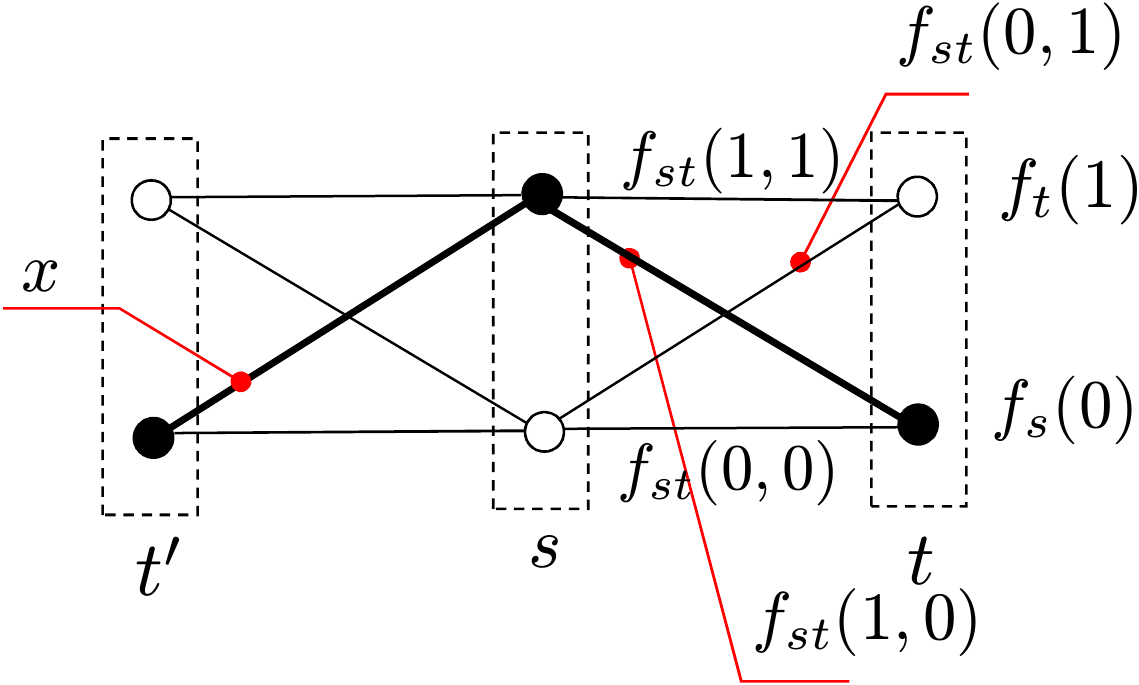}
\caption{Graphical notation by \citet{Schlesinger-76}. Variables $x_s, x_t, x_{t'}$ are depicted as boxes, their possible states as circles and states of pairs of variables as lines. In the first order model (pairwise) the energy of a labeling $x$ is the sum of the selected unary and pairwise costs.
}\label{fig:energy}
\end{figure}

%
A hypergraph $(\V,\E)$ is given by the set of nodes $\V$ and the set of hyperedges $\E \subset 2^\V$. We assume that $\V$ is totally ordered and each hyperedge $\c\in\E$ is identified with the tuple of elements of $\c$ ordered \wrt the total order of $\V$. We will further assume that $\emptyset\in\E$ and $(\forall s\in\V)\ \{s\}\in\E$. 
Let $\X_s$ be a finite set of {\em labels} associated to a node $s\in\V$. For a subset of nodes $\c \subset \V$ the set 
$\X_\c$ denotes the Cartesian product $\prod_{s\in \c} \X_s$ in the order defined on $\V$ and $\X = \X_\V$. 
The assignment of labels to all nodes $x\colon \V \to \X$ is called a {\em labeling}. Let $x_\c$ denote the restriction of $x$ to $\c\subset\V$ (thus $x_s$ is just a single coordinate) and $x_\emptyset = \emptyset$.
Let us define the following functions (terms):
\begin{align}
\tag{general hyperedge term}
& (\forall \c\in\E) \ \ f_{\c} \colon \X_\c \to \Real. 
\end{align}
The special cases read
\begin{subequations}
\begin{align*}
f_{\emptyset} & \colon \{\emptyset \} \to \Real & \mbox{(constant term)},\\
f_{s} &\colon \X_s \to \Real 
& \mbox{(unary / 0 order term)}, \\
f_{\{s,t\}} & \colon \X_{\{s,\,t\}} \to \Real
& \mbox{(pairwise / 1st order term)}
\end{align*}
\end{subequations}
and so on. The constant term $f_{\emptyset}$ is nothing but a single number. 
The {\em energy function} $E_{f} \colon \X \to \Real$ is defined by
\begin{equation}\label{the energy}
E_{f}(x) = \sum_{\c\in\E}f_{\c}(x_\c).
\end{equation}
It is a partially separable function of discrete variables $x$. In this paper we will use a graphical notation of the energy explained in \Figure{fig:energy}.
\par
The general energy minimization problem is NP-hard to approximate\footnote{\eg, inapproximability of the traveling salesman problem~\cite{Orponen90onapproximation}.}. On the other hand, there are tractable subclasses. Works by \citet{Thapper-12,Thapper-13} and \citet{Kolmogorov-12-LP-power} characterized all {\em languages} of energy functions with terms from a fixed finite set and unrestricted structure. They showed that there are no tractable languages other than those that can be solved by the basic LP relaxation (defined in~\Section{sec:Simplex-LP}), which proves that the relaxation is a universal and powerful technique. 
%
\par
%
%
\subsection{General Polyhedral Relaxation}\label{sec:general-polyhedral}
\begin{figure}[t]
\centering
\includegraphics[width=\linewidth]{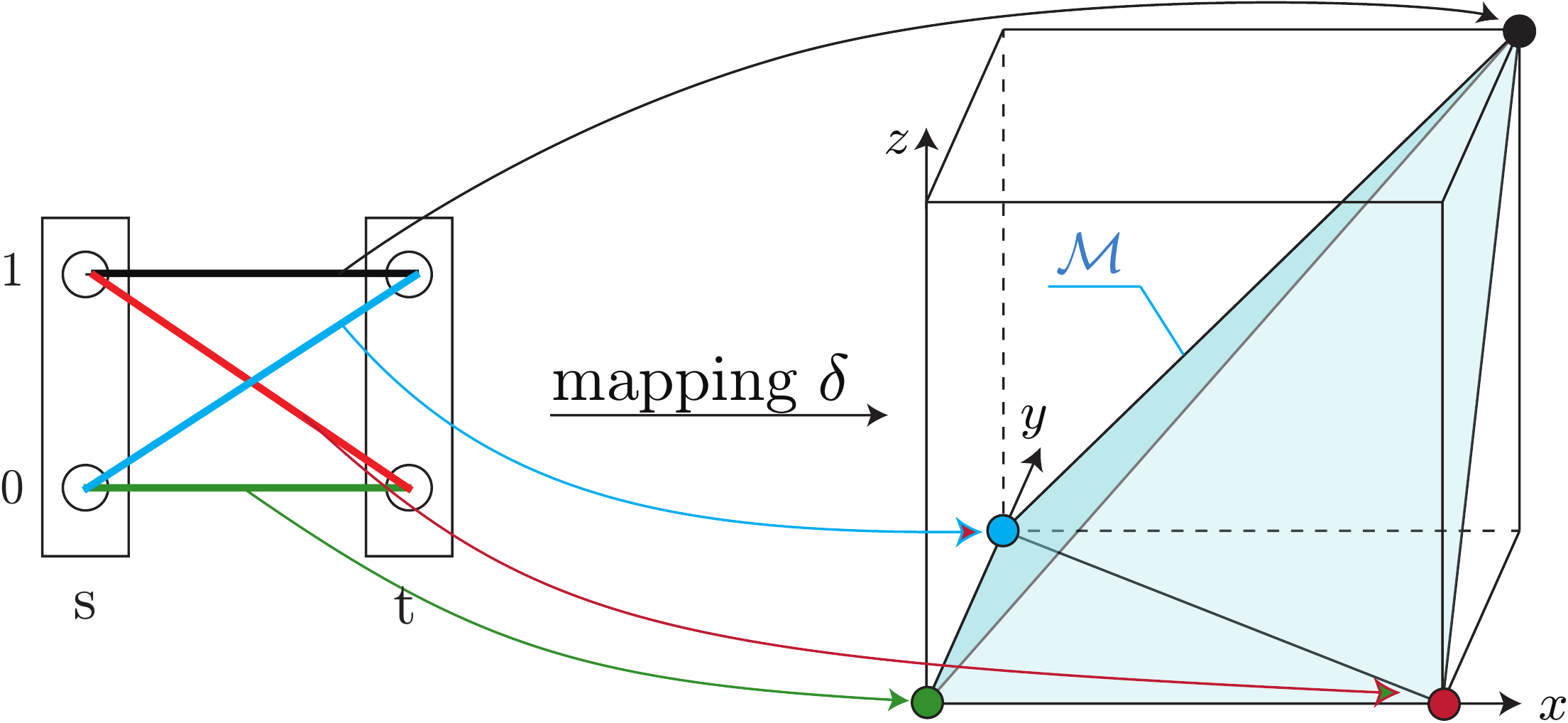}
\caption{Mapping $\delta$ embeds discrete labelings as points in the space $\Real^\I$. Left: 2 variables with 2 states, lines of different colors show possible assignments. Right: to each labeling $x$ there correspond a point $\delta(x) \in\Real^\I$. Axis $x,y,z$ in the figure correspond respectively to $\delta_{s}(1)$, $\delta_{t}(1)$ and $\delta_{st}(1,1)$. In this representation the energy function is a linear functional. The minimization domain can be extended equivalently from the set of points $\delta(\X)$ to their convex hull, the marginal polytope $\M$.
} 
\label{f:embed1}
\end{figure}

In this section we embed the energy minimization problem into the Euclidean space. A labeling $x$ is represented as a $0$-$1$ vector in order to linearize the energy and write it as scalar product of this vector with the cost vector $f$ consisting of all components $f_{\c}(x'_\c)$ for $\c\in\E$, $x'_\c\in\X_\c$.
According to these components 
let us define the following set of {\em indices} $\I = \{(\c, x'_{\c}) \mid \c\in\E,\, x'_\c \in \X_\c\}$. The {\em embedding} $\delta \colon \X \to \Real^\I \colon \x \to \delta(x)$ is defined by its components
\begin{equation}
(\forall \c\in\E,\, x'_\c\in\X_\c)\ \ \delta(x)_\c(x'_{\c})  = \leftbb x_\c{=}x'_\c \rightbb.
\end{equation}
The special cases read
\begin{subequations}
\begin{align}
& \delta(x)_\emptyset = 1,\\
& \delta(x)_s(x_s') = \leftbb x_s{=}x'_s\rightbb,\\
& \delta(x)_{\{s,t\}}(x_{\{s,t\}}') = \leftbb x_s{=}x'_s\rightbb \leftbb x_t{=}x'_t \rightbb
\end{align}
\end{subequations}
and so on.
Let $\<\cdot,\cdot\>$ denote the scalar product in $\Real^{\I}$. 
We can write the energy using the embedding $\delta$ as a linear function:
\begin{equation}
\f(x) = \sum_{\c\in\E}\sum_{x'_\c\in\X_\c}f_\c(x'_\c) \delta(x)_\c(x'_\c) = \<f,\phi(x)\>.
\end{equation}
The embedding $\delta$ is illustrated in \Figure{f:embed1}. The energy minimization can be expressed as:
\begin{align}
\label{energy-linear}
\min_{x\in \LL} \<f,\phi(x)\>
 & = \min_{\mu\in \delta(\LL) }\<f,\mu\>  = \min_{\mu\in \M }\<f,\mu\>,
\end{align}
where $\delta(\X)$ is the image of the set of labelings, \ie, the set of corresponding points in $\Real^\I$ and $\M = \conv \delta(\X)$ is their convex hull, called {\em marginal polytope}~\cite{Wainwright-08}. The second equality follows from the fact that a convex combination of solutions is also a solution. Polytope $\M$ has in general exponentially many facets.
A {\em relaxation} of the problem is obtained by replacing $\M$ with an outer approximation $\Lambda\supset \M$:
\begin{align}\label{LP-Lambda}
\min_{\mu\in\Lambda}\<f,\mu\>.
\end{align}
A vector $\mu\in \Lambda \subset \Real^\I$ will be called a {\em relaxed labeling}.
We will consider polyhedral relaxations of the following general form:
\begin{equation}\label{Lambda-relaxation}
\begin{aligned}
\Lambda = \{ \mu\in\Real^\I \mid A \mu \geq 0;\ \mu_\emptyset = 1;\ \mu \geq 0 \},
\end{aligned}
\end{equation}
where we assume that $A\in \Real^{m \times |\I|}$ is such that $\Lambda$ is bounded and $\M \subset \Lambda$. Since $\M$ is non-empty, it follows that $\Lambda$ is non-empty. By these assumptions, relaxation~\eqref{Lambda-relaxation} is a feasible and bounded linear program.
%
Note that general inhomogenous equality and inequality constraints can be represented in this form by utilizing the component $\mu_\emptyset$. 
The dual problem to~\eqref{LP-Lambda} and the conical hull of $\Lambda$ are expressed conveniently as follows. Recall that for a convex set $\Lambda \subset \Real^\I$ its conical hull is the set:
\begin{equation}\label{coni-Lambda-def}
\coni(\Lambda) = \{ \alpha \mu \mid \mu \in \Lambda, \alpha\geq 0\}.
\end{equation}%
\begin{restatable}{lemma}{LconiLambda}\label{L:coni-Lambda}
The conical hull of a relaxation polytope $\Lambda$ (in the form~\eqref{Lambda-relaxation}, non-empty and bounded) is obtained by dropping the constraint $\mu_\emptyset = 1$:
\begin{equation}\label{coni-Lambda}
\coni(\Lambda) = \{ \mu\in\Real^\I \mid A \mu \geq 0; \mu \geq 0 \}.
\end{equation}
\end{restatable}
%
\par
The linear program~\eqref{LP-Lambda} and its dual are expressed as
\begin{equation}\label{LP}
\begin{array}{rrlr}
& \min \<f,\mu\> & = \hskip-0.3cm & \max \psi \,,\\
&
\setlength{\arraycolsep}{0.2em}
\begin{array}{rl}
A \mu &\geq 0\\
\mu_\emptyset &= 1\\
\mu &\geq 0
\end{array}
& &
\setlength{\arraycolsep}{0.2em}
\begin{array}{rl}
\varphi & \in \Real^m_+\\
\psi &\in \Real \\
f - A\T \varphi - e_\emptyset \psi & \geq 0\\
\end{array}
\end{array}
\tag{LP}
\end{equation}
where vector $e_\emptyset\in\Real^\I$ is the basis vector for the component $\emptyset$ and the equality between the primal and the dual formulations holds because the primal problem is feasible and bounded.
Let us introduce the notation $f^{\varphi} := f - A\T \varphi$. Later on, when we consider equality constraints of the form $A \mu = 0$, the vector $f^\varphi$ will obtain the meaning of an equivalent problem and for now it is just an abbreviation.
\section{Maximum Persistency}\label{sec:max-persistency}
A {\em partial assignment} $y_\A \in \X_\A$, where $\A \subset \V$, is called {\em weakly persistent} if there exists an optimal solution $x$ such that $x_\A = y_\A$. In other words, $y_\A$ can be extended to a global solution.
Partial assignment $y_\A$ is called {\em strongly persistent} if $x_\A = y_\A$ holds for all optimal solutions $x$.  
\par
It may seem that there are no practical reasons to distinguish strongly and weakly persistent partial assignments as long as they allow to simplify the problem. However, it will become clear later that they have different theoretical properties leading to polynomially solvable versus NP-hard maximum persistency problems. It turns out that strong persistency is more tractable, whereas proofs are generally easier to obtain in the weak form and most results in the literature deliver weak persistency.
%
\par
\begin{figure}
\centering
\includegraphics[width=0.9\linewidth]{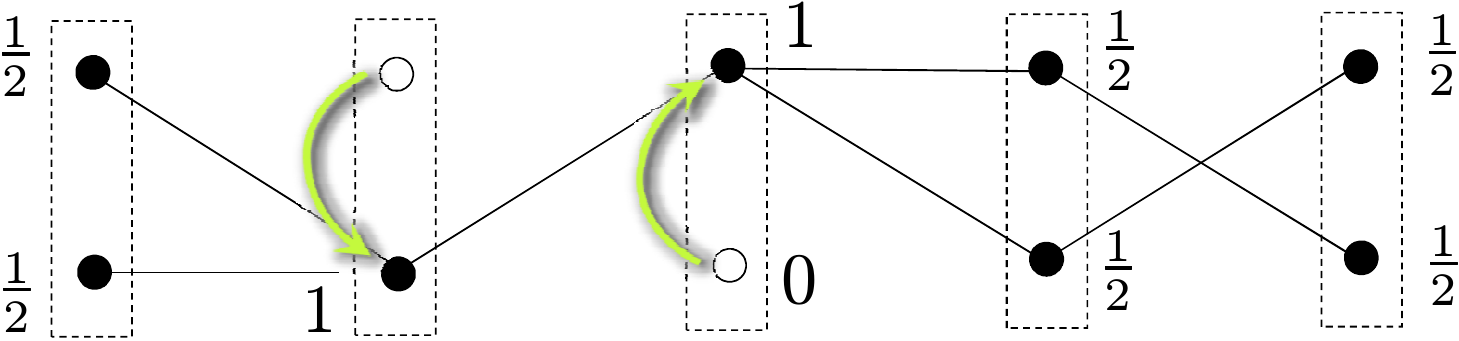}
\caption{Quadratic pseudo-Boolean case (roof dual). There exist half-integral optimal solution to the relaxation (indicated by numbers). Its integer part (the assignment indicated by 1's) is persistent, \ie, optimal to the original discrete optimization problem. The arrows show how an arbitrary labeling can be changed in order to improve the energy: switch to the optimal assignment for integer nodes and keep the assignment of the remaining nodes.
}
\label{fig:qpbo}
\vskip10pt
\includegraphics[width=0.9\linewidth]{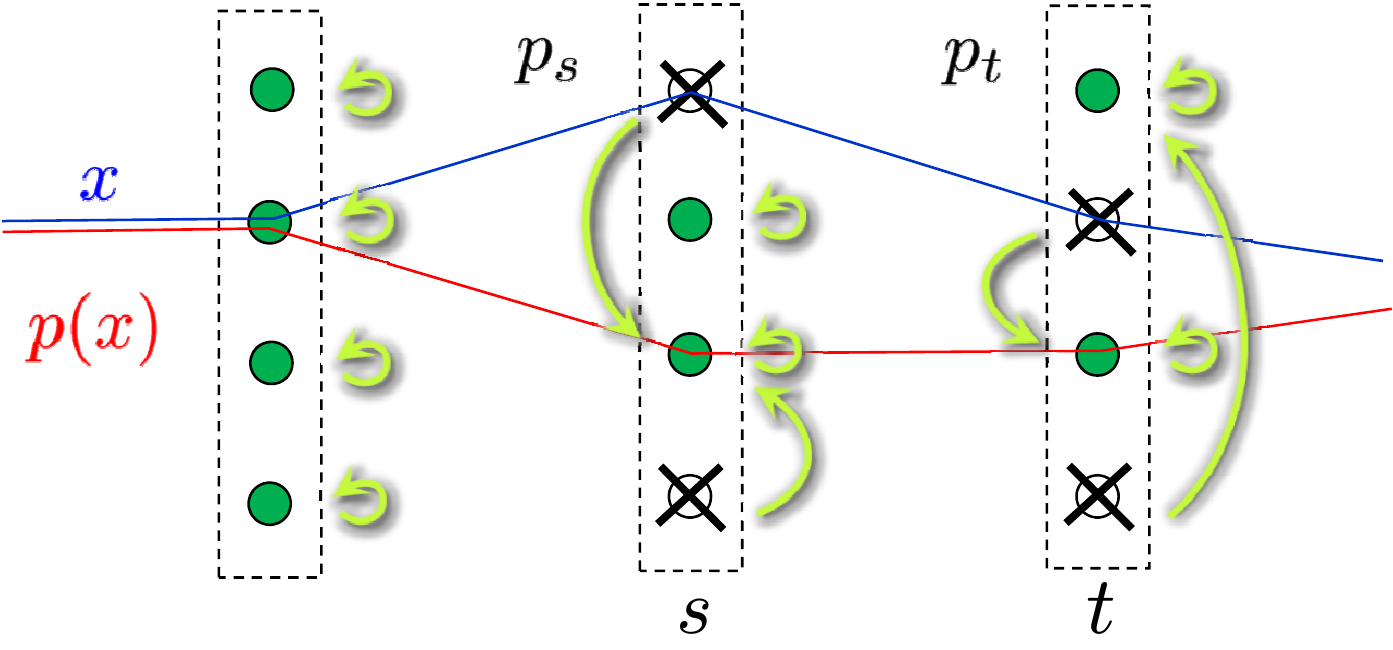}
\caption{
Improving mapping is a generalization of autarky. To every variable $s$ a there is an associated mapping $p_s\colon\X_s \to \X_s$, shown by arrows. For any labeling $x$ its image $p(x)$ has the same or better energy: $\f(p(x)) \leq \f(x)$. 
If label $i$ is not in the range $p_s(\X_s)$ then it can be eliminated as shown by crosses.
}
\label{fig:improving-map}
\end{figure}

In the case of quadratic pseudo-Boolean functions the roof dual relaxation~\cite{BorosHammer02} is persistent: for any relaxed solution its integral part defines a partial assignment $y_\A$ which is optimal to the discrete problem. Moreover, for any labeling $x$, not necessarily optimal, replacing part of $x$ on $\A$ with $y_\A$, the {\em overwrite} operation, denoted in~\cite{BorosHammer02} by $x [\A{\leftarrow}y]$, has the following {\em autarky} property:
\begin{align}\label{eq:autarky}
(\forall x\in\X)\ \ \f(x[\A{\leftarrow}y]) \leq \f(x),
\end{align}
illustrated in \Figure{fig:qpbo}.
We will generalize this property to the multilabel setting.
%
%

\subsection{Improving Mapping}\label{subsec:improving-map}
The overwrite operation discussed above can be represented by a discrete mapping $p\colon \X \to \X \colon x \mapsto x[\A{\leftarrow}y]$. The following generalization of autarky to an arbitrary mapping is proposed.
\begin{definition}\label{def:improving mapping}
A mapping $p \colon \LL \to \LL$ is called {\em (weakly) improving} for $f$ if
\begin{equation}\label{improving mapping}
(\forall x\in\LL) \tab \f(p (x)) \leq \f (x),
\end{equation}
and {\em strictly improving} if
\begin{equation}\label{s-improving mapping}
(p(x)\neq x) \ \Rightarrow\ \f(p (x)) < \f (x).
\end{equation}
\end{definition}\noindent
The idea of the improving mapping is illustrated in \Figure{fig:improving-map}. 
It easily follows from the definition that if $p$ is improving then there exists an optimal solution $x\in p(\X)$ and if $p$ is strictly improving then all optimal solutions are contained in $p(\X)$. In this way an improving mapping reduces the search space from $\X$ to $p(\X)$.
\par
We will consider {\em node-wise} mappings, of the form
$p(x)_s = p_s(x_s)$,
where $(\forall s\in \V)$ $p_s \colon \LL_s \to \LL_s$. Furthermore, we restrict ourselves to idempotent mappings, \ie, satisfying $p\circ p = p$. This restriction is without loss of generality. Indeed, for an improving node-wise mapping $p$ its compositional power $p^k$ will be idempotent for some $k$ (\eg, for $k$ = $(\max_{s}|\X_s|)!$, which turns all cycles in the map to identity) and provides equally good or better reduction with $p^k(\X) \subseteq p(\X)$.
Idempotent maps have two following properties. Let $X$ be a set and $p\colon X \to X$ idempotent.
\begin{itemize}
\item If $p(x) \neq x$ then no $y\in X$ is mapped to $x$;
\item For $Y = p(X)$ the restriction of $p$ to $Y$ is the identity map $x \mapsto x$ and there holds $p(X) = \{x \in X \mid p(x) = x\}$;
\end{itemize} 

It follows that knowing an improving mapping $p$, we can eliminate labels $(s,i)$ for which $p_s(i)\neq i$ and there will remain at least one global minimizer of $\f$. 

Given a mapping $p$, the verification of the improving property~\eqref{improving mapping} is NP-hard since already in the quadratic pseudo-Boolean case the verification of autarky property~\eqref{eq:autarky} is NP-hard~\cite{Boros:TR06-probe}. 
A tractable sufficient condition will be constructed by embedding the mapping into the space $\Real^\I$ and applying the relaxation there.
\subsection{Relaxed Improving Mapping}\label{sec:relaxed-improviing}
\begin{definition}\label{def:relaxed-improving}
A {\em linear extension} of $p\colon \LL \to \LL$ is a linear mapping $P\colon \Real^\I \to \Real^\I$ that satisfies
\begin{equation}\label{p-ext}
(\forall x\in\LL)\tab \delta(p(x)) = P\delta(x).
\end{equation} 
\end{definition}
\begin{figure}
\centering
\includegraphics[width=\linewidth]{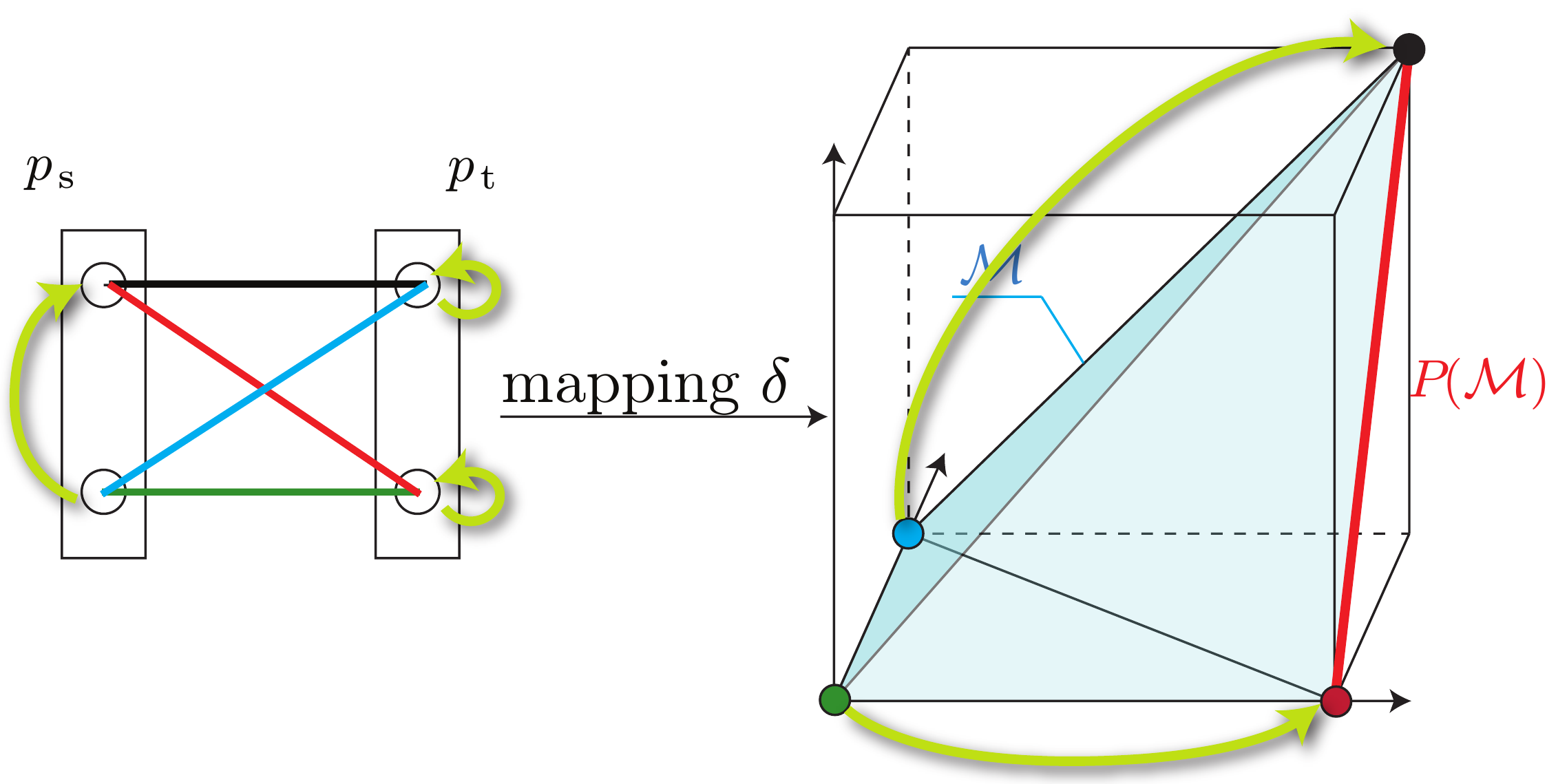}
\caption{Embedding of a discrete mapping in $\Real^\I$ (continues the example in \Figure{f:embed1}).
Left: discrete node-wise mapping $p\colon \X\to\X$ is shown by arrows, it sends the green labeling to red and the blue one to black. 
Right: there is a corresponding linear map $P\colon \Real^\I\to\Real^\I$ with this action on labelings embedded as vertices. It is an oblique projection which maps polytope $\M$ onto the red facet $P(\M)$.}\label{f:embed2}
\end{figure}
See \Figure{f:embed2} for illustration. Avoiding the discussion of uniqueness\footnote{When a linear extension exists, its restriction to the affine hull of $\delta(\X)$ is unique.}, we will only use the following linear extension for a node-wise mapping $p\colon \LL \to \LL$, which will be denoted $[p]$. 
%
The linear extension $P=[p]$ is defined by
\begin{align}\label{pixel-ext}
(P\mu)_{\c}(x_\c) = \sum_{x_\c'} P_{\c,x_\c,x'_\c} \mu_{\c}(x_\c')
\end{align}
with coefficients
\begin{align}\label{pixel-ext-comp}
P_{\c,x_\c,x'_\c} := \prod_{s\in \c} \leftbb p_s(x'_s){=}x_s\rightbb = \leftbb p_\c(x'_\c){=}x_\c \rightbb.
\end{align}
These coefficients should be understood as a ``matrix'' representation of $P$.
To verify that~\eqref{p-ext} holds true we simply substitute an integer labeling $\delta(x)$ and expand the components as
\begin{equation}
\begin{aligned}
(P\delta(x))_{\c}(x'_\c) & = 
\sum_{x''_\c} \prod_{s\in \c} \leftbb p_s(x''_s){=}x'_s \rightbb \leftbb x_{\c}{=}x''_\c \rightbb\\
& = \delta(p(x))_{\c}(x'_\c).
\end{aligned}
\end{equation}
%
\par
Using the linear extension $P$ of $p$ we can write
\begin{equation}
\f(p(x)) = \<f, \delta (p(x))\>  = \<f, P \delta(x) \>. 
\end{equation}
This allows to express the condition of improving mapping~\eqref{improving mapping} as 
\begin{equation}\label{improving-conj}
(\forall x\in\LL)\tab  \<f, P \delta(x) \> \leq  \<f, \delta(x) \>
\end{equation}
or equivalently, fully in the embedding, as
\begin{equation}\label{improving-conj-m}
(\forall \mu\in \delta(\X))\tab  \<f, P \mu \> \leq  \<f, \mu \>.
\end{equation}
Taking convex combinations in~\eqref{improving-conj-m}, we obtain an equivalent condition
\begin{equation}\label{improving-conj-M}
(\forall \mu\in \M)\tab  \<f, P \mu \> \leq  \<f, \mu \>.
\end{equation}
Thus we have linearized the inequalities necessary for an improving mapping. However, the marginal polytope $\M$ is not tractable. We introduce a sufficient condition by requiring that the same inequality~\eqref{improving-conj-M} is satisfied over a larger (tractable) polytope $\Lambda \supset \M$.
\begin{definition}
A linear mapping $P \colon \Real^\I \to \Real^\I$ is (weak) {\em $\Lambda$-improving} for $f$ if
\begin{align}\label{L-improving-conj-a}
& (\forall \mu\in\Lambda)\tab  \<f, P \mu \> \leq \<f, \mu \>;
\end{align}
and is {\em strict $\Lambda$-improving} for $f$ if
\begin{align}
\label{Def:SI-eq}
&(\forall \mu\in\Lambda,\ P\mu \neq \mu) \tab  \<f, P \mu \> < \<f, \mu \>.
\end{align}
\end{definition}
%
%
\par
\begin{statement}Let $P\colon\Real^\I\to\Real^\I$ be a linear extension of $p\colon \X\to \X$ and $\Lambda$ a relaxation polytope. If $P$ is $\Lambda$-improving for $f$ then $p$ is improving for $f$.
\end{statement}
\begin{proof}
A relaxed-improving mapping $P$ satisfies inequality~\eqref{L-improving-conj-a} over a superset $\Lambda$ of $\M$, therefore condition~\eqref{improving-conj-M} is satisfied, which by the definition of extension is equivalent to~\eqref{improving mapping}.
\end{proof}
\par
The set of mappings for which~\eqref{L-improving-conj-a} (resp. ~\eqref{Def:SI-eq}) is satisfied will be denoted \WI (resp. \SI). For convenience, we will use the term {\em relaxed improving} when the relaxation is clear from the context.
\par
Naturally, a strict relaxed improving map is relaxed improving, \ie, $\SI\subset \WI$. This is so because for all $\mu\in\Lambda$ such that $P\mu = \mu$ the inequality~\eqref{L-improving-conj-a} is trivially satisfied.
\par
Next we show that the verification of $P\in\WI$ (resp. $P\in\SI$) for a given $P$ can be solved (decided) in polynomial time.
The definition~\eqref{L-improving-conj-a} of $P\in\WI$ is equivalent to the expression
\begin{equation}\label{L-LP}
\min_{\mu\in\Lambda} \<(I-P\T)f,\mu \> \geq 0.
\end{equation}
The optimization problem in~\eqref{L-LP} will be therefore called the {\em verification LP}. As a linear program over a tractable polytope $\Lambda$, it can be solved in polynomial time and hence the decision problem $P\in\WI$ is solvable in polynomial time.
\par
In order to show that the verification of $[p]\in\SI$ can also be decided in polynomial time we introduce the following equivalent reformulation.
\begin{restatable}{statement}{SIO}\label{SI-O}Let $\O = \argmin\limits_{\mu\in\Lambda}\<f,(I-P)\mu\>$. 
There holds $P \in\SI$ iff
\begin{align}
\label{Def:SI-b}
\O & = P(\Lambda).
\end{align}
\end{restatable}
The statement says that a strictly relaxed improving mapping must not change the set of all optimal  solutions to the verification LP. This can be further expressed in components of the mapping and of the support set $\O$:
%
%
\begin{restatable}{statement}{TSIcomponents}\label{T:SI-components}
Let $\O_\c = \{x_\c\in\X_\c \mid (\exists \mu \in \O )\ \mu_\c(x_\c)>0 \}$. There holds $[p]\in\SI$ iff
\begin{align}
\label{SI-components}
(\forall \c\in\E)\ \ \O_\c = p_\c(\X_\c).
\end{align}
\end{restatable}
Now, in order to solve the verification of $p\in \SI$ in polynomial time we can 
solve the verification LP in~\eqref{L-LP}, obtain $\c$-support sets of its optimal solutions $\O_\c$ and check condition~\eqref{SI-components}.
\par
\subsection{Properties}\label{sec:properties}
%
We next give necessary conditions for $p$ in order that $[p] \in \WI$ or $[p] \in \SI$. They help to narrow down the set of maps to be considered. A relaxed improving map must preserve optimality of solutions to the relaxation and consequently their support set (again in components).
%
\begin{restatable}[Necessary conditions I]{lemma}{necessaryLI}\label{necessary-LI} 
Let $p\colon \X\to \X$ be node-wise and $P=[p]$. Let $\O = \argmin_{\mu\in\Lambda}\<f,\mu\>$ and $\O_\c = \{x_\c \in \X_\c \mid (\exists \mu\in \O)\ \ \mu_{\c}(x_\c) > 0 \}$. Then
\begin{enumerate}
\item[(i)]For $p\in\WI$ there holds 
\begin{subequations}
\begin{align}\label{necessary-wi-gen}
& P(\O) \subset \O;\\
\label{necessary-wi-comp}
& (\forall \c \in \E)\ \ p_\c(\O_\c) \subset \O_\c; 
\end{align}
\end{subequations}
\item[(ii)]For $p\in\SI$ there holds
\begin{subequations}
\begin{align}
\label{necessary-si-gen}
& P(\O) = \O;\\
\label{necessary-si-comp}
& (\forall \c \in \E)\ \ p_\c(\O_\c) = \O_\c; 
\end{align}
\end{subequations}
\end{enumerate}
\end{restatable}
Next, we reformulate problems $[p] \in \WI$ and $[p] \in \SI$ dually, \ie, not with quantifier $(\forall x\in \Lambda)$ as in~\eqref{L-improving-conj-a} but with existence quantifiers. This will become important in the formulation of the maximum persistency problem where we optimize over $p$ subject to the constraints $[p] \in \WI$ (resp. $[p] \in \SI$). 
%
Recall that the set $\WI$ is defined for the relaxation polytope $\Lambda = \{\mu \in \Real^\I \mid A\mu \geq 0;\ \mu_\emptyset = 1;\ \mu \geq 0\}$, where $A\in\Real^{m \times |\I|}$.
\begin{theorem}[Dual representation of $\WI$]\label{S:WI-dual} Set $\WI$ can be expressed as
\begin{equation}\label{WI-dual}
\{P\colon\Real^\I\to\Real^\I \mid (\exists \varphi\in \Real^m_+)\ \ f - A\T \varphi - P\T f \geq 0 \}.
\end{equation}
\end{theorem}
\begin{proof}
Denote $g = (I-P\T)f$. Condition~\eqref{L-LP}, equivalent to~\eqref{L-improving-conj-a}, can be stated yet equivalently for the conical hull of $\Lambda$: 
\begin{equation}\label{L-improving-coni}
\inf_{\mu\in\coni(\Lambda)} \<g, \mu \> \geq 0.
\end{equation}
This is because for any $\mu \in \Lambda$ and any $\alpha \geq 0$ vector $\alpha \mu$ will satisfy RHS of~\eqref{L-improving-conj-a} as well.
Using the expression for the conical hull of $\Lambda$ in~\eqref{coni-Lambda}, we can write the minimization problem in~\eqref{L-improving-coni} and its dual as\\[-5pt]
\begin{equation}\label{LP'}
\begin{array}{rclr}
& \inf \<g,\mu\> &\ \ 
& \max 0\,.\\
&
\setlength{\arraycolsep}{0.2em}
\begin{array}{rl}
A \mu & \geq 0 \\
\mu & \geq 0
\end{array}
& &
\setlength{\arraycolsep}{0.2em}
\begin{array}{rl}
\varphi & \in \Real^m_+\\
g - A\T \varphi & \geq 0
\end{array}
\end{array}
\end{equation}
Inequality~\eqref{L-improving-coni} holds iff the primal problem is bounded, and it is bounded iff the dual is feasible, which is the case iff $(\exists\varphi\in\Real^m_+)$ $(f-A\T\varphi)-P\T f \geq 0$.
\qqed 
\end{proof}
\par
The set $\SI$ is defined via a more complicated quantifier $(\forall \mu \in \Lambda, P \mu \neq \mu)$. Fortunately, the following dual reformulation holds for node-wise maps:
\begin{restatable}[Dual representation of $\SI$]{theorem}{SSIdual}\label{S:SI-dual} Let $p\colon \Labels\to \Labels$ be node-wise. Then: (i) there exists $\epsilon>0$ such that $[p] \in \SI$ iff
\begin{align}
\label{SI-dual}
&\ \ (\exists \varphi\in \Real^m_+)\ \ f - A\T \varphi - [p]\T f \geq \epsilon h,
\end{align}
where $h$ is a function such that $h \geq 0$ and $h_\c(x_\c) = 0$ iff $p_\c(x_\c) = x_\c$; 
and (ii) for rational inputs (including $h$) the value of $\varepsilon$ in (i) is a rational number of polynomial bit length.
\end{restatable}
The constraint $[p]\in\SI$ can thus be reduced to nearly the same representation as~\eqref{WI-dual}, with an addition of an $\varepsilon h$ slack term. By construction, this term is zero iff $[p]\mu = \mu$.
In practice, taking a larger value of $\varepsilon$ always results in a sufficient condition for $\SI$ and hence does not break correctness. 
In theory, we want a very small $\varepsilon$ but not so small that it would break polynomiality of the reformulation, which is ensured by part (ii). Note, while the set $\SI$ in the space of all maps $\Real^\I \to \Real^\I$ was convex but not closed (as seen from definition~\eqref{Def:SI-eq}), the theorem encloses the discrete maps of our interest, $\{[p] \mid p\colon \X \to \X \mbox{ node-wise} \}$ in a closed (convex) polytope.
\par
Finally we give a necessary condition for $\WI$. 
The theorem has a primal and a dual counterpart. The primal counterpart states that when solving the verification LP, because its objective $(I-P\T)f$ is in the null space of $P$, the constrains of the problem can be projected onto the same subspace providing a simplification. The dual counterpart states that there always exist dual multipliers such that the improving property holds component-wise for reparametrized costs. This is useful in proofs, providing an alternative reformulation of local inequalities~\eqref{WI-dual}.
\begin{restatable}[Necessary conditions II]{theorem}{TChar}\label{T:Char}
Let $P\colon \Real^\I\to \Real^\I$ be idempotent, $P(\Lambda)\subset \Lambda$ and $P \in \WI$. Then 
\begin{subequations}
\begin{align}
\label{necessary II - inf}
& \inf\csub{
A(I-P)\mu \geq 0 \\ \mu\geq 0} \<(I-P)\T f,\mu\> = 0;\\
\label{component-equalities}
& (\exists \varphi\in\Real^m_+)\ \ (I-P\T) (f-A\T\varphi) \geq 0.
\end{align}
\end{subequations}
\end{restatable}
These conditions become necessary and sufficient for standard relaxations as discussed in \Section{sec:L1-s}. The constraint $A(I-P)\mu \geq 0$ in~\eqref{necessary II - inf} replaces the constraint $A\mu\geq 0$ in~\eqref{LP'} and simplifies the problem.
%
\subsection{Maximum Relaxed Improving Mapping}\label{sec:maximum persistency}
We showed in~\Section{sec:relaxed-improviing} that weak/strict relaxed-improving property can be verified in polynomial time and have described sets $\WI$, $\SI$. Any relaxed-improving map, with the exception of the identity, eliminates some labels as non-optimal.
%
%
%
Recall that the label $(s,i)$ is eliminated by node-wise mapping $p$ if $p_s(i)\neq i$. We formulate the following {\em maximum persistency} problem:
%
\begin{align}\label{best L-improving}
\tag{{\sc max-wi}}
\max_{p} \sum_{s,i} \leftbb p_s(i){\neq i}\rightbb \tab \mbox{s.t. $[p]\in\WI$,}\quad
\end{align}
\ie we directly maximize the number of eliminated labels.
The strict variant, with constraint $[p]\in\SI$, will be denoted {\bf\sc max-si}.

\begin{table}
\begin{center}
\small
\setlength{\tabcolsep}{3pt}
\renewcommand{\arraystretch}{1.2}
\begin{tabular}{|c|c|c|c|c|c|}
\hline
order & labels & maps & \maxsi & \maxwi\\
\hline
pairwise  & 2  & all & P  & P \\
higher-order & 2  & all & P  & NP-hard\\
any & any & $\P^{1,y}$ or $\P^{2,y}$ & P & P \\
any & any & $\P^{1}$ & P & NP-hard \\
pairwise & ${\geq}4$ & $\P^{2}$ & NP-hard & NP-hard \\
\hline
\end{tabular}\\[5pt]
\end{center}\caption{Polynomiality of finding the maximum strict/weak relaxed-improving mapping for a general relaxation. 
}\label{table:P-cases}
\end{table}
\par
The problem may look difficult to solve. Indeed, it optimizes over discrete maps and involves a general polyhedral relaxation in the specification of constraints. Nevertheless, if we place some additional restrictions on the set of maps, it turns out to be solvable in polynomial time in a number of cases summarized in Table~\ref{table:P-cases}. One of them is the pseudo-Boolean case, where there are only 3 possible idempotent maps for every node: $(0,1) \mapsto (1,1)$, $(0,1) \mapsto (0,0)$ and $(0,1) \mapsto (0,1)$. Problem \maxsi turns out to be solvable in this case. For multilabel problems, node-wise mappings are more diverse. Motivated by the goal to include/generalize existing multilabel methods, the following sets of maps are introduced: 
%
%
%
%
\par
{\em all-to-one maps}. The set $\P^{1,y}$ of maps $p$ of the form $p \colon x \mapsto x [\A{\leftarrow}y]$ for all $\A\subset\V$ and fixed $y\in \LL$. This class is a straightforward generalization of the overwrite operation in the autarky~\eqref{eq:autarky}. A mapping $p\in\P^{1,y}$ is illustrated in \Figure{fig:Kovtun}(a). There are only two possible choices for every node $s$. The mapping $p_s$ either contracts $\X_s$ to a single label $\{y_s\}$ or retains $\X_s$ unchanged.
This class allows to explain one-against-all method of \citet{Kovtun03} and the central part of the method of \citet{Swoboda-14} as discussed in \Section{sec:kovtun}, \Section{sec:swoboda}.
\par
{\em all-to-one-unknown} maps. Set $\P^1 = \bigcup_{y\in\LL} \P^{1,y}$. A mapping $p\in\P^1$ has the same form as above, $p \colon x \mapsto x [\A{\leftarrow}y]$, however the labeling $y$ is not fixed now but a part of the specification of the mapping, see \Figure{fig:Swoboda}. In every node there are $|\X_s| + 1$ choices for $p_s$: send all labels to a single one (which may be chosen) or change nothing. It is easy to see that in the case of two labels, $\P^1$ contains all idempotent node-wise maps.
As will be shown later the \maxsi problem over this class decomposes into sufficient conditions to determine $y$ from the integral part of the solution to the relaxation and the \maxsi problem over $\P^{1,y}$.
\par
\begin{figure}
\centering
\includegraphics[width=0.9\linewidth]{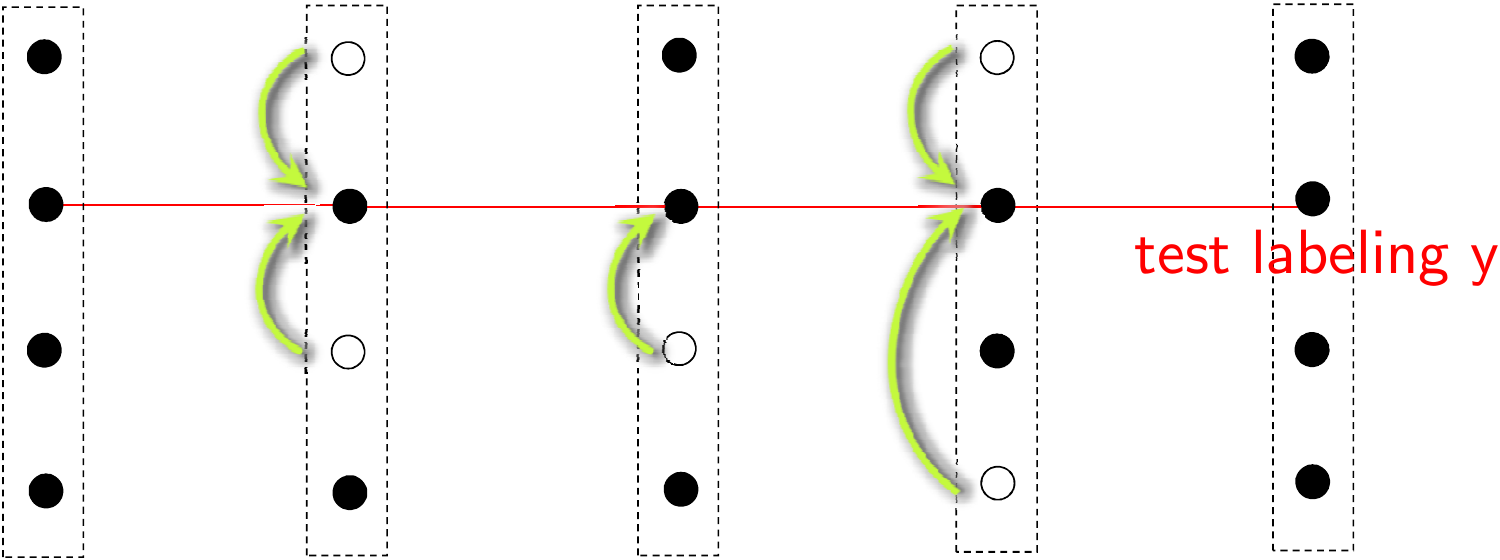}
\caption{Example of a map in the subset-to-one class $\P^{2,y}$. Labeling $y$ is fixed while a map $p$ can select a subset of labels in every node $s$ that are sent to $y_s$. Nodes without an outgoing arrow are mapped to themselves. }
\label{fig:subset-maps}
\end{figure}

{\em subset-to-one maps}. 
The set of maps $\P^{2,y}$ is defined as follows. Let $V = \{(s,i)\mid s\in\V, i\in \LL_s\}$ -- the set of labels in all nodes. Let $\zeta \in\{0,1\}^V$. Mapping $p_\zeta \in \P^{2,y}$ in every node either preserves the label $x_s$ or overwrites it with $y_s$:
\begin{equation}\label{subset-to-one-map}
p_\zeta(x)_s = 
\begin{cases}
y_s &\ \IF\ \zeta_{s, x_s} = 0,\\
x_s &\ \IF\ \zeta_{s, x_s} = 1.
\end{cases}
\end{equation}
Vector $(\zeta_{s,i}\mid i\in\LL_s)$ serves as the indicator of the subset of labels in node $s$ that stay immovable while all other labels are mapped to $y_s$, see \Figure{fig:subset-maps}. In a node $s$ there are $2^{|\X_s|-1}$ choices for $p_s$. Clearly, this class generalizes $\P^{1,y}$. 
\par
The main result of this paper is that both \maxwi and \maxsi problems are tractable for the class $\P^{2,y}$. Other tractability results in Table~\ref{table:P-cases} are obtained as corollaries. Intractability results are shown to hold for the basic LP relaxation in \Section{sec:L1-s}.

\subsection{Formulation for Subset-to-one Maps}
In the following three subsections we gradually show that~\maxwi problem over $\P^{2,y}$ class can be written as a mixed integer linear program in which integrality constraints can be relaxed without loss of tightness and thus we obtain an equivalent LP formulation. 
\par
Using the dual representation of the constraint $[p]\in\WI$~\eqref{WI-dual} and the form of the mapping~\eqref{subset-to-one-map}, the problem \eqref{best L-improving} becomes 
\begin{subequations}\label{best L-improving-W}
\begin{align}
& \min \sum_{s\in\V}\sum_{i\in \X_s} \zeta_{s,i}\\
\notag
& (\zeta_{s,i}\in\{0,1\} \mid s\in\V,\ i\in\X_s); \  \varphi \in \Real^m_+;\\
\label{p_zeta-complicated}
& f - A\T \varphi -[p_\zeta] \T f \geq 0.
\end{align}
\end{subequations}

\par
Constraints~\eqref{p_zeta-complicated} involve a complicating expression $[p_\zeta]$.
Let us express coefficients $P_{c,\x_\c,x'_\c}$~\eqref{pixel-ext-comp} of the linear extension $P=[p_\zeta]$. 
Substituting mapping $p_\zeta$~\eqref{subset-to-one-map} they are expressed as polynomials in $\zeta$:
\begin{align}\label{P-zeta-components}
P_{\c,x_\c, x'_\c}  = &\prod_{s\in \c} \leftbb p_\zeta(x')_s{=} x_s \rightbb\\
\notag
 = &\prod_{s\in \c}\Big(\leftbb x'_s{=}x_s \rightbb \zeta_{s,x'_s} + \leftbb y_s{=}x_s \rightbb(1-\zeta_{s,x'_s}) \Big).
\end{align}
It appears that after expanding $[p_\zeta]$ using~\eqref{P-zeta-components} the constraint that we need to represent~\eqref{p_zeta-complicated} will involve products of binary variables $\prod_{s\in \d}\zeta_{s, x'_s}$ for all $\c\in\E$, $\d\subset\c$ and $x'_\d \in\X_\d$. To reach the ILP formulation we are going to replace each such product with a substitute variable $\zeta_{\d,x'_\d}$. This is achieved with the help of the relaxation of~\citet{SheraliA90}.
\subsection{Relaxation of Sherali and Adams}\label{sec:SA-relaxation}
The relaxation of \citet{SheraliA90} is applicable to polynomial programs with binary variables $z\in\{0,1\}^\V$. The relaxation of order $d$ performs a simultaneous lifting for all subsets of variables $\c\subset \V$ with $|\c| = d$. 
\begin{table}
\normalfont
\normalsize
{
\centering
\setlength{\tabcolsep}{3pt}
\renewcommand{\arraystretch}{1.2}
\begin{tabular}{p{0.47\linewidth}cp{0.4\linewidth}}
\hline
monomial & & new variable \hfill \anchor[S1]{Sone}\\
$\prod_{s\in\d} z_s$, $\d\subset\c$ & $\rightarrow$ & $\zeta_{\d} \in \Real$, $\zeta_\emptyset = 1$ \\
\hline
multilinear polynomial & & linearization \hfill \anchor[S2]{Stwo}\\
$g(z) = \sum_{\d \subset \c} \alpha_\d \prod_{s\in\d} z_s$ & $\rightarrow$ & $G(\zeta) = \sum_{\d\subset \c} \alpha_\d \zeta_\d$ \\
\hline
linearization properties & & \hfill \anchor[S3]{Sthree}\\
$g_1(z) + g_2(z)$ & $\rightarrow$ & $G_1(\zeta) + G_2(\zeta)$\\
$\alpha g(z)$ & $\rightarrow$ & $\alpha G(\zeta)$\\
$\forall z\in \{0,1\}^\c\ g(z) = 0$ & $\Leftrightarrow$ & $G(\zeta) \equiv 0$\\
\hline
identity inequality \newline for $\b\subset\c$
&  & new constraint \hfill \anchor[S4]{Sfour}\\ 
$\prod\limits_{s\in\b} z_s \prod\limits_{s\in\c\backslash \b} (1-z_s) \geq 0 $
& $\rightarrow$ & $\sum\limits_{\d\subset\c\backslash \b}(-1)^{|\d|}\zeta_{\d\cup \b} \geq 0$ \\
\hline
any other \newline identity inequality & & \hfill \anchor[S5]{Sfive} \newline identity inequality\\
$\forall z\in \{0,1\}^\c \ \ g(z) \geq 0$ & $\Leftrightarrow$ & $(\forall \zeta\in\Zeta_\c)\ G(\zeta) \geq 0$\\
\hline
\end{tabular}
\vskip2mm
}
\caption{Summary of correspondences in the relaxation approach~\cite{SheraliA90} within a hyperedge $\c$.}\label{SA-table}
\end{table}
Let us focus on a single hyperedge $\c$ chosen for generality from the set of hyperedges $\E \subset 2^\V$. The construction and its properties (within hyperedge $\c$) are summarized in Table~\ref{SA-table}.
For every product $\prod_{s\in\d}z_s$, $\d\subset\c$, a new variable $\zeta_{\d}$ is introduced \Sone. A pseudo-Boolean function $g \colon \{0,1\}^\c \to \Real$ is linearized by writing it as a multilinear polynomial and replacing each monomial $\prod_{s\in\d}z_s$ with the new variable $\zeta_{\d}$, \Stwo. From this definition we have linearity properties \Sthree, in particular:
\begin{restatable}[Identity Equality \Sthree]{lemma}{SAequality}\label{L:SA-equality} Let $G(\zeta)$ be the linearization of $g(z)$. Then $g(z) = 0$ for all $z\in\{0,1\}^\c$ iff $G(\zeta) = 0$ for all $\zeta\in\Real^{2^\c}$.
\end{restatable}
\par
Next, constraints on new variables are added which correspond to identity inequalities $\prod\limits_{s\in\b} z_s \prod\limits_{s\in\c\backslash \b} (1-z_s) \geq 0$ for each $\b\subset\c$. Clearly this inequality holds for all $z\in\{0,1\}^\c$.
By expanding this expression one obtains its equivalent multilinear polynomial
 $g(z) = \sum\limits_{\d\subset\c\backslash \b}(-1)^{|\d|}\prod_{s\in\d\cup\b}z_s \geq 0$. 
Constraints \Sfour ensure that the linearization of this expression is non-negative.
The set of all such constraints defines the polytope
\begin{align}\label{Zeta-C-polytope}
\Zeta_\c = \{ \zeta \in\Real^{2^\c} \mid
\zeta_\emptyset = 1,\ (\forall \b\subset \c)\ \sum\limits_{\d\subset\c\backslash \b}(-1)^{|\d|}\zeta_\d \geq 0
\}.
\end{align}
\par
In fact, polytope $\Zeta_\c$ is the convex hull of all binary vectors $\zeta$ corresponding to configurations $z$: 
\begin{restatable}[Convex hull]{lemma}{SAconvexhull}\label{L:SA-convex-hull}
 Polytope $\Zeta_\c$ equals the convex hull
\begin{align}\label{zeta-set}
\conv \Big\{ \zeta(z)\in\Real^{2^\c} \Mid \zeta(z)_\d = \prod_{s\in\d}z_s,\ \forall \d\subset\c, \forall z\in\{0,1\}^\c \Big\}.
\end{align}
\end{restatable}
%
From the convex hull representation there naturally follows an equivalence of identity inequalities before and after linearization:
\begin{restatable}[Identity inequality \Sfive]{lemma}{SAinequality}\label{L:SA-inequality}Let $G(\zeta)$ be the linearization of $g(z)$. Then $g(z) \geq 0$ for all $z\in\{0,1\}^\c$ iff $G(\zeta) \geq 0$ for all $\zeta \in \Zeta_\c$. 
\end{restatable}
%
In particular, for $\zeta\in\Zeta_\c$ there holds $0 \leq \zeta_{\d} \leq 1$ for $\d\subset\c$, a relation which is rather difficult to prove directly form~\eqref{Zeta-C-polytope}. 
Finally, for our construction the next two results are necessary.
\begin{theorem}[Lemma 2 of~\cite{SheraliA90}]\label{T:SA} If $\zeta \in \Zeta_\c$ and unary components $\zeta_{s}$ are integer (\ie, equal to some $z_s\in\{0,1\}$) for all $s\in\c$, then there holds $\zeta_{\d} = \prod_{s\in \d}z_{s}$ for all $\d\subset\c$.
\end{theorem}
%
%
\begin{restatable}[Product]{lemma}{SAproduct}\label{L:product}\label{L:SA-product}
For $\zeta \in \Zeta_\c$ there holds $\zeta^2 \in \Zeta_\c$, where the product $\zeta^2 = \zeta\zeta$ is component-wise. 
\end{restatable}
When applying the linearization to all hyperedges simultaneously, a variable $\zeta_{\a \cap \b}$ is introduced only once for (overlapping) hyperedges $\a,\b\in\E$. All local properties described above continue to hold for each hyperedge $\c\in\E$ individually but of course they need not hold for the whole set $\V$.
\subsection{Solution via Linear Program Formulation}\label{sec:LP-solution} 
Let us return to the reformulation~\eqref{best L-improving-W} of \maxwi. It is clear that by opening brackets in~\eqref{P-zeta-components}, the coefficients $P_{\c,x_\c, x'_\c}$ can be expressed as
\begin{align}\label{P_c-poly-expression}
P_{\c,x_\c, x'_\c} = \sum_{\d \subset \c} c_{\c,\d}(x_\c,x'_\d) \prod_{s\in \d}\zeta_{s,x'_s},
\end{align}
where $c_{\c,\d}(x_\c,x'_\d)$ are appropriate constants not depending on $\zeta$ (detailed in \Section{L1-proofs}).
Because for $x_s'=y_s$ there holds $p_\zeta(y)_s = y_s$ irrespectively of $\zeta_{s,y_s}$ (label $y_s$ is always mapped to itself) we may assume that $\zeta_{s,y_s} = 0$ as well as all products involving it.
\par
The relaxation of Sherali and Adams is applied as follows. Let us denote $\tilde \X_s = \X_s\backslash\{y_s\}$ and respectively $\tilde \X_\c = \prod_{s\in\c} \tilde \X_s$. We substitute new variables $\zeta_{\d, x'_\d} \in \Real$ in place of products $\prod_{s\in \d}\zeta_{s, x'_s}$ in~\eqref{P_c-poly-expression}. For zero products, \ie, for $x'_\d\notin \tilde \X_\d$, we let $\zeta_{\d, x'_\d} = 0$. 
From now on, let $\zeta$ denote the vector of relaxed variables 
\begin{align}\label{zeta-relax}
\zeta = (\zeta_{\d, x_\d} \in\Real \mid \forall \c \in\E,\ \d\subset \c, x_\d \in \X_\d).
\end{align}
New variables $\zeta$ must satisfy the following constraints, defining a polytope $\Zeta$:
%
\begin{subequations}\label{Zeta-set}
\begin{align}\label{Zeta-set-a}
& \zeta_\emptyset = 1,\\
\label{Zeta-set-b}
&(\forall \c\in\E,\ \forall x'_\c\in\X_\c \backslash \tilde\X_\c,\ \forall \d\subset\c)\ \ \zeta_{\d, x'_\d} = 0,\\
\label{Zeta-set-c}
&(\forall \c\in\E,\ \forall x'_\c\in\tilde\X_\c,\ \forall \d\subset\c)\ \ \sum\limits_{\d\subset\c\backslash \b}(-1)^{|\d|}\zeta_{\d,x'_\d} \geq 0.
\end{align}
\end{subequations}
\par
Polytope $\Zeta$ is the intersection of polytopes $\Zeta_\c$~\eqref{Zeta-C-polytope} lifted to the space of all variables over $\c\in\E$ and $\x'_\c \in\tilde\X_\c$.
Let $P_\zeta$ denote the extension-linearization of $p_\zeta$~\eqref{subset-to-one-map}, according to~\eqref{P_c-poly-expression} and \Stwo defined by:
\begin{align}
(P_\zeta)_{\c,x_\c,x'_\c} = \sum_{\d \subset \c} c_{\c,\d}(x_\c,x'_\d) \zeta_{\d,x'_\d}.
\end{align}
For our purpose it is necessary that the linearized map $P_\zeta$ preserves the relaxation polytope $\Lambda$: $P_\zeta(\Lambda) \subset \Lambda$. This constraint expresses as $(\forall \mu\in\Lambda)$
\begin{subequations}\label{P preserve Lambda}
\begin{align}
\label{P_zeta c0}
& (P_\zeta \mu)_\emptyset = 1;\\
\label{P_zeta c1}
& P_\zeta \mu \geq 0; \\
\label{P_zeta c2}
& A P_\zeta \mu \geq 0.
\end{align}
\end{subequations}
We trivially have $(P_\zeta)_\emptyset = \zeta_\emptyset = 1$. It is also easy to show that $(P_\zeta)_{\c,x_\c,x'_\c} \geq 0$ for $\zeta\in\Zeta$: before linearization, coefficients $P_{\c,x_\c,x'_\c}$ in the expression~\eqref{P-zeta-components} are clearly non-negative and by property \Sfive it is guaranteed that $(P_\zeta)_{\c,x_\c,x'_\c} \geq 0$ holds on $\Zeta$. Then for $\mu \geq 0$ there holds $P_\zeta \mu \geq 0$.
Interestingly, the converse is also true (but this result is not necessary in the subsequent construction):
\begin{restatable}{theorem}{TSAequiv}\label{T:SA-equiv}
Inequalities~\eqref{Zeta-set-c} in the definition of polytope $\Zeta$ can be equivalently replaced with $P_\zeta \geq 0$.
\end{restatable}
There remains constraint~\eqref{P_zeta c2}.
In the case of standard local relaxations (to be defined later) constraint~\eqref{P_zeta c2} holds automatically and needs not be enforced. 
To account for general relaxations, we include constraint~\eqref{P_zeta c2} explicitly by representing it similarly to \Theorem{S:WI-dual} in the dual form as:
\begin{equation}
(\exists \Phi\in \Real^{m \times |\I|}_+)\ \ A P_\zeta - \Phi\T A \geq 0.
\end{equation}
We arrive at the following relaxation of \maxwi as 
a linear program: 
\begin{subequations}\label{best L-improving ILP}
\begin{align}
\label{L1}
\tag{L1}
& \min\csub{\zeta,\varphi, \Phi} \sum_{s,i}\zeta_{s,i} \\ 
\label{dual-WI-constraint}
& (I-P\T_\zeta)f - A\T \varphi \geq 0;\ \ \varphi \geq 0;\\
\label{IL1-closed}
& A P_\zeta - \Phi\T A \geq 0;\ \ \Phi \geq 0;\\
\label{IL1-closed2}
&\zeta \in \Zeta. 
\end{align}
\end{subequations}
Constraint~\eqref{dual-WI-constraint} ensures that mapping $P_\zeta$ is relaxed-improving, constraints~\eqref{IL1-closed} that it preserves the polytope: $P_\zeta (\Lambda) \subset \Lambda$ and constraint~\eqref{IL1-closed2} ensures that for each $\c\in\E$ relaxed variables $(\zeta_{\d,x_\d} \mid \d\subset \c)$ stay in the local convex hull for $\c$.
\par
We claim that this relaxation is tight. 
As shown below, rounding down all components of $\zeta$ in a feasible solution maintains feasibility (with possibly different  values of $\varphi$, $\Phi$) and can only improve the objective. The rounding is performed by constructing the composite mapping $P_\zeta P_\zeta$. If $P_\zeta$ is relaxed-improving then so is $(P_\zeta)^2$ provided that it satisfies all feasibility constraints. The auxiliary lemma below establish this feasibility: it verifies that $P_\zeta^2 = P_{\zeta^2}$. 
Starting from a non-integer $\zeta$ and building a feasible sequence by taking $\zeta \mapsto \zeta^2$ we get each next point closer and closer to the integer limit.
\begin{restatable}{lemma}{Lsquaretozeta}\label{L:square to zeta}
For $\zeta\in\Zeta$ there holds $P_\zeta^2 = P_{\zeta^2}$. 
\end{restatable}
\begin{theorem}\label{T LP1 tight}
In a solution $(\zeta,\varphi, \Phi)$ to~\eqref{L1} vector $\zeta$ is integer. 
\end{theorem}
\begin{proof}
Because $\zeta$ is feasible to~\eqref{L1}, the mapping $P_\zeta$ is $\Lambda$-improving for $f$. Note, at this point, unless $\zeta$ is integer it is not guaranteed that $P_\zeta(\M) \subset \M$ and we cannot draw any partial optimality from it, neither $P_\zeta$ is guaranteed to be idempotent. 
By constraints~\eqref{IL1-closed},~\eqref{IL1-closed2}, 
there holds $P_\zeta (\Lambda) \subset \Lambda$. Therefore
\begin{equation}
(\forall \mu\in\Lambda)\tab \<f, P_\zeta P_\zeta \mu \> \leq \<f, P_\zeta \mu \> \leq \<f, \mu \>.
\end{equation}
It follows that $P_\zeta^2 =  P_\zeta P_\zeta$ is $\Lambda$-improving. Since $P_\zeta (\Lambda) \subset \Lambda$
, it is also $P_\zeta^2 (\Lambda) \subset P_\zeta (\Lambda) \subset \Lambda$.

By \Lemma{L:square to zeta}, there holds $P_\zeta^2 = P_{\zeta^2}$ and by \Lemma{L:product} $\zeta^2\in\Zeta$. By induction, there holds $P_\zeta^{2^n}  = P_{\zeta^{2^n}}$, $P_\zeta^{2^n}(\Lambda) \subset \Lambda$ and $P_\zeta^{2^n}$ is $\Lambda$-improving. 
%
Let 
\begin{equation}\label{xi star}
\zeta^*_{\c, x_\c}  = \lim_{n\to\infty} \zeta_{\c, x_\c}^{2^n} = \leftbb \zeta_{\c, x_\c}{=}1\rightbb.
\end{equation}
Since $P_{\zeta^*}$ is $\Lambda$-improving and $\zeta^*\in\Zeta$, it is feasible to~\eqref{L1}. 
Assume for contradiction that there exist $(s',i')$ such that $0< \zeta_{s'i'} < 1$. From~\eqref{xi star} we have $\zeta^*_{si} \leq \zeta_{si}$ for all $s,i$ and $\zeta^*_{s'i'} < \zeta_{s'i'}$. It follows that $\zeta^*$ achieves a strictly better objective value, which contradicts the optimality of $\zeta$. If all unary components $\zeta_{s,i}$ are integer then by \Theorem{T:SA} $\zeta$ is integer. 
\qqed 
\end{proof}
%
\begin{corollary}The optimal solution to~\eqref{L1} is unique.
\end{corollary}
\begin{proof}
Assume for contradiction that $\zeta_1, \zeta_2$ are two distinct integer solutions to~\eqref{L1}. Since~\eqref{L1} is a linear program, their combination $\zeta = (\zeta_1+\zeta_1)/2$ is an optimal solution (values of $\varphi$, $\Phi$ are omitted for clarity). But if $\zeta_1 \neq \zeta_2$ then $\zeta$ is not integer, which contradicts \Theorem{T LP1 tight}. 
\end{proof}
Clearly, for an integer vector $\zeta\in\Zeta$ the linearization $P_\zeta$ coincides with the extension of the discrete mapping $p_\zeta$~\eqref{subset-to-one-map}. It follows that the unique optimal solution to~\eqref{L1} is the solution to \maxwi.
%
%
%
%
\subsection{Perturbation for Strong Persistency}
Problem \maxsi over $\P^{2,y}$ can be reduced to \maxwi with a perturbed cost vector $\tilde f$ as follows. 
It is sufficient to show that dual representation~\eqref{SI-dual} of constraint $[p] \in \SI$ can be reduced to that of $[p] \in \WIx{\tilde{f}}$. 
For $p\in\P^{2,y}$ we can choose components of vector $h$ in the dual representation~\eqref{SI-dual} of $\SI$ as
\begin{subequations}\label{perturbation-vector}
\begin{align}
& h_\c(x_\c) = g_\c(p_\c(x_\c))-g_\c(x_\c),\\
& g_\c(x_\c) = \sum_{ s\in\c} \leftbb x_s{=}y_s \rightbb.
\end{align}
\end{subequations}
Clearly, $p_\c(x_\c) = x_\c$ iff $h_\c(x_\c) = 0$ and for $p_\c(x_\c) \neq x_\c$ there holds $1 \leq h_\c(x_\c) \leq |\c|$. With such a vector $h$ the dual representation of $\SI$ can be written as
\begin{align}
(f+\varepsilon g) - [p]\T (f+\varepsilon g) - A\T\varphi \geq 0,
\end{align}
\ie, the same constraint as~\eqref{WI-dual} must hold but for an $\epsilon$-perturbed cost vector 
\begin{equation}\label{perturbation}
\tilde f := f+\varepsilon g.
\end{equation}
Since the solution $\zeta$ to the perturbed problem is integer and unique it is the optimal solution to \maxsi.
%
%
%
\subsection{Two-Phase Method}\label{sec:two-phase}
Let us consider the class $\P^{1,y}$. Formulation~\eqref{L1} can be adopted by incorporating additional constraints on $\zeta$ (making variables $\zeta_{s,x_s}$ equal for all $\x_s$). The proof of \Theorem{T LP1 tight} is based on the fact that for a feasible $\zeta$ also $\zeta^2$ is feasible. Clearly, this property is not destroyed by any equality constraints between components of $\zeta$. Therefore \Theorem{T LP1 tight} continues to hold and thus both \maxsi and \maxwi problems over $\P^{1,y}$ are tractable. 
%
%
\par
For class $\P^{1}$ the problem \maxsi can be solved as proposed by \Algorithm{Alg1}. It first solves the LP-relaxation in order to determine the {\em test labeling} $y$ and then solves the \maxsi problem for fixed $y$ using perturbed \eqref{L1} for class $\P^{1,y}$. 
\begin{algorithm}
	$\mu\in\argmin_{\mu\in\Lambda} \<f,\mu\>$\tcc*{solve~\eqref{LP}}
	For all $s$ if there exists $i\in\LL_s$ such that $\mu_{s}(i)=1$ then set $y_s = i$, otherwise set $y_s$ arbitrarily\;
	For strong persistency apply perturbation~\eqref{perturbation}\label{A:step-3}\;
	Solve the problem~\eqref{L1} for the class of maps $\P^{2,y}$ or $\P^{1,y}$\label{A:step-4}\;
	\caption{Two Phase Method\label{Alg1}}
\end{algorithm}
\begin{theorem}\Algorithm{Alg1} solves \maxsi over $\P^{1}$.
\end{theorem}
\begin{proof}
The necessary conditions of \Lemma{necessary-LI} for the optimal solution of LP-relaxation require that a strictly-improving mapping does not change optimal relaxed solutions. From the component-wise condition~\eqref{necessary-si-comp} follows that when $\mu_{s}$ if fractional for some $s$ then $p_s$ (assuming $p \in \P^{1}$) must be identity. When $\mu_{s}$ is integer, the only possible value of $y_s$ qualifying necessary conditions must correspond to $\mu_{s}(y_s) = 1$.
Applying perturbation in step~\ref{A:step-3} and optimizing over $\P^{1,y}$ in step~\ref{A:step-4} we obtain the optimal solution.
\end{proof}
\par
As a general heuristic, we can apply the same two-phase method, optimizing in step 4 over $\P^{1,y}$ or $\P^{2,y}$ with or without perturbation. The persistent assignment found by the heuristic is guaranteed to be at least as large as the solution of \maxsi over $\P^1$.
%
%
\section{Local LP Relaxations}\label{sec:Simplex-LP}
In this section we consider a special case of {\em local} (or standard) LP relaxations in energy minimization~\cite{Schlesinger-76,SheraliA90,Chekuri-01,Koster-98,Wainwright03nips}, see also the survey by \citet{Werner-PAMI07}. In our notation local relaxations are described by the polytope $\Lambda$ of the form 
\begin{align}\label{local-Lambda}
\Lambda = \{\mu\in\Real^\I \mid A\mu = 0;\ \mu_\emptyset=1;\ \mu \geq 0 \}.
\end{align}
\par
Recall that in the embedding $\delta$, different components of a relaxed labeling $\mu$, \eg, $\mu_\c$ and $\mu_\d$ for $\d \subsetneq \c$ represent overlapping subsets of variables. In order that they represent all discrete labelings consistently they must satisfy marginalization constraints of the form
\begin{equation}\label{marg-d}
(\forall x_\d \in \X_{\d}) \ \sum\limits_{x_{\c \backslash \d }} \mu_{\c}(x_\c) = \mu_\d(x_\d).
\end{equation}
\par
%
%
%
\par

\citet{Werner-10} considers a family of LP relaxations generated by enforcing constraint~\eqref{marg-d} for some pairs of subsets $\c\in\E, \d\subsetneq \c$. The set of such pairs is called the {\em coupling structure}~\cite{Werner-10}. 
For $\c,\d \subset \V$ we define coupling relation $\d \Esubset \c$ of order $d$: let $\d \Esubset \c$ iff
\begin{align}
\d \subsetneq \c,\ \c, \d \in \E,\ |\d|\leq d.
\end{align}
Subsequently, we will consider two possibilities: to include only first order constraints or all of them. 
%
%
Zero order constraints~\eqref{marg-d} define just normalization:
$$
\sum\limits_{x_{\c}}\mu_{\c}(x_\c) =  \mu_\emptyset.
$$
Together with non-negativity they guarantee boundedness (which was assumed in the general case \Section{sec:general-polyhedral}). The first order constraints~\eqref{marg-d} add marginalization constraints of the form
\begin{equation}
(\forall s \in \c,\ \forall x_s \in \X_{s} ) \ \sum\limits_{x_{\c \backslash \{s\} }} \mu_{\c}(x_\c) = \mu_s(x_s).
\end{equation}
And so on. 
By specifying larger $d$, we introduce more coupling between relaxed variables. 
\par
Note that any relaxation in the form~\eqref{local-Lambda} is local, \ie, tied to the hypergraph. We cannot add more facets (inequalities) without increasing the number of variables and the variables are defined by the fixed embedding $\delta$.
Tightening the relaxation is thus only possible by enlarging the hypergraph (adding zero interactions in~\cite{Werner-PAMI07}), which results in an exponential increase in the number of relaxed variables. An example of a non-local relaxation is the cutting plane method~\cite{Sontag-12}, which progressively adds facet-defining inequalities coupling many variables at a time. While general results of \Section{sec:max-persistency} are applicable, the local representation would not be tractable.
\par
The primal and dual LP relaxation problems for coupling $\Esubset$ are expressed as follows:
\begin{equation}\label{SLP}
\begin{array}{llr}
& \min \<f,\mu\>\tab\tab\tab \ \ \ = & \max \psi\ \ \ \ \ \ \ \\
(\forall \d\Esubset \c) & \sum\limits_{x_{\c \backslash \d }} \mu_{\c}(x_\c) = \mu_\d(x_\d),  & \varphi_{\d,\c}(x_\d) \in \Real,\\
& \mu_\emptyset = 1, & \psi \in \Real,\\
(\forall \c \in\E,\ \forall x_\c)& \mu_{\c}(x_\c) \geq  0, & \hskip-2cm f^\varphi_{\c}(x_\c) \geq \begin{cases}
0, & \hskip-2mm \c \neq \emptyset,\\
\psi, & \hskip-2mm \c = \emptyset.
\end{cases}
\end{array}
\notag
\end{equation}
Matrix $A$ corresponds to primal equality constraints.
Vector $f^\varphi = f - A\T \varphi$ is an {\em equivalent transformation}~\cite{Schlesinger-76} or {\em reparametrization}~\cite{Wainwright03nips} of $f$. Its components are expressed as
\begin{align}\label{reparam-expand}
f^\varphi_{\c}(x_\c) & = f_{\c}(x_\c) -\hskip-0.1cm \sum\clim{ \d\Esubset  \c }\hskip-0.1cm \varphi_{\d,\c}(x_\d) + \hskip-0.1cm \sum\clim{ \h\Esupset \c }\hskip-0.1cm \varphi_{\c,\h}(x_\c).
\end{align}
In particular, components $f^\varphi_\emptyset$ and $f^\varphi_{s}(x_s)$ are expressed as 
\begin{subequations}
\begin{align}
f^\varphi_\emptyset & = f_\emptyset + \sum\clim{\c\Esupset \emptyset}\varphi_{\emptyset,\c};\\
f^\varphi_s(x_s) & = f_s(x_s) - \varphi_{\emptyset,s} + \sum\clim{ \h\Esupset \{s\}} \varphi_{s,\h}(x_s).
\end{align}
\end{subequations}
For any $\varphi\in\Real^{m}$ there holds 
\begin{equation}
(\forall \mu\in\Lambda)\ \ \<f^{\varphi},\mu\> = \<f,\mu\> - \<\varphi,A\mu\> = \<f,\mu\>.
\end{equation}
Since $\Lambda \supset\M \supset \delta(\LL)$, it follows that $\f(x) = E_{f^{\varphi}}(x)$ for all $x\in\LL$. Hence $f^\varphi$ is indeed equivalent to $f$ in defining the energy function.

\par
The dual problem can be equivalently written as
\begin{equation}
\max \{ f^\varphi_\emptyset \mid (\forall \c\neq \emptyset)\ f^\varphi_{\c}(x_\c) \geq 0,\ \varphi \in\Real^m \},
\end{equation}
we therefore can speak of the dual solution as just $\varphi$.

\begin{figure}
\centering
\begin{tabular}{cc}
\includegraphics[height=0.25\linewidth]{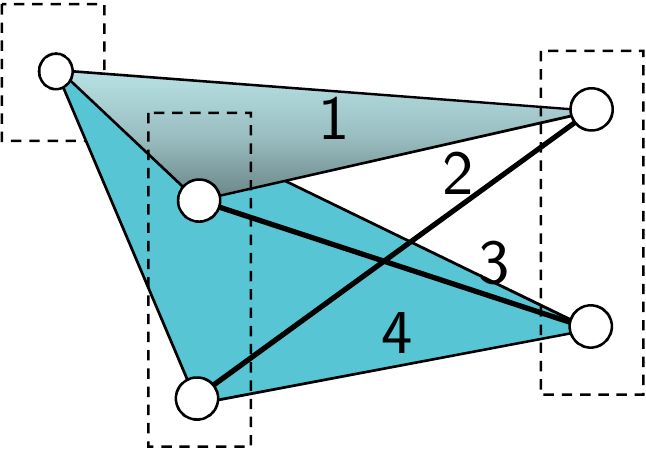}\quad &
\ \ \ \includegraphics[height=0.25\linewidth]{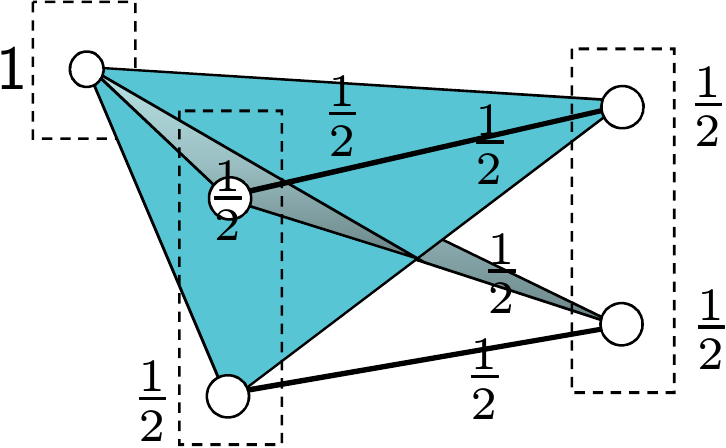}\\
(a) & (b) \\
\ \ \ \ \ \ \includegraphics[width=0.28\linewidth]{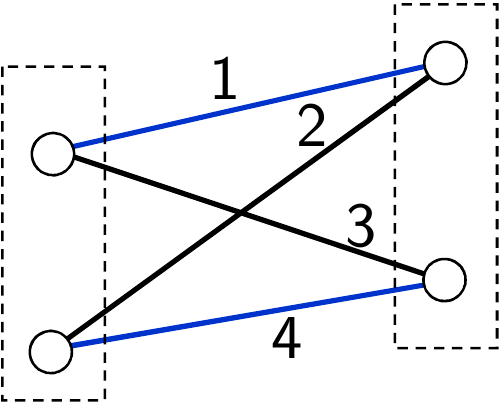} & \\
(c)
\end{tabular}
\caption{
An example when fixing a variable in BLP relaxation is not equivalent to eliminating it. (a) Energy in 3 variables (the leftmost variable has only one possible assignment). Costs $2$ and $3$ are assigned to the pairwise term (solid lines) and $1$ and $4$ to triplewise (faces), other costs are zero. (b) An optimal solution to BLP relaxation (of cost 0). BLP relaxation does not enforce marginalization between the triple and the pair. (c) The energy after elimination of the dummy variable. Now BLP relaxation is tight and can determine the optimal labeling of cost $1$.
}\label{fig:BLP}
\end{figure}

\paragraph{Complementary slackness}
Complementary slackness for (LP) reads that 
a feasible primal-dual pair $(\mu,\varphi)$ is optimal iff $(\forall \c\in\E\backslash \{\emptyset\},\ \forall x_\c)$ 
\begin{align}
\label{slack-b}
\mu_{\c}\ind{x_\c} > 0\ \Rightarrow\ \ & f^\varphi_{\c}\ind{x_\c} = 0. 
\end{align}
Because a feasible dual solution satisfies $f^\varphi_{\c} \geq 0$, the condition on the RHS of~\eqref{slack-b} implies that assignment $x_\c$ is {\em locally minimal} for $f^\varphi_\c$: $x_\c \in \argmin f^\varphi_\c(\cdot)$. 
\paragraph{Strict Complementarity}
Let $(\mu,\varphi)$ be a feasible primal-dual pair for~\eqref{LP}. This pair is called {\em strictly complementary} if
\begin{subequations}\label{sslackness}
\begin{align}
\label{sslack-b}
\mu_{\c}(x_\c) > 0\ & \Leftrightarrow\ f^\varphi_{\c}(x_\c) = 0. 
\end{align}
\end{subequations}
Clearly, a strictly complementary pair is complementary and thus it is optimal. Such a pair always exists and can be found by interior point algorithms (see \eg, \cite{Vanderbei2001}). It is known that $\mu$ is a relative interior point of the primal optimal facet and $\varphi$ is a relative interior point of the dual optimal facet.
\paragraph{Arc Consistency}
The following conditions, known as {\em arc consistency} (AC,~\eg, \cite{Werner-10}), are satisfied for strictly complementary pairs:
\begin{itemize}
\item If $f^\varphi_\c(x_\c) = 0$ then $(\forall \d \Esubset \c)$ $f^\varphi_\d(x_\d) = 0$.
\item If $f^\varphi_\d(x_\d) = 0$ then $(\forall \c \Esupset \d)$ $(\exists x'_{\c} \in \X_\c \mid x'_\d = x_\d )$ $f^\varphi_\c(x'_\c) = 0$.
\end{itemize}
These conditions say that the set of local minimizers must be consistent over overlapping hyperedges. Arc consistency is a necessary but, in general, not sufficient condition for strict complementarity.
\paragraph{BLP}
The relaxation with marginalization constraints of order $1$ is known as Basic LP relaxation (BLP)~\cite{Werner-10}. 
Note, if we do not enforce marginalization constraints of at least order $1$ there may occur integer feasible solutions to the relaxation which are not consistent, \ie, do not correspond to a global assignment. 
Out of all local relaxations BLP is the least constrained useful one. It is remarkable that it is tight for all tractable languages~\cite{Thapper-12,Thapper-13,Kolmogorov-12-LP-power}. However, for certain purposes BLP is not sufficient, as can be illustrated with pseudo-Boolean functions. Suppose we would like to express a pseudo-Boolean function of 3 variables as a cubic polynomial. We know it can be expressed in this form, however, such a desired equivalent transformation of the problem appears to be not equivalent for BLP and hence not equivalent for the maximum persistency problem. Another difficulty is that fixing a variable to its optimal value is not the same as eliminating this variable. Example in \Figure{fig:BLP} illustrates that eliminating a persistent variable tightens the relaxation.
It follows that under BLP relaxation we won't be able to compare theoretically neither to quadratization techniques (as they perform general equivalent transformations) nor to generalized roof duality~\cite{KahlS11}, which incrementally eliminates persistent variables.
%
\paragraph{FLP}
The relaxation with all marginalization constraints present will be refereed to as Full local LP relaxation (FLP). For every hyperedge all its subsets are assumed to be contained in $\E$ and all constraints of the form~\eqref{marg-d} with $d$ equal to the order of the problem are included.
In case of pairwise model, individual nodes are the only proper subsets of edges and hence BLP and FLP are the same. In the pseudo-Boolean case, FLP matches the relaxation of~\citet{SheraliA90} as discussed in \Section{sec:adams-proofs}.
%
\subsection{Maximum Persistency with Local Relaxations}\label{sec:L1-s}
In this section we summarize how the general construction and formulation of~\eqref{L1} simplifies for local relaxations. First, the constraint $P_\zeta(\Lambda) \subset \Lambda$ holds automatically and needs not be enforced. It is shown in two steps: first we consider the linear extension $[p]$ of any node-wise mapping $p$ and then the linearized mapping $P_\zeta$, $\zeta\in\Zeta$.
\begin{restatable}{lemma}{ppreservesLambda}\label{[p] preserves Lambda}
 Node-wise mapping $[p]$ preserves the local polytope $\Lambda$.
\end{restatable}
\begin{restatable}{lemma}{Lisclosed}\label{L is closed}
Mapping $P_\zeta$ for $\zeta\in\Zeta$ preserves the local polytope: $P_\zeta(\Lambda) \subset \Lambda$.
\end{restatable}
In short, $P_\zeta$ satisfies all the equality constraints satisfied by $[p]$ and has all components non-negative for $\zeta\in\Zeta$. 
As a consequence of~\Lemma{L is closed}, the constraint of polytope preservation~\eqref{IL1-closed} in the maximum persistency problem~\eqref{L1} can be dropped. We can write \maxwi as 
\begin{align}\label{L1-simplex}
& \min\csub{\zeta,\varphi} \sum_{s,i}\zeta_{s,i} \\ 
\notag
& (I-P\T_\zeta)f - A\T \varphi \geq 0;\ \varphi\in\Real^m;\\
\notag
& \zeta\in\Zeta.
\end{align}

Further properties of improving mappings for local relaxations are as follows.
Conditions that are necessary for $[p]\in\WI$ in the general case (\Theorem{T:Char}) become necessary and sufficient for local relaxations and can be now summarized together with the dual representation \Theorem{S:WI-dual}: 
\begin{restatable}[Characterizations]{theorem}{Tcharacterization}\label{T:characterization}
For a local relaxation $\Lambda$ all of the following are equivalent:
\begin{enumerate}
\item \label{char-a} $P \in \WI$;
\item \label{char-b} $(\exists \varphi \in\Real^m) \ f^\varphi-P\T f \geq 0$;
\item \label{char-c} $(\exists \varphi \in\Real^m) \ (I-P\T) f^\varphi \geq 0$;
\item \label{char-d} $\inf \{ \<f-P\T f,\mu\> \mid \mu\in\Real^\I_+, \ \ A(I-P)\mu = 0\} = 0$.
\end{enumerate}
\end{restatable}
We have transitions from a global property $(\forall \mu\in\Lambda)\ \ \<(I-P\T)f,\mu\> \geq 0$ (a) to component-wise local inequalities~\ref{char-b} and~\ref{char-c}. Inequalities~\ref{char-c} offer an equivalent reparametrization $f^\varphi$ in which mapping $p$ improves every component independently:
\begin{align}
(\forall \c\subset\E,\ \forall x_\c\in\X_\c)\ f^\varphi_{\c}(p_\c(x_\c)) \leq f^\varphi_{\c}(x_\c).
\end{align}
This is a fairly simple condition similar in spirit to the idea of equivalent transformations by~\citet{Schlesinger-76} (find an equivalent $f^\varphi$ such that the global minimum may be recovered from independent component-wise minima). 
Condition~\ref{char-d} is a primal reformulation which has fewer equality constraints than the verification LP and hence is simpler.

Some properties expressed for all hyperedges $\c\in\E$ can be simplified if we assume at least the BLP relaxation. In \Statement{T:SI-components} it is sufficient that only unary components satisfy $(\forall s\in\V)\ \ \O_s = p_s(\X_s)$. For other components the constraint is implied by marginalization. For the same reason, in the perturbation~\eqref{perturbation} it is sufficient to have only unary components $f_s(y_s)$ increased by $\epsilon$ for all $s$ and leave higher-order terms intact. 
%
\par
%
Lastly, there are following NP-hardness results with BLP relaxation:
\begin{restatable}{theorem}{TmaxwiNPhard}\label{T: maxwi NP hard}
Problem \maxwi over the $\P^1$ class of maps and the BLP relaxation is solvable in polynomial time for the quadratic pseudo-Boolean case and otherwise (when the problem is multilabel or higher order) it is NP-hard. 
\end{restatable}
\begin{restatable}{theorem}{TmaxsiNPhard}\label{T: maxsi NP hard}
Problem \maxsi with 4 or more labels over the class of maps $\P^2= \bigcup _{y\in \X}\P^{2,y}$ and BLP relaxation is NP-hard. 
\end{restatable}
We see that the difference between weak and strong persistency leads to different complexity classes for the maximum persistency problem. The question of complexity of \maxsi with 3 labels is not resolved.
%
%
%

\section{Comparison Theorems}\label{sec:comparison}
This section is devoted to theoretical comparison between different persistency techniques. The firs result is the following:
\begin{theorem}\label{T:poly-inclusion}
Let $\Lambda \subset \Lambda'$ and $P$ be a $\Lambda'$-improving mapping for $f$. Then $P$ is $\Lambda$-improving for $f$.
\end{theorem}
\begin{proof}
The claim follows from Definition~\ref{def:relaxed-improving} and nesting of polytopes $\Lambda \subset \Lambda'$.
\end{proof}
We therefore have a natural hierarchy: if we can identify some variables as persistent by the proposed sufficient condition with relaxation $\Lambda'$ then for any tighter relaxation $\Lambda \subset \Lambda'$ we are guaranteed to find at least the same persistent variables.
Other nesting results under different reformulations of the problem are obtained in \Section{sec:quadratization}, \Section{sec:hocr-proof}.
\par
\autoref{table:comparisons} gives an overview of the obtained comparisons to other methods. The first comparison column establishes that all listed methods correspond to a relaxed-improving mapping under standard relaxations (recall that in the pairwise case FLP = BLP). For cases when \Algorithm{Alg1} is optimal, as indicated in \autoref{table:P-cases}, it is guaranteed to find the same or larger set of persistent labels than any other method. This fills the second comparison column in \autoref{table:comparisons}.
In the remainder of this section we give a more detailed overview of different methods and comparison results. 
\par
\begin{table}[t]
\centering
\setlength{\tabcolsep}{3pt}
\renewcommand{\arraystretch}{1.2}
\scriptsize
\resizebox{\linewidth}{!}{%
\begin{tabular}{|p{4.5ex}|c|c|c|}
\cline{2-4}
\multicolumn{1}{l|}{}&\multicolumn{3}{r|}{Dominated by \Algorithm{Alg1}}\\
\cline{2-3}
\multicolumn{1}{l|}{}&\multicolumn{2}{r|}{Corresponds to a BLP/FLP-improving mapping} & \\
\cline{1-2}
\parbox[t]{2mm}{\multirow{5}{*}{\rotatebox[origin=c]{90}{\parbox{10ex}{pairwise\newline multilabel}}}}
& Simple DEE~\cite{Goldstein-94-dee} & \checkmark & - \\
& MQPBO~\cite{kohli:icml08} & \checkmark & - \\
& Kovtun's one-against-all~\cite{Kovtun03} & \checkmark & \checkmark \\
& Kovtun's iterative~\cite{Kovtun-10} & \checkmark & - \\
& \citet{Swoboda-14}{\small **} & \checkmark & \checkmark \\
\cdashline{2-2}[1pt/2pt]
\parbox[t]{2mm}{\multirow{4}{*}{\rotatebox[origin=c]{90}{\parbox{17.5ex}{higher order\newline pseudo-Boolean}}}} 
& Roof dual / QPBO~\cite{Nemhauser-75,Hammer-84-roof-duality,BorosHammer02,Kolmogorov-Rother-07-QBPO-pami} & \checkmark & ={\small *} \\
\cdashline{2-2}[1pt/2pt]
& Reductions: HOCR~\cite{Ishikawa-11}, \citet{Fix-11} & FLP & \checkmark{\small *} \\
& Bisubmodular relaxations~\cite{Kolmogorov10-bisub}{\small ***} & BLP & \checkmark{\small *} \\
& Generalized Roof Dualilty~\cite{KahlS11} & FLP & \checkmark{\small *} \\
& Persistency by \citet{Adams:1998} & FLP & \checkmark{\small *} \\[5pt]
\hline
\end{tabular}}
\vskip5pt
\caption{Summary of theoretical comparisons. {\small *}Result holds for strong persistency variants and resp. strict version of \Algorithm{Alg1}. {\small **}\cite{Swoboda-14} is higher order but the comparison proof is for the pairwise case. {\small ***}Result holds for sum of bisubmodular functions over the same hypergraph as the BLP relaxation.
}\label{table:comparisons}
\end{table}
\subsection{DEE}
\begin{figure}[!t]
\centering
\includegraphics[width=0.8\linewidth]{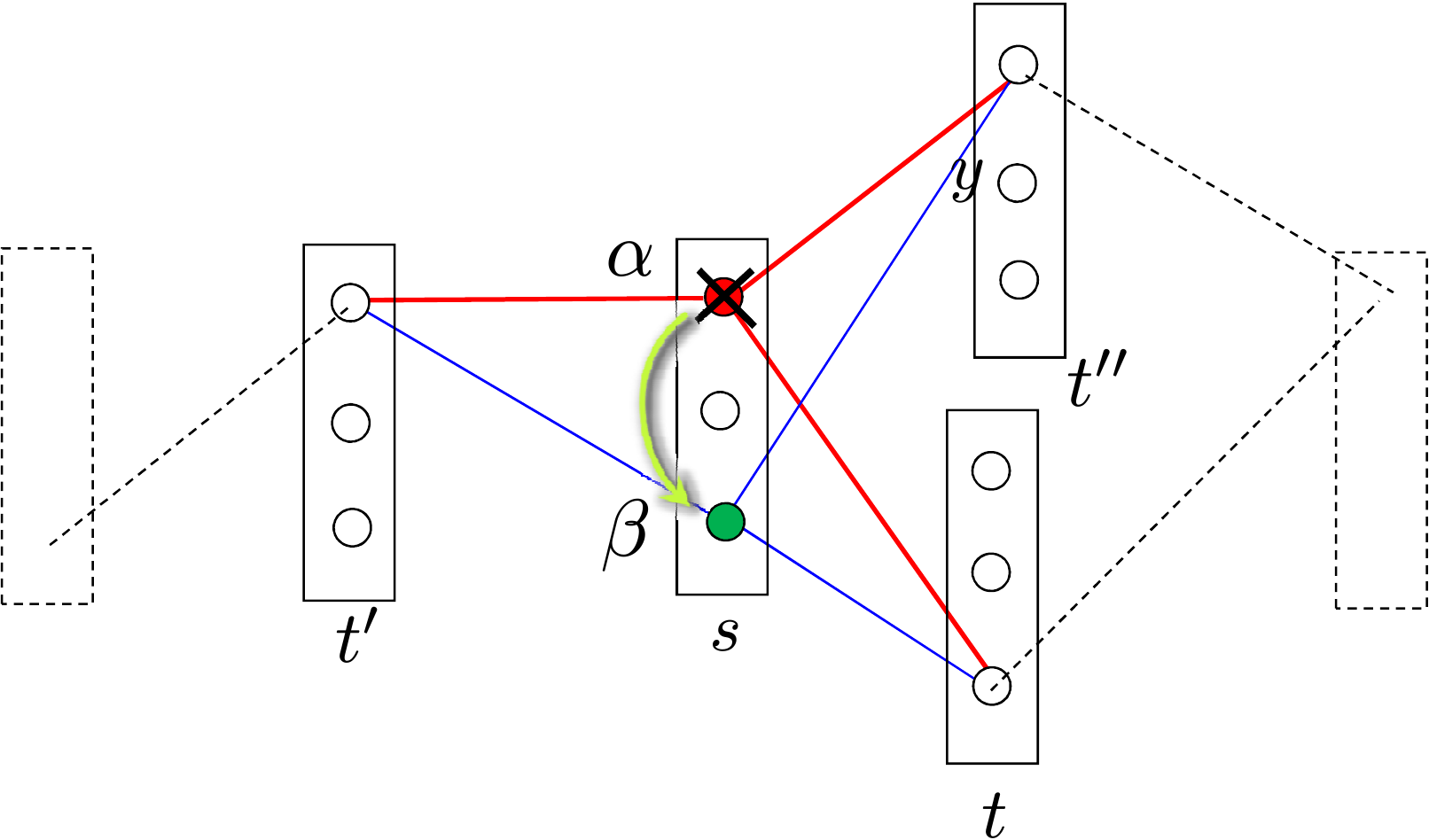}
\caption{Improving mapping corresponding to an individual DEE condition. The full DEE method iterates over all nodes and labels and composes the found improving maps.}
\label{fig:dee}
\end{figure}

We will consider Goldstein's {\em simple} DEE~\cite{Goldstein-94-dee} (which is stronger than original DEE by~\citet{Desmet-92-dee}) in the pairwise setting. For every node $s$ this method considers its neighbors in the graph, $\N(s)$, and for a pair of labels $\alpha,\beta\in\LL_s$ verifies the condition
\begin{align}\label{DEE-simple}
& (\forall x\in \LL_{\N(s)}) \\
\notag
& f_s(\alpha)-f_s(\beta)
+\sum_{t\in\N(s)} [f_{\{s,t\}}(\alpha,x_t)-f_{\{s,t\}}(\beta,x_t)] \geq 0,
\end{align}
illustrated in \Figure{fig:dee}. If the condition is satisfied it means that a (weakly) improving switch from $\alpha$ to $\beta$ exists for an arbitrary labeling $x$. In this case $(s,\alpha)$ can be eliminated while preserving at least one optimal assignment. 
\par
It is trivial to construct an improving mapping for this case. We let $p_s(\alpha) = \beta$, $p_s(i) = i$ for $i\neq \alpha$; and $p_t(i)=i$ for all $t\neq s$. The non-zero terms of the problem $g = (I-P\T)f$ form a tree with root node $s$ and other nodes $t\in \N(s)$ being leaves. It is known that in this case the FLP relaxation is tight and therefore $p$ is FLP-improving. 
Similarly, the strict inequality in~\eqref{DEE-simple} implies $[p] \in \SI$.

\par
\subsection{QPBO}
Let $\LL_s = \Bool$.
The {\em weak persistency} theorem~\cite{Nemhauser-75,Hammer-84-roof-duality} can be formulated as follows.
Let $\mu\in \argmin_{\mu\in\Lambda}\<f,\mu\>$.
Let $\O_s = \{i\in\Bool \mid \mu_s(i) > 0\}$.
Then
\begin{equation}\label{qpbo-weak}
(\exists x\in \argmin_x \f(x))\ (\forall s\in\V) \ x_s\in \O_s.
\end{equation}
In the case $|\O_s|=1$ vector $\mu_{s}$ is necessarily integer and the theorem states that there is an optimal solution $x$ to the discrete problem which is consistent with the integer part of the relaxed solution $\mu$. The largest weakly persistent assignment is obtained in the case $\mu$ is the solution with the maximum number of integer components.
%
\par
The {\em strong persistency} theorem~\cite{Nemhauser-75,Hammer-84-roof-duality} can be formulated as follows.
\begin{theorem}[\cite{Nemhauser-75,Hammer-84-roof-duality}]
Let $(\mu,\varphi)$ be a strictly complementary primal-dual pair. 
Let $\O_s$ be defied as above. Then 
\begin{equation}\label{qpbo-strong}
(\forall x\in \argmin_x \f(x))\ (\forall s\in\V) \ x_s\in \O_s.
\end{equation}
\end{theorem}
The difference to~\eqref{qpbo-weak} is in the quantifier $\forall$ \vs. $\exists$. Note, a strictly complementary solution has the minimum number of integer components.
%
%

\begin{theorem}[\cite{shekhovtsov-14-TR}]\label{T:QPBO-W}
Weak (resp. strong) persistency by QPBO corresponds to an FLP-improving (resp. strict FLP-impriving) mapping.
\end{theorem}
The mapping is defined by $p_s(i) = 0$ if $O_s=\{0\}$, $p_s(i) = 1$ if $O_s=\{1\}$ and $p_s(i) = i$ otherwise. 
The idea of the proof is to show that the dual optimal solution $\varphi$ provides the reparametrization in which the mapping improves every component independently, \ie, satisfies the inequalities of the characterization \Theorem{T:characterization}(c).
\par
It follows from the theorem that solution by \Algorithm{Alg1} with perturbation coincides with the strong QPBO persistency. 
\subsection{MQPBO}
\begin{figure}
\centering
\includegraphics[width=0.9\linewidth]{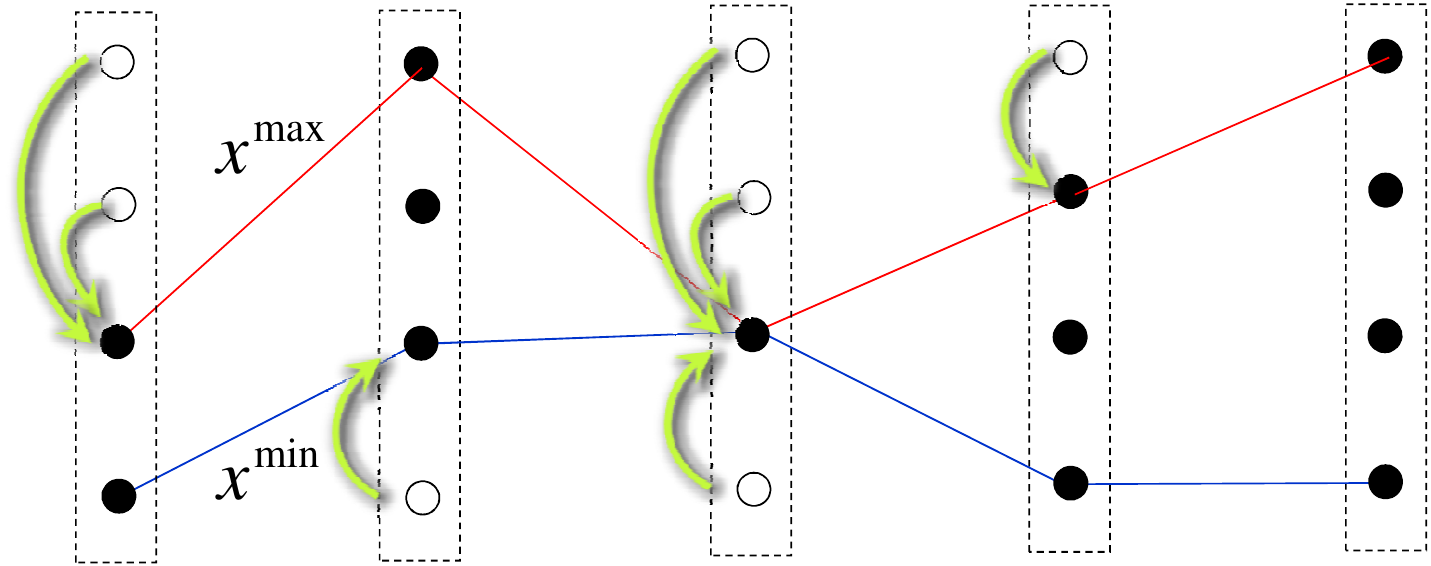}
\caption{Improving mapping found by the MQPBO method: all labels above $x^{\rm max}$ are mapped to $x^{\rm max}$ and all lables below $x^{\rm min}$ to $x^{\rm min}$. Labelings $x^{\rm max}$ and $x^{\rm min}$ are determined by the method from the underlying roof dual relaxation. }
\label{fig:MQPBO}
\end{figure}

The MQPBO method~\cite{kohli:icml08} extends partial optimality properties of QPBO to multilabel problems via the reduction of the problem to $0$-$1$ variables. The reduction, known as "$K$ to $2$" transform~\cite{DSchlesinger-K2} ($K = |\X_s|$), depends on the linear ordering of labels in $\X_s$. The method outputs two labelings $x^{\rm min}$ and $x^{\rm max}$ with the guarantee that there exists an optimal labeling $x$ that satisfies $x_s \in [x^{\rm min}_s,\, x^{\rm max}_s]$. The corresponding improving mapping has the form $p \colon x \mapsto (x \vee x^{\rm min}) \wedge x^{\rm max}$, where $\vee$ and $\wedge$ are component-wise minimum and maximum, resp. in a given ordering of labels. The mapping is illustrated in~\Figure{fig:MQPBO}.
Because the reduction is component-wise and component-wise inequalities hold for QPBO it follows that the component-wise conditions of \autoref{T:characterization}\ref{char-c} hold for $p$ (proof in~\cite{shekhovtsov-14-TR}).
For $(x^{\rm min}, x^{\rm max})$ obtained from weak (resp. strong) persistency by QPBO there holds $[p]\in \WI$ (resp. $[p]\in \SI$). Since the class of mappings of the form $x \mapsto (x \vee x^{\rm min}) \wedge x^{\rm max}$ is not among the cases for which \Algorithm{Alg1} is optimal, the question of tractability of \maxsi for this class remains open.
%

\par
\subsection{Auxiliary Submodular Problems by Kovtun}\label{sec:kovtun}
\begin{figure}
\centering
\begin{tabular}{cc}
(a)\ \ \ \ \ & \begin{tabular}{c}\includegraphics[width=0.9\linewidth]{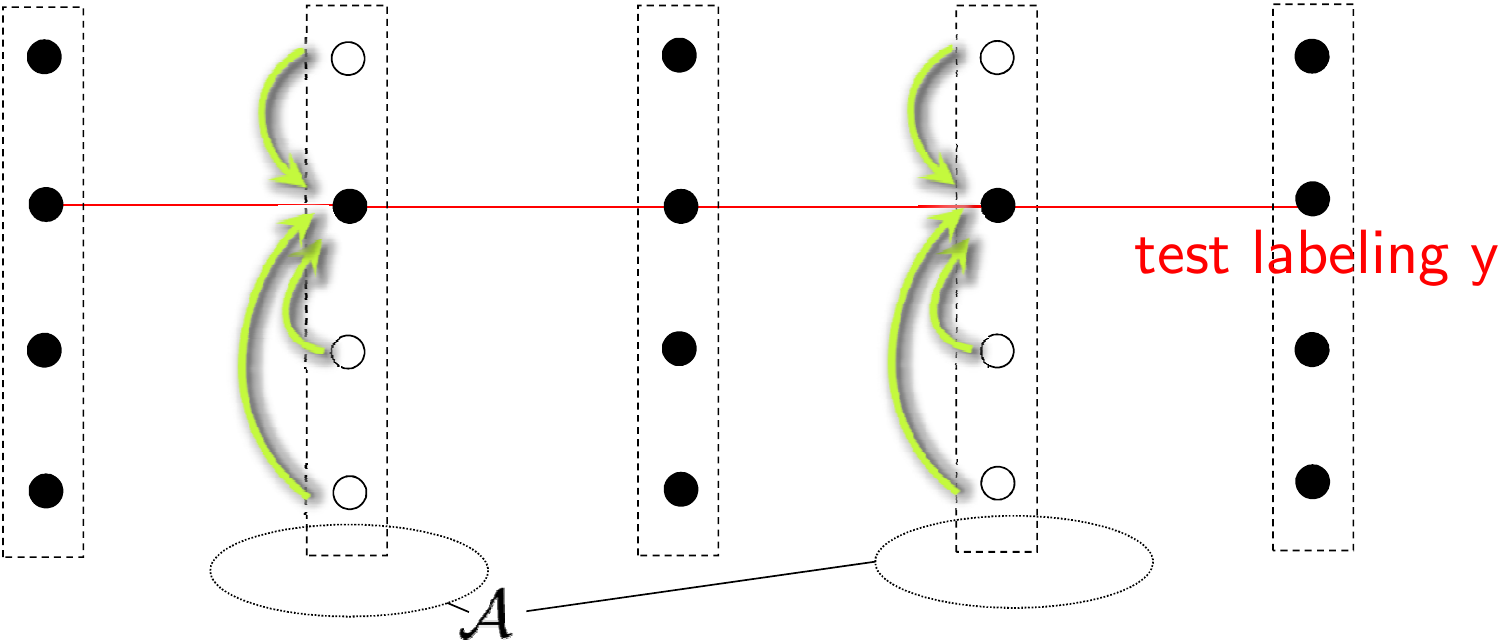}\end{tabular} \\
(b)\ \ \ \ \ & \begin{tabular}{c}\includegraphics[width=0.9\linewidth]{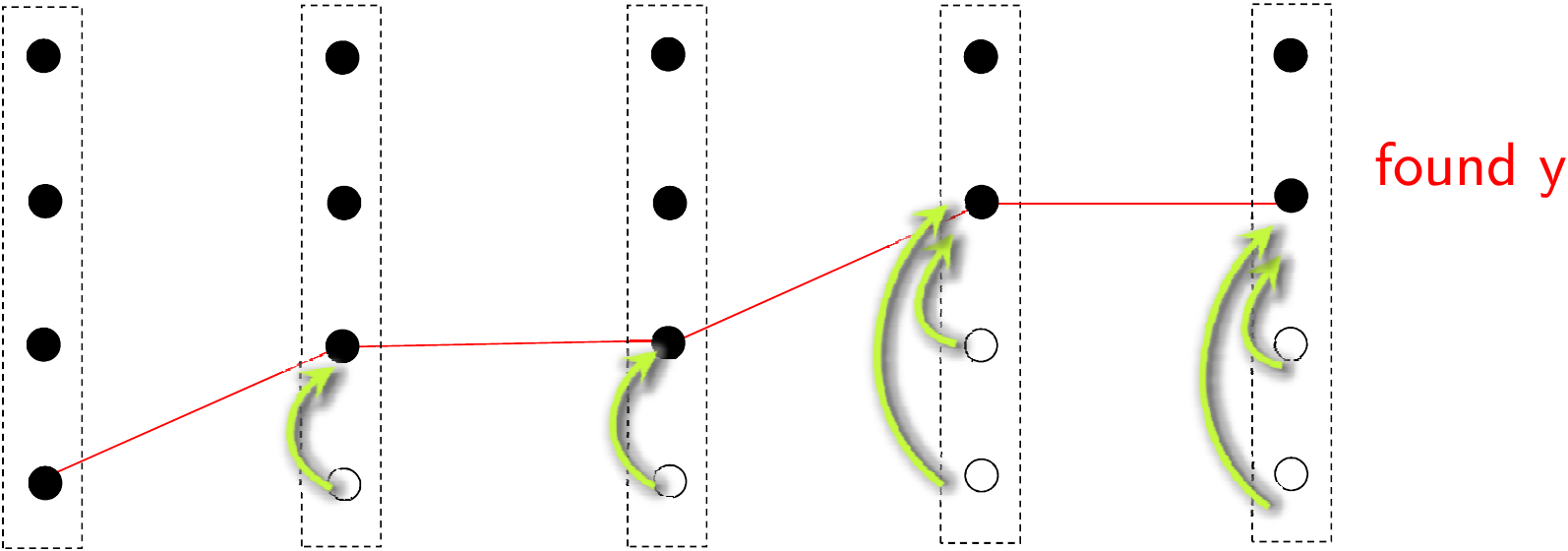}\end{tabular}
\end{tabular}
\caption{Improving mappings in Kovtun's methods. (a) One-against-all method for fixed test labeling $y$. The method determines a subset $\A$ of vertices for which the optimal labeling is $y$. 
(b) Iterative method~\cite{Kovtun-10} in which labeling $y$ is found incrementally and with respect to a predefined ordering of labels.}
\label{fig:Kovtun}
\end{figure}

There were several methods proposed~\cite{Kovtun03,Kovtun-10} which differ in detail. All methods construct an auxiliary submodular energy $\g$. A minimizer $y$ of $\g$ has the property that $\g(x \vee y) \leq \g(x)$, implied by submodularity.  
It follows that mapping $p \colon x \mapsto x \vee y$ is improving for $g$. \Figure{fig:Kovtun} illustrates such mappings found by two of the methods in~\cite{Kovtun-10}. 
In case (a) the test labeling $y$ must be the highest (the maximum) in the selected order of label. It follows that the mapping $p$ is essentially of the form $x \mapsto x [\A{\leftarrow}y]$, \ie, from the class $\P^{1,y}$.
The construction of the auxiliary function ensures that improvement in $f$ is at least as big as improvement in $g$ 
and so $p$ is improving for $f$. 
\begin{theorem}[\cite{shekhovtsov-14-TR}]Persistency by any method~\cite{Kovtun03,Kovtun-10} corresponds to an FLP-improving mapping.
\end{theorem}
Since in the case (a) the mapping is in the class $\P^{1}$, we know that the strict version of the method is dominated by \Algorithm{Alg1}. 
In the case (b), the class of maps is a subset of maps considered in MQPBO and tractability of \maxsi problem is also open.
\par
Computationally, methods of~\citet{Kovtun-10} have an advantage as they rely on the minimization of a pairwise submodular function. In the case of the Potts model, the method~\cite{Kovtun03} for all ``flat'' test labelings $y_s = \alpha$  for $\alpha=1\dots K$, ($K = |\X_s|$), 
can be efficiently performed using $\log(K)$ maximum flow computations~\cite{Gridchyn-13}. It is very practical in some vision problems (\eg results~\cite{Kovtun03,Alahari-08,Gridchyn-13}), where unary costs are determining. At the same time experiments on difficult random problems in \Section{sec:experiments} reveal very poor performance of this method.

%
\subsection{Iterative Pruning by Swoboda \etalb}\label{sec:swoboda}

\begin{figure}
\centering
\includegraphics[width=0.8\linewidth]{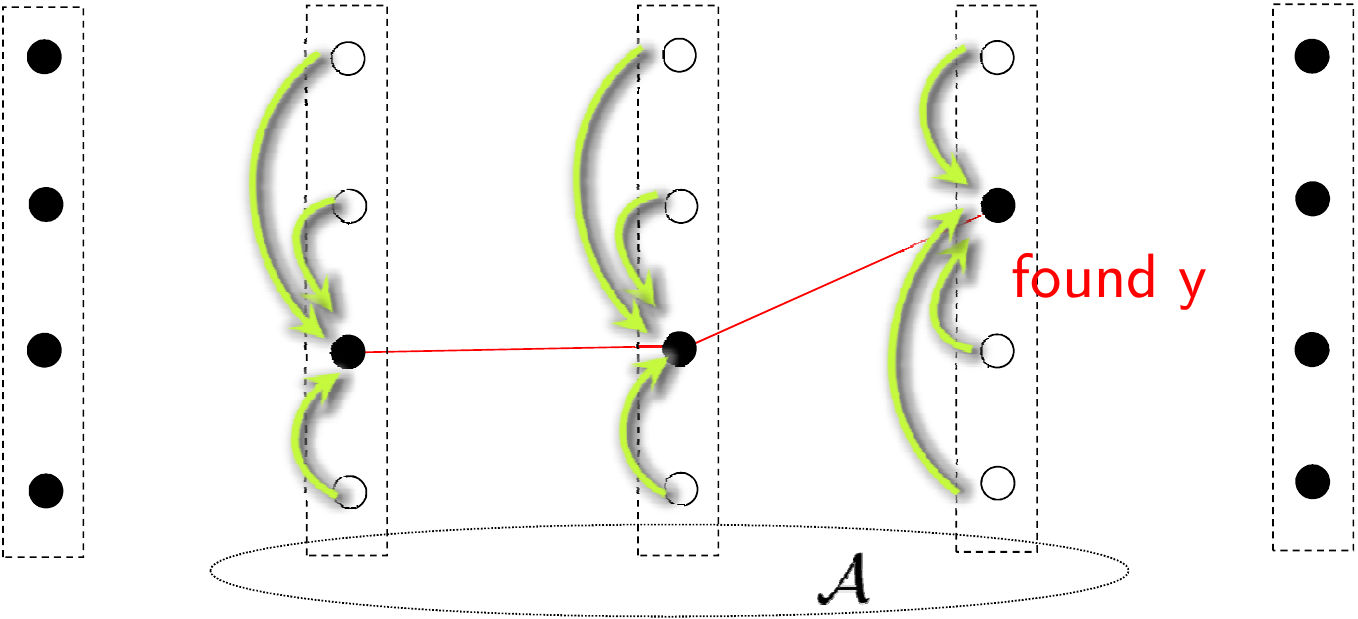}
\caption{Improving mapping in the method of \citet{Swoboda-14}. The method finds the labeling $y$ and a subset $\A$. Outside $\A$ the mapping is identity.}
\label{fig:Swoboda}
\end{figure}

The iterative Pruning method was first proposed~\cite{Swoboda-13} for the Potts model and then extended to general pairwise and higher order energies~\cite{Swoboda-14}. 
The method can be interpreted as finding an improving mapping in the class $\P^1$ (\Figure{fig:Swoboda}).
\begin{restatable}{theorem}{Tswobodacomp}\label{T:swoboda-comp}
Persistency by method~\cite{{Swoboda-14}} in the pairwise multilabel case corresponds to an FLP-improving mapping.
\end{restatable}
In fact the optimal value of $y$ is determined in~\cite{Swoboda-14} by the initial relaxation, similarly to how it is determined in \Algorithm{Alg1}. 
Therefore, \Algorithm{Alg1} without perturbation identifies the same or better weak persistency. 
\Algorithm{Alg1} with perturbation identifies the same or larger set $\A_{\rm strong}$ as theoretically guaranteed in~\cite{Swoboda-14}. 

\subsection{Quadratization Techniques}\label{sec:quadratization}
We now turn to the higher order pseudo-Boolean case. There is a number of different reductions proposed~\cite{Ishikawa-11,Fix-11,Boros-12} which represent the initial function of $0$-$1$ variables as a minimum of a quadratic function over auxiliary $0$-$1$ variables. 
Persistency is obtained then by applying the QPBO method to the reduced problem. 
Since QPBO solves the FLP relaxation, our goal is to compare local relaxations as well as relaxed-improving maps before and after the reduction. Fortunately, full reductions~\cite{Ishikawa-11,Fix-11} are defined by chaining certain elementary reductions applied to separate cliques or groups of cliques. 
We define a sufficient set of atomic reductions with the following property: the maximum persistent subset by an FLP-improving mapping for the reduced problem is not larger than that one for the initial problem. Chaining these atomic reductions we obtain the following comparisons.
\begin{restatable}{theorem}{THOCR}\label{T:HOCR}
Persistency by Higher Order Clique Reduction (HOCR) of \citet{Ishikawa-11} corresponds to an FLP-improving mapping.
\end{restatable}
\begin{restatable}{theorem}{TFix}\label{T:Fix}
Persistency by method of~\citet{Fix-11} corresponds to an FLP-improving mapping.
\end{restatable}
\citet{Ishikawa-11} proposed a family of elementary reductions (called $\gamma$-flipping) and posed the problem of what sequence of reductions gives in a certain sense the best overall reduction. This is a difficult combinatorial problem. While we do not address it directly, it follows that FLP maximum persistency by \Algorithm{Alg1} dominates persistencies that can be obtained by any reduction from the family and hence also the best one. 
%

%
\subsection{Bisubmodular Relaxations}
Submodular/bisubmodular relaxations were introduced by~\citet{Kolmogorov10-bisub} as a natural generalization of the roof duality approach to higher order pseudo-Boolean functions. Kolmogorov showed that all totally half-integral relaxations are bisubmodular relaxations and vice versa.
Similar to roof duality, (bi)submodular relaxations have a global persistency property. However, to a given function many different (bi)submodular relaxations can be constructed. There are two challenges in this approach. One is that the class of all (bi)submodular relaxations is very large and it is not tractable to parametrize it. The other challenge is to answer the question of which relaxation provides the largest persistent assignment.
\par 
\citet{KahlS11} build upon graph-cut reducible submodular relaxations. 
They propose that the relaxation which corresponds to the best lower bound on the energy is the optimal one. Their algorithm solves a series of linear programs to build the tightest graph-cut reducible submodular relaxation. However, not all submodular relaxations are graph-cut reducible (it is a hard problem to determine which ones actually are~\cite{Zivny-08-binary} with the exception of cubic functions). Moreover, it is not clear whether the relaxation that gives the best lower bound is also the best one \wrt the size of the persistent assignment.
\par
We consider a more general case when the relaxation is a sum of bisubmodular functions (SoB) over the same hypergraph as $f$. This class includes all graph-cut reducible submodular relaxations. Exploiting the property that SoB function can be minimized exactly by BLP relaxation~\cite{Thapper-12} and properties~\cite{Kolmogorov10-bisub}, we obtain the following theorem. 
\begin{restatable}{theorem}{TbisubmodularBLP}\label{T:bisubmodularBLP}
Persistency by SoB relaxation~\cite{Kolmogorov10-bisub} corresponds to a BLP-improving mapping. 
\end{restatable}

The work of~\citet{Lu-Williams-1987} is a special case of SoB relaxation, it follows that their result corresponds to a BLP-improving mapping as well.
%
%
\subsection{Generalized Roof Duality}
%
As discussed above, the method of~\citet{KahlS11} finds persistencies by SoB relaxation and all such relaxations are dominated by BLP-improving maps. There is however a catch. The method reduces the problem progressively by finding in each iteration a BLP-improving map. 
While \Lemma{necessary-LI} guarantees that fixing variables to their persistent values does not tighten the BLP-relaxation, eliminating persistent variables actually does (as explained in \Figure{fig:BLP}). It follows that their method is not in general dominated by a single BLP-improving map.
%
On the other hand, we can easily claim domination by a single FLP-improving map (as it is stronger than BLP and is not tightened by the elimination of persistent variables), which is also confirmed experimentally.
\par
While avoiding the difficult question~\cite{Kolmogorov10-bisub,KahlS11} of how to find the best SoS or SoB relaxation, we give an answer to how to find same or larger strong persistent assignment. In the case of 3rd order energies (quartic terms), \citet{Kolmogorov10-bisub} gives an example where there is a tight bisubmodular relaxation but no tight submodular relaxation. It follows that in this case~\Algorithm{Alg1} can determine strictly larger strong persistent assignment than~\cite{KahlS11}. We give experimental confirmation of larger persistent set for both cubic and quartic problems. The proposed two-phase algorithm is seen more computationally attractive than the series of LPs of~\citet{KahlS11}.

\par
\citet{Windheuser-et-al-eccv12} extended generalized roof duality~\cite{Kolmogorov10-bisub,KahlS11} to multilabel case. For pairwise models they showed equivalence with MQPBO. For higher order models 
the approach can be seen as a combination of $K$ to $2$ transform~\cite{DSchlesinger-K2} and application of submodular relaxation~\cite{Kolmogorov10-bisub,KahlS11}. As we have given comparison with (bi)submodular relaxations and $K$ to $2$ transform is component-wise, it should follow that the sufficient condition of~\cite{Windheuser-et-al-eccv12} corresponds to an FLP-improving mapping. 
%
%
%
%

%
%
\subsection{Persistency in 0-1 Polynomial Programming by Adams \etalb}
\citet{Adams:1998} proved a persistency result for the $0${-}$1$ polynomial programming problem. 
The result is based on the relaxation of~\citet{SheraliA90}, which can be identified with the FLP relaxation. 
They proposed a sufficient condition on the dual multipliers in the relaxation which provides a persistency guarantee. The sufficient condition is a linear feasibility program that can be verified for a given partial assignment, similar in spirit to our verification LP. No method to find a persistent partial assignment except for the case when the integer part of the optimal relaxed solution turns out to be persistent is proposed. We show the following.
\begin{restatable}{theorem}{TAdamscondition}\label{T:Adams-condition}
Persistency by the sufficient condition of~\citet[Lemma 3.2]{Adams:1998} corresponds to an FLP-improving mapping.
\end{restatable}
In fact their sufficient conditions splits the problem into two overlapping parts: one part, where an optimal assignment is unknown (call it the inner problem) and the second part, containing all the assigned variables and the coupled unassigned variables (call it outer problem).
It can be seen that the sufficient condition guarantees that any choice of unassigned variables together with the assigned ones delivers an optimal solution to the outer problem. Thus what happens in the inner problem can be efficiently ignored. Any  feasible solution of the inner problem is optimal to the outer one. 
%
%
%

\section{Experiments}\label{sec:experiments}
We propose two families of experiments: for pairwise multilabel energies and higher-order pseudo-Boolean energies. We first discuss linear program~\eqref{L1} for FLP relaxation in these two cases.
\subsection{Details of L1 Program}

Explicit form of~\eqref{L1} for the pairwise multilabel case is given in~\cite{shekhovtsov-14-TR}.
It is expressed with variables $\xi$ which are related to $\zeta$ by 
$\xi_{s,i} = 1-\zeta_{s,i}$; $\xi_{st,ij} = 1-\zeta_{s,i}-\zeta_{t,j}+\zeta_{st,ij}$, 
so that $\xi_{st,ij}$ is the linearization of $\xi_{s,i}\xi_{t,j} = (1-\zeta_{s,i})(1-\zeta_{t,j})$. This parametrization is more convenient in the pairwise case. Its drawback is a more complex representation of $P_\zeta^2$ (which is however not needed except for the proof).
\par
Let us consider now the pseudo-Boolean case. Without loss of generality we assume that $y = 0$ (otherwise variable values can be flipped). Let $f$ be given in the form of a multilinear polynomial: $f_\c(x_\c) = \eta_\c \prod_{s\in\c} \x_s$. We let $\zeta_{\c}$ denote $\zeta_{\c, 1_\c}$. The expression $P_\zeta f$ simplifies as
\begin{align}
& (P_\zeta f)_\c (x'_\c) = \sum_{x_\c\in\X_\c}\sum_{\d\subset \c}c_{\c,\d}(x_\c,x'_\c)\zeta_{\d,x'_{\d}} f_\c(x_\c)\\
\notag
& = \sum_{\d\subset \c}\prod_{s\in\d}(\leftbb x'_s{=}1\rightbb -\leftbb y_s{=}1\rightbb)\prod_{s\in \c\backslash \d}\leftbb y_s{=}1\rightbb \zeta_{\d,x'_{\d}} f_\c(1_\c)\\
\notag
& = \leftbb x'_\c {=} 1_\c\rightbb \zeta_{\c} \eta_\c. 
\end{align}

Components of $(I-P_\zeta\T)f$ simplify as 
\begin{align}
\notag
&f_\c(x'_\c) - \leftbb x'_\c {=} 1_\c\rightbb \zeta_{\c} \eta_\c 
=\leftbb x'_\c {=} 1_\c\rightbb (1 - \zeta_{\c})\eta_\c.
\end{align}
The set $\tilde \X_s = \X_s\backslash \{y_s\}$ equals simply $\{1\}$ and thus polytope $\Zeta$ simplifies as
\begin{align}\label{Zeta-set-binary}
& \zeta_\emptyset = 1,\\
\notag
&(\forall \c\in\E,\ \forall \d\subset\c)\ \ \sum\limits_{\d\subset\c\backslash \b}(-1)^{|\d|}\zeta_{\d} \geq 0.
\end{align}
Problem~\eqref{L1} can be written as
\begin{align}
& \min_{\zeta,\varphi} \sum_{s\in\V} \zeta_s \\
\notag
&\begin{array}[b]{lr}
(\forall \c\in\E) \ & \zeta_{\c} \eta_\c - (A\T \varphi)_\c(1_\c) \geq 0;\\
(\forall \c\in\E)(\forall x'_\c\neq 1_\c) \ & - (A\T \varphi)_\c(x'_\c) \geq 0;\\
& - (A\T \varphi)_\emptyset \geq 0;\\
& \zeta \in \Zeta.
\end{array}
\end{align}
Implementation in matlab is available at \url{http://www.icg.tugraz.at/Members/shekhovtsov/persistency} for research purposes.
%
\subsection{Evaluation}\label{sec:evaluation}
\begin{figure*}[!t]
\ifblacknwhite
\includegraphics[width=0.5\linewidth]{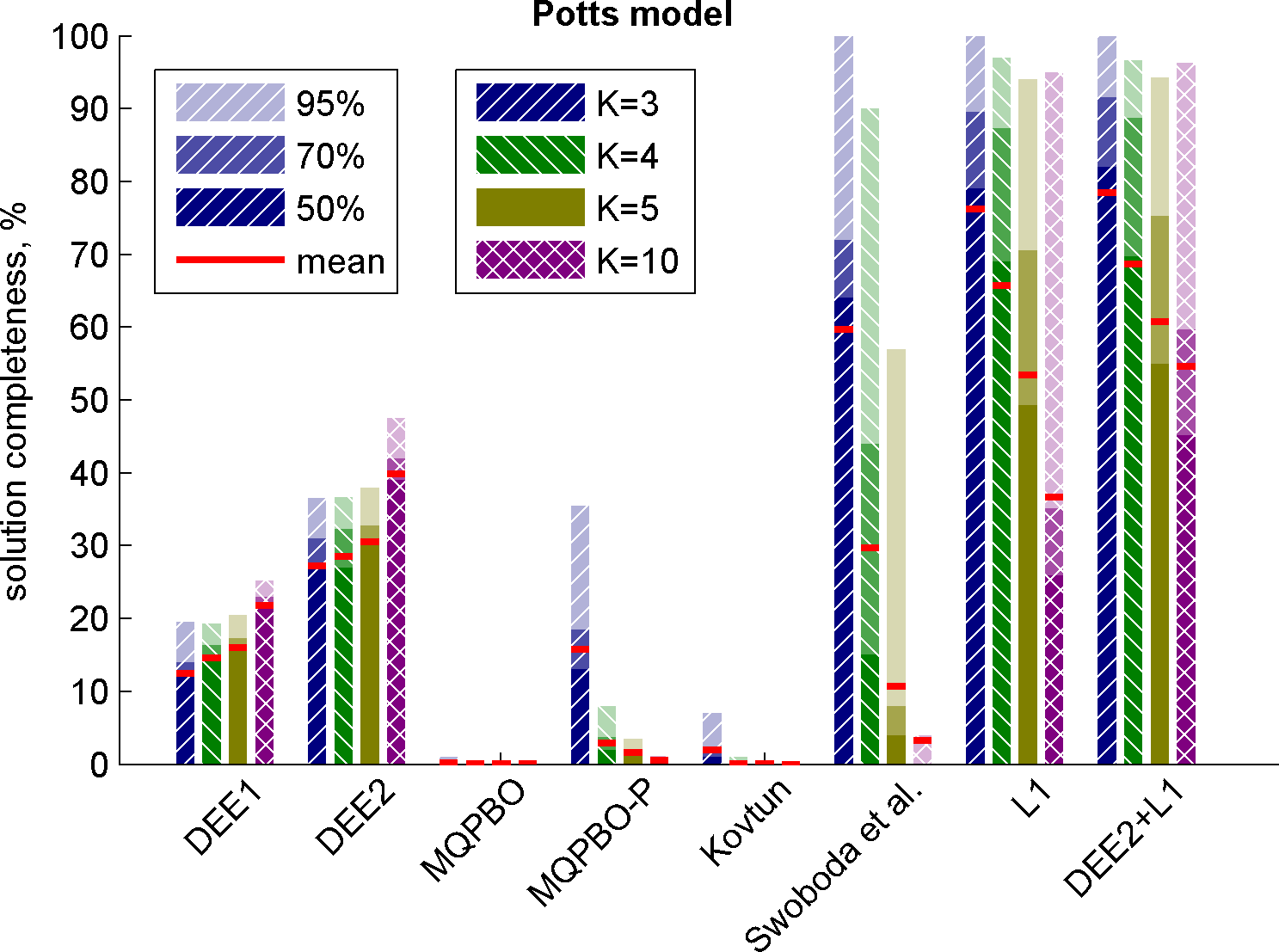}
\includegraphics[width=0.5\linewidth]{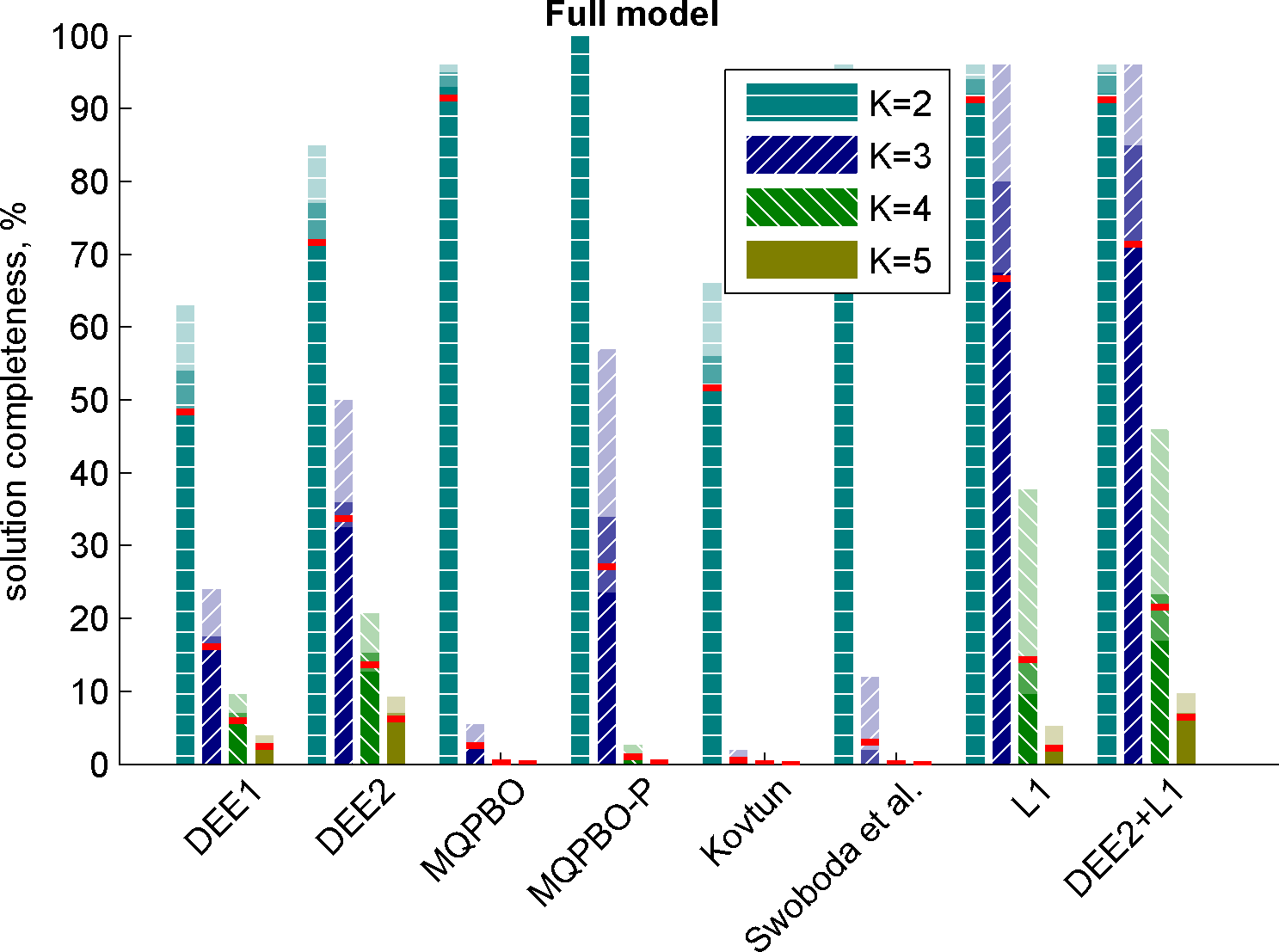}
\else
\includegraphics[width=0.5\linewidth]{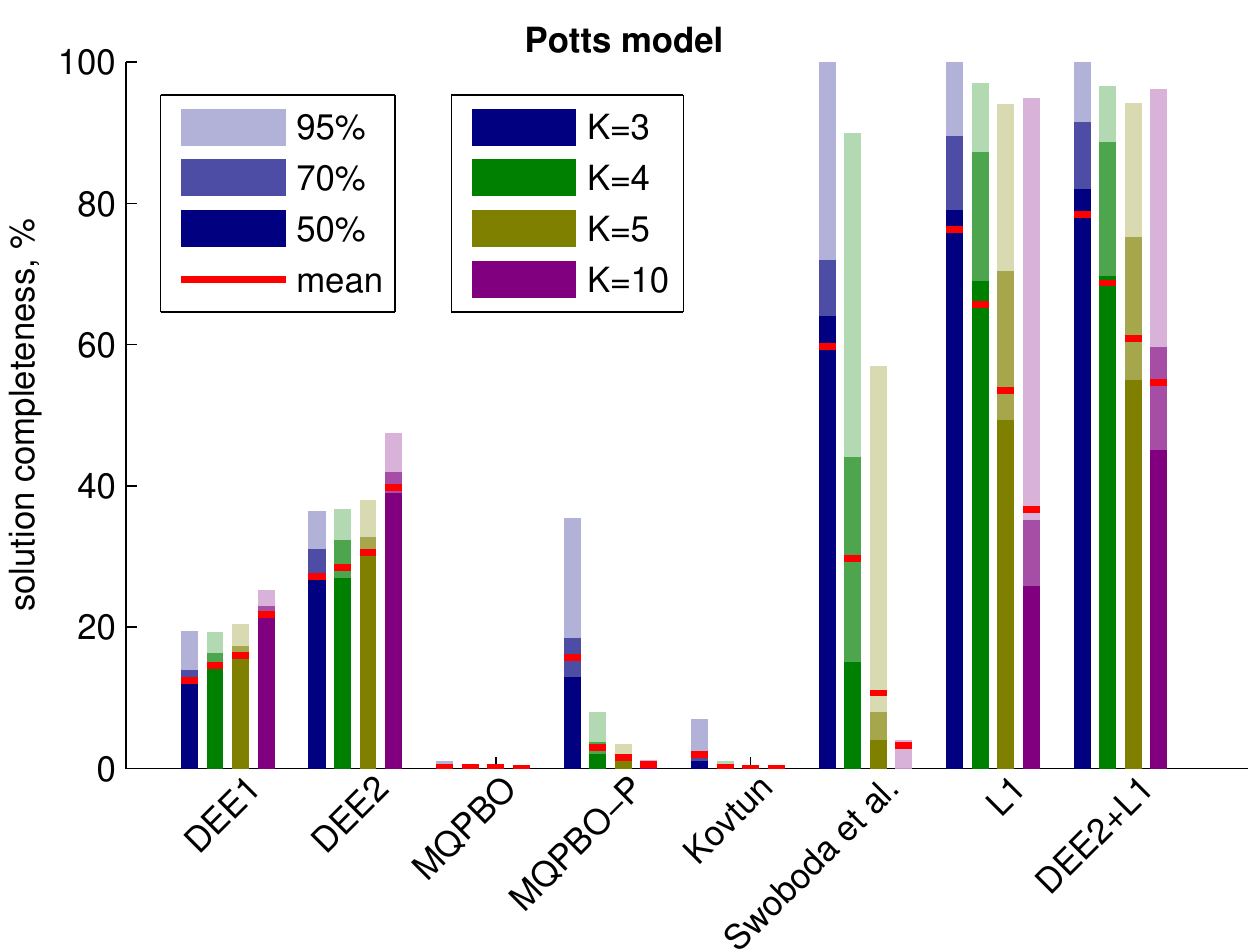}
\includegraphics[width=0.5\linewidth]{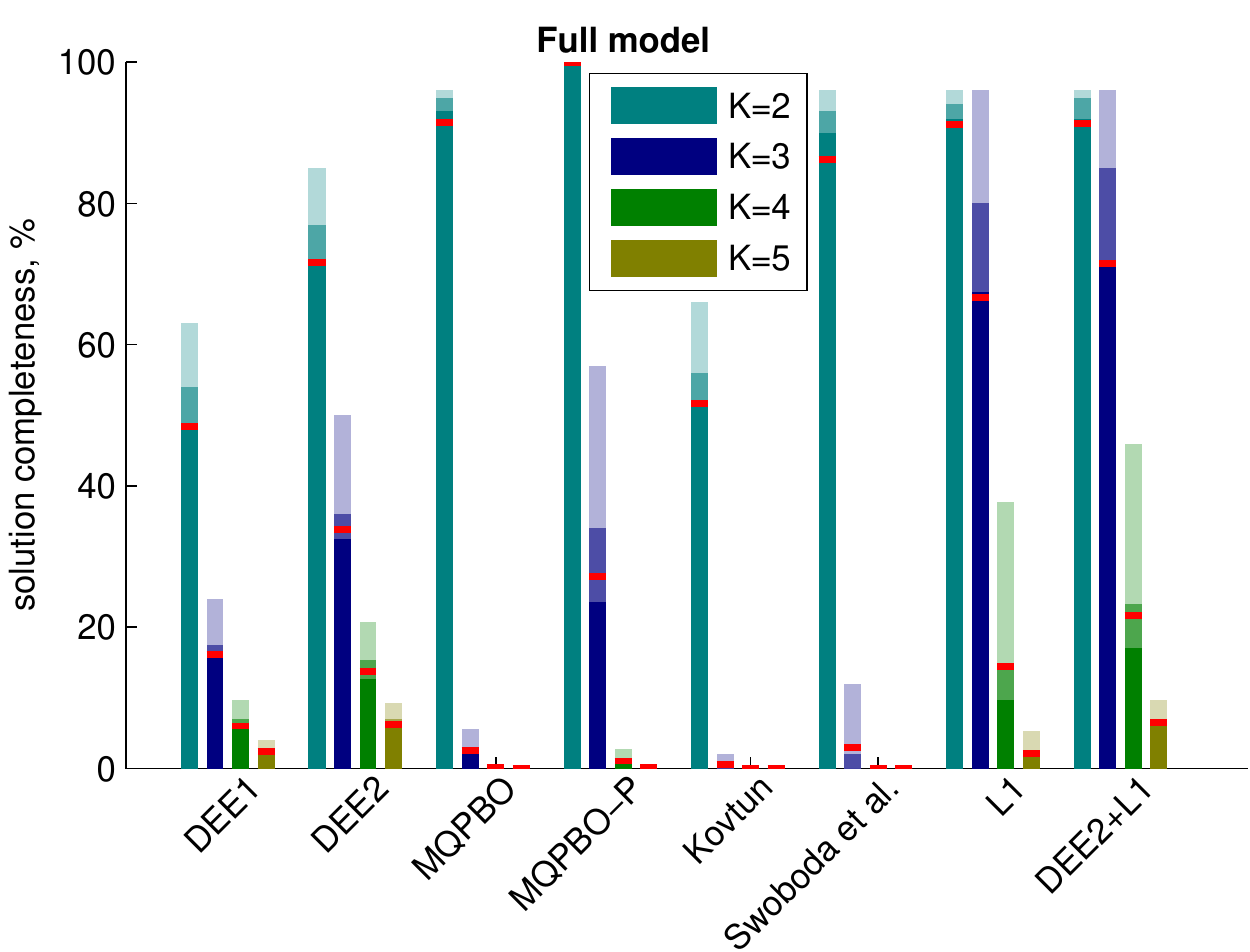}
\fi
\caption{Evaluation of pairwise multilabel methods. Problems size is 10x10, 4-connected. Bars of different shades indicate the portion of the sample under the given solution completeness value (statistics over 100 instances).}
\label{f:exp-rand}
\end{figure*}

\begin{figure*}[!t]
\ifblacknwhite
\includegraphics[width=0.5\linewidth]{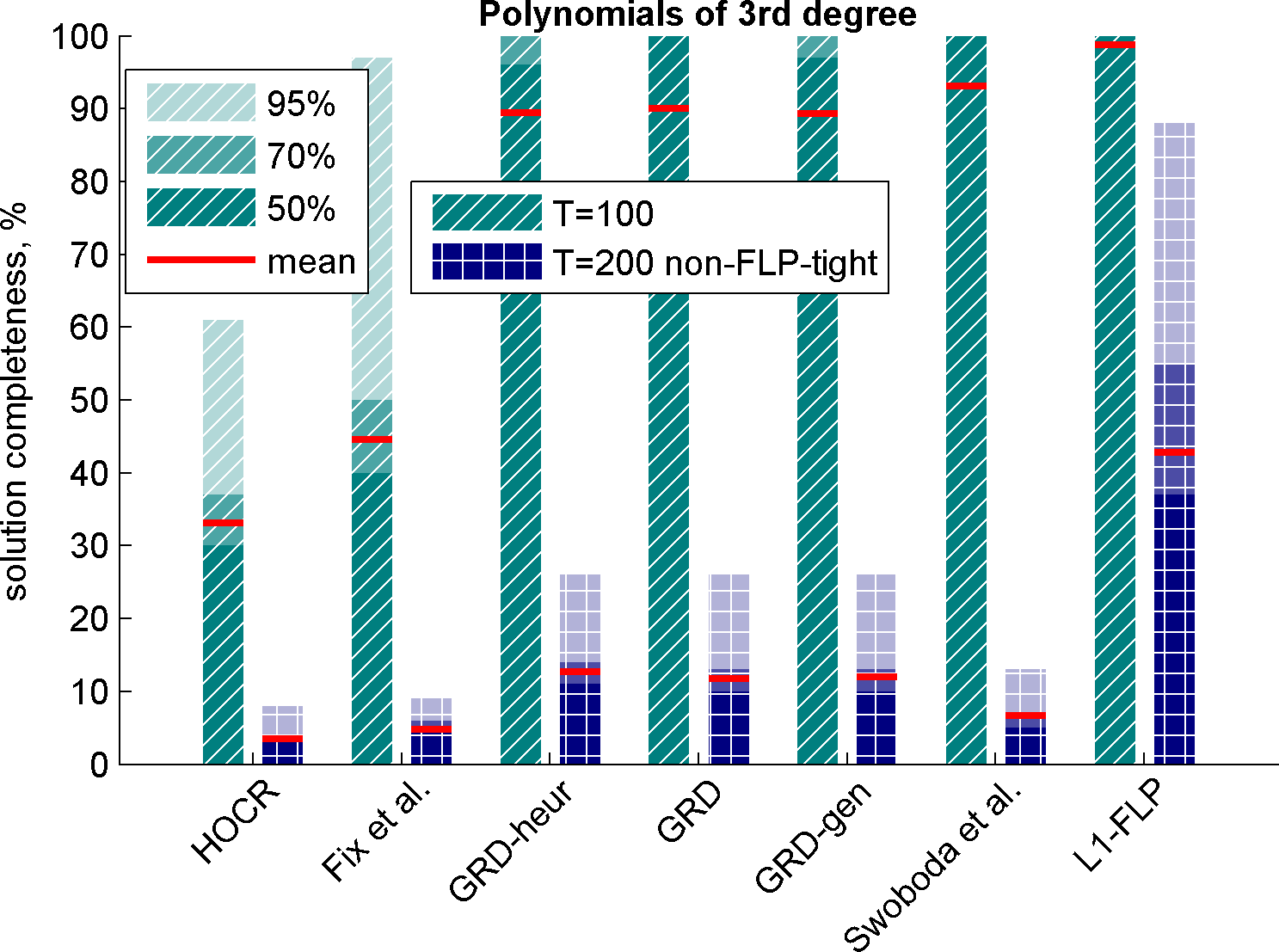}
\includegraphics[width=0.5\linewidth]{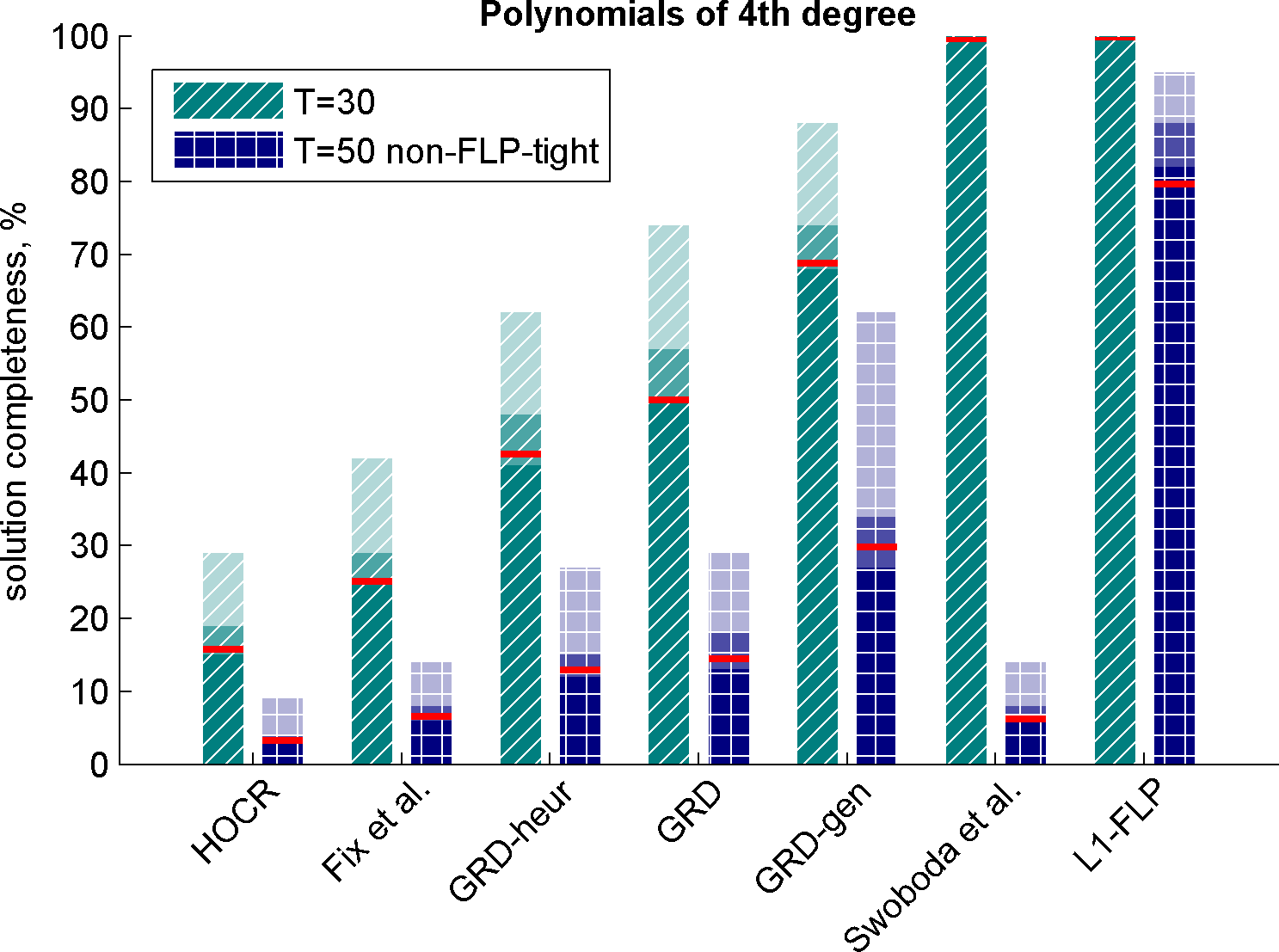}
\else
\includegraphics[width=0.5\linewidth]{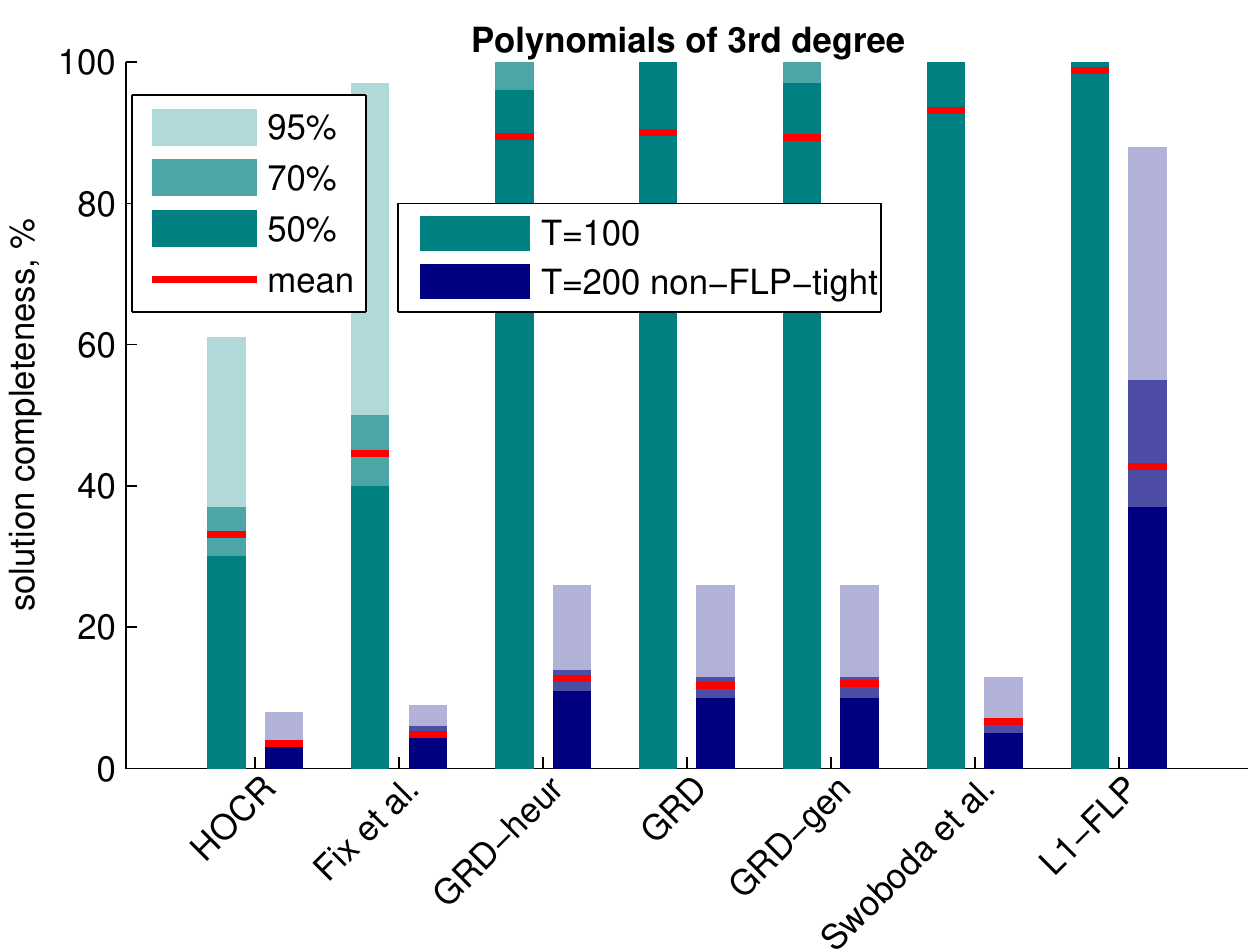}
\includegraphics[width=0.5\linewidth]{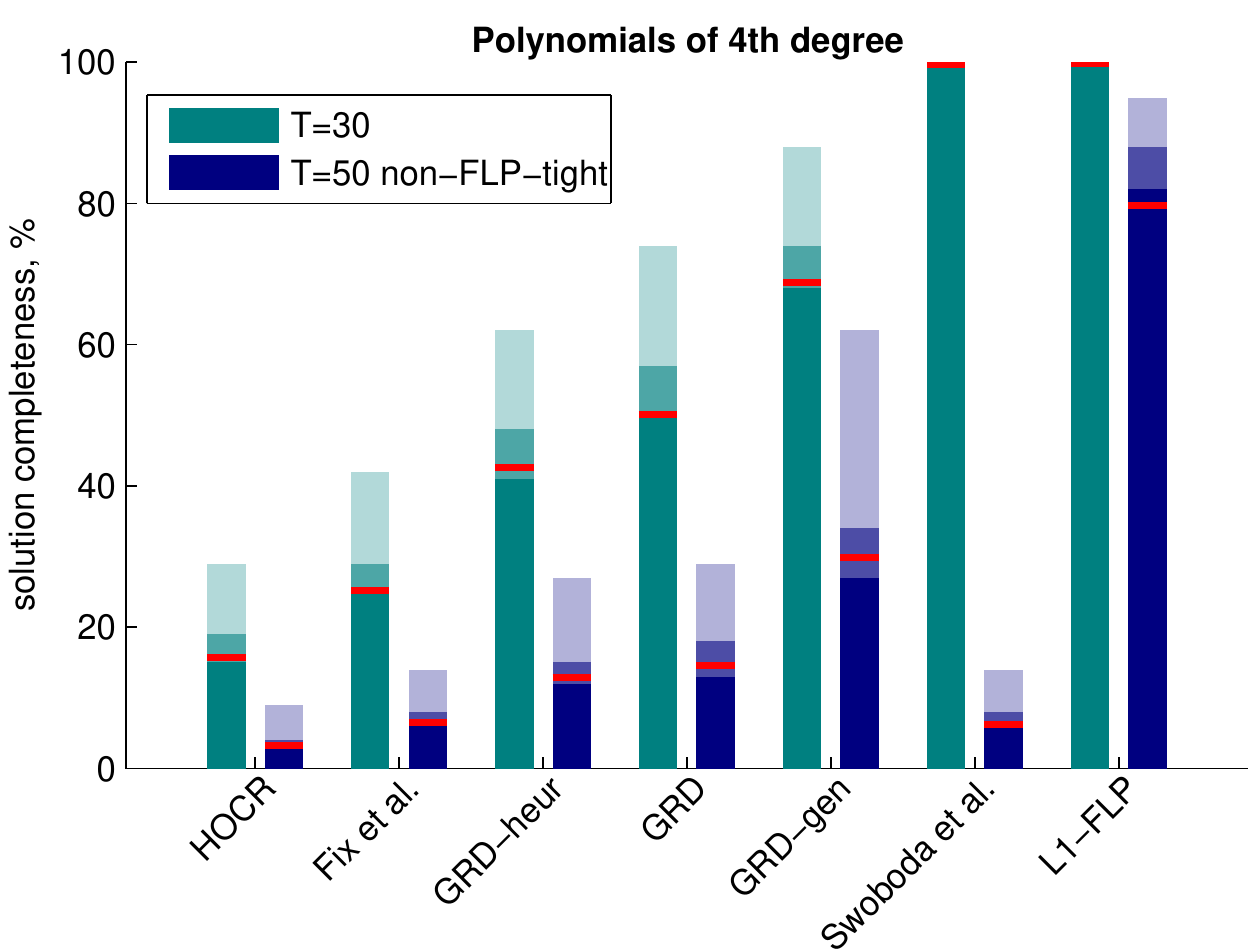}
\fi
\caption{Evaluation of higher-order binary methods on random polynomials generated as~\cite{KahlS-12} for $n=100$ variables. Plots with $(d=3,T=100)$ and $(d=4,T=30)$ reproduce results reported in~\cite{KahlS-12}. In is seen that most of the instances are FLP-tight and thus solved exactly by L1. We increase complexity by evaluating $(d=3,T=200)$ and $(d=4,T=50)$ and selecting only non FLP-tight instances.}
\label{f:exp-rand-h}
\end{figure*}

We evaluated all methods on small random problems. The purpose of the experiments is to validate the theory and to verify whether the improvement obtained by the new method is not negligible. For all methods, including ours, for each instance we numerically verified that:
\begin{itemize}
\item map $p$ constructed by the method is relaxed improving \wrt FLP relaxation (by solving the verification LP~\eqref{L-LP}).
\item the persistency guarantee is correct (by solving exactly the initial and the reduced problems).
\end{itemize}
We measure {\em solution completeness} as $\frac{n_{\rm elim}}{|\V|(K-1)}100\%$, where $K$ is the number of labels in every node ($|\X_s| = K$) and $n_{\rm elim}$ is the total number of labels $(s\in\V,i\in\X_s)$ eliminated by the method as non-optimal.
\paragraph{Pairwise Multilabel Models}
We report results on random problems with Potts interactions and full interactions. Both types have unary weights $f_s(i) \sim U[0,\,100]$ (uniformly distributed). Full random energies have pairwise terms $f_{st}(i,j) \sim U[0,\,100]$ and Potts energies have $f_{st}(i,j) = -\gamma_{st}(i)\leftbb i{=}j\rightbb$, where $\gamma_{st}(i)\sim U[0,\, 50]$. All costs are integer to allow for exact verification of correctness. Only instances with non-zero integrality gap \wrt FLP relaxation are considered (non-FLP-tight). The results are shown in Figure~\ref{f:exp-rand}, while Table~\ref{table1} gives details of the methods.
%
%
\begin{table*}
{\small
\setlength{\tabcolsep}{3pt}
\renewcommand{\arraystretch}{1.2}
\begin{tabular}{|p{0.076\linewidth}|p{0.900\linewidth}|}
\hline
\scriptsize DEE1 & Goldstein's Simple DEE~\cite{Goldstein-94-dee}: If $f_s(\alpha)-f_s(\beta) +\sum_{t\in\N(s)} \min_{x_t}[f_{st}(\alpha,x_t)-f_{st}(\beta,x_t)] \geq 0$ eliminate $\alpha$. Iterate until no elimination possible.\\
\scriptsize DEE2 & Similar to DEE1, but including also the pairwise condition:
$f_s(\alpha_s)-f_s(\beta_s) + f_t(\alpha_t)-f_t(\beta_t)
+f_{st}(\alpha_{st})-f_{st}(\beta_{st})
+\sum\lsub{t'\in\N(s)\backslash\{t\}} \min_{x_{t'}}[f_{st'}(\alpha_s,x_{t'})-f_{st'}(\beta_s,x_{t'})]
+\sum\lsub{t'\in\N(t)\backslash\{s\}} \min_{x_{t'}}[f_{tt'}(\alpha_t,x_{t'})-f_{tt'}(\beta_t,x_{t'})]
 \geq 0.$\\
\scriptsize MQPBO(-P) & The method of Kohli \etal~\cite{kohli:icml08}. The problem reduced to $\{0,1\}$ variables is solved by QPBO(-P)~\cite{Rother:CVPR07}, where ``-P'' is the variant with probing~\cite{Boros:TR06-probe}. In the options for probing we chose: ``use weak persistencies'', ``allow all possible directed constraints'' and ``dilation=1''.\\
\small Kovtun & One-against-all Kovtun's method~\cite{Kovtun-10}. We run a single pass over $\alpha = 1,\dots K$ (test labelings are $(y_s=\alpha\mid s\in\V)$). Labels eliminated in earlier steps are taken correctly into account in the subsequent steps. Reimplementation.\\
\small Swoboda \etal & Iterative Pruning method of~\citet{Swoboda-14} using CPLEX~\cite{CPLEX} for each iteration. Reimplementation.\\
\scriptsize L1 & The proposed method in~\Algorithm{Alg1} for class $P^{2}$ without perturbation, both phases solved with CPLEX~\cite{CPLEX}. 
\\
\scriptsize DEE2+L1 & Sequential application of DEE2 and L1. Note, DEE2 uses a condition on pairs which is not covered by the proposed sufficient condition under pairwise BLP relaxation. 
\\
\hline
\end{tabular}
}
\caption{List of tested methods for pairwise multilabel evaluation.}
\label{table1}
\end{table*}
\paragraph{Higher Order Binary Models}
The proposed evaluation of higher order $0$-$1$ models is based on the {\tt submodular} library~\cite{submodular,KahlS-12}.
The library interfaces quadratization techniques {\bf HOCR}~\cite{Ishikawa-11} and {\bf Fix \etal}~\cite{Fix-11} and implements three variants of generalized roof duality~\cite{KahlS-12}, {\bf GRD*}. \Figure{f:exp-rand-h} shows evaluation on random polynomials of degrees $3$ and $4$, sampled by the library. In the first series, we reproduce results~\cite{KahlS-12} with similar parameters but smaller problems (\eg, $n=100$ variables and $T=30$ multilinear terms \vs $n=1000$ and $T=300$ in~\cite{KahlS-12}). The results for baseline methods are consistent with~\cite{KahlS-12}. It turned out however that most of the instances are FLP-tight. Our method, as well as~\cite{Swoboda-14}, reduces in this case to solving the FLP relaxation and gives the trivial $100\%$ persistency result. In the second series we increased the complexity by adding more terms as well as selecting only non-FLP-tight instances. The proposed approach determines a significantly larger persistent assignment.

\begin{figure*}[!t]
\centering
\begin{tabular}{cc}
\begin{tabular}{c}
\ifblacknwhite
\includegraphics[width=0.5\linewidth]{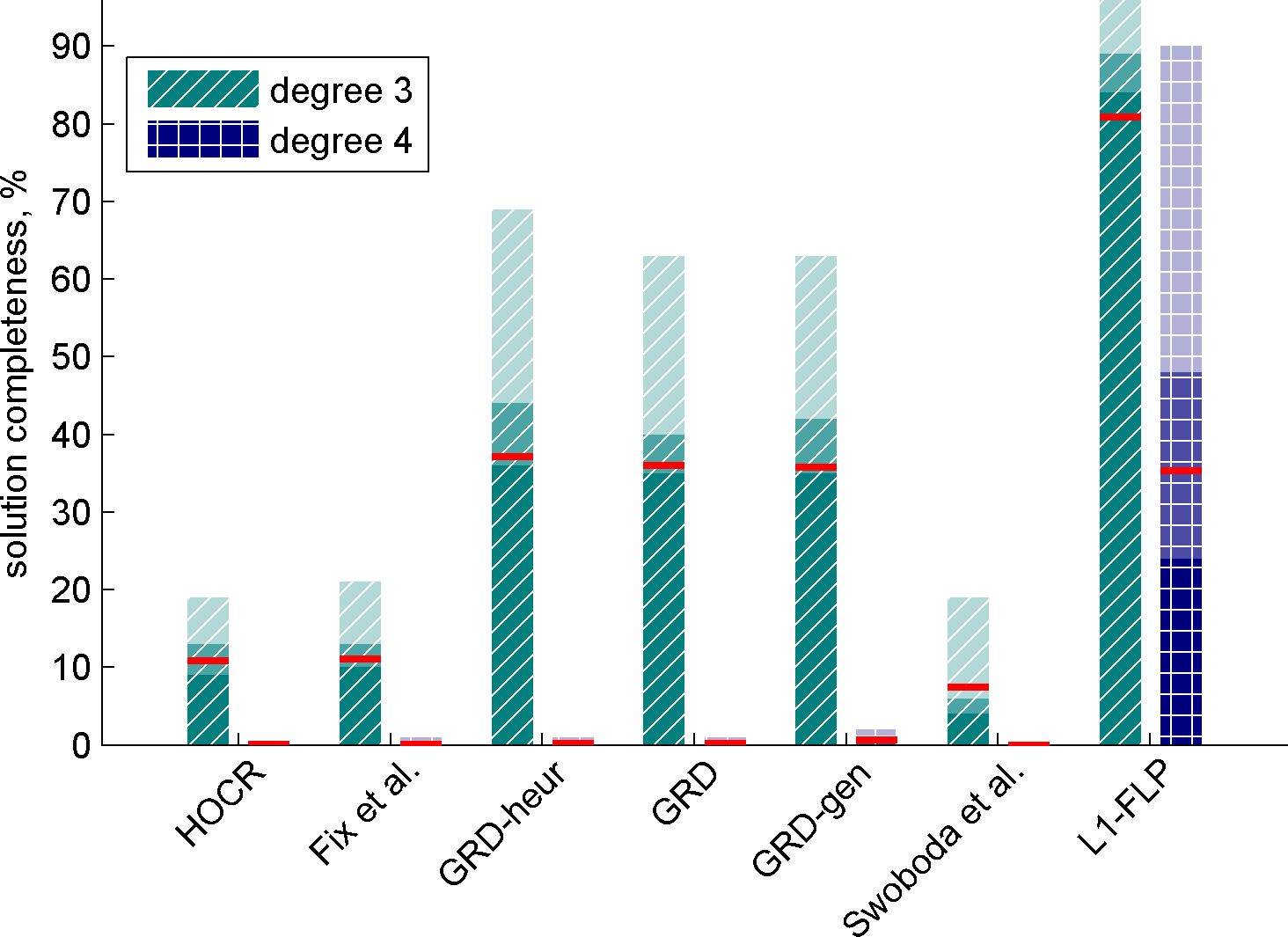}
\else
\includegraphics[width=0.5\linewidth]{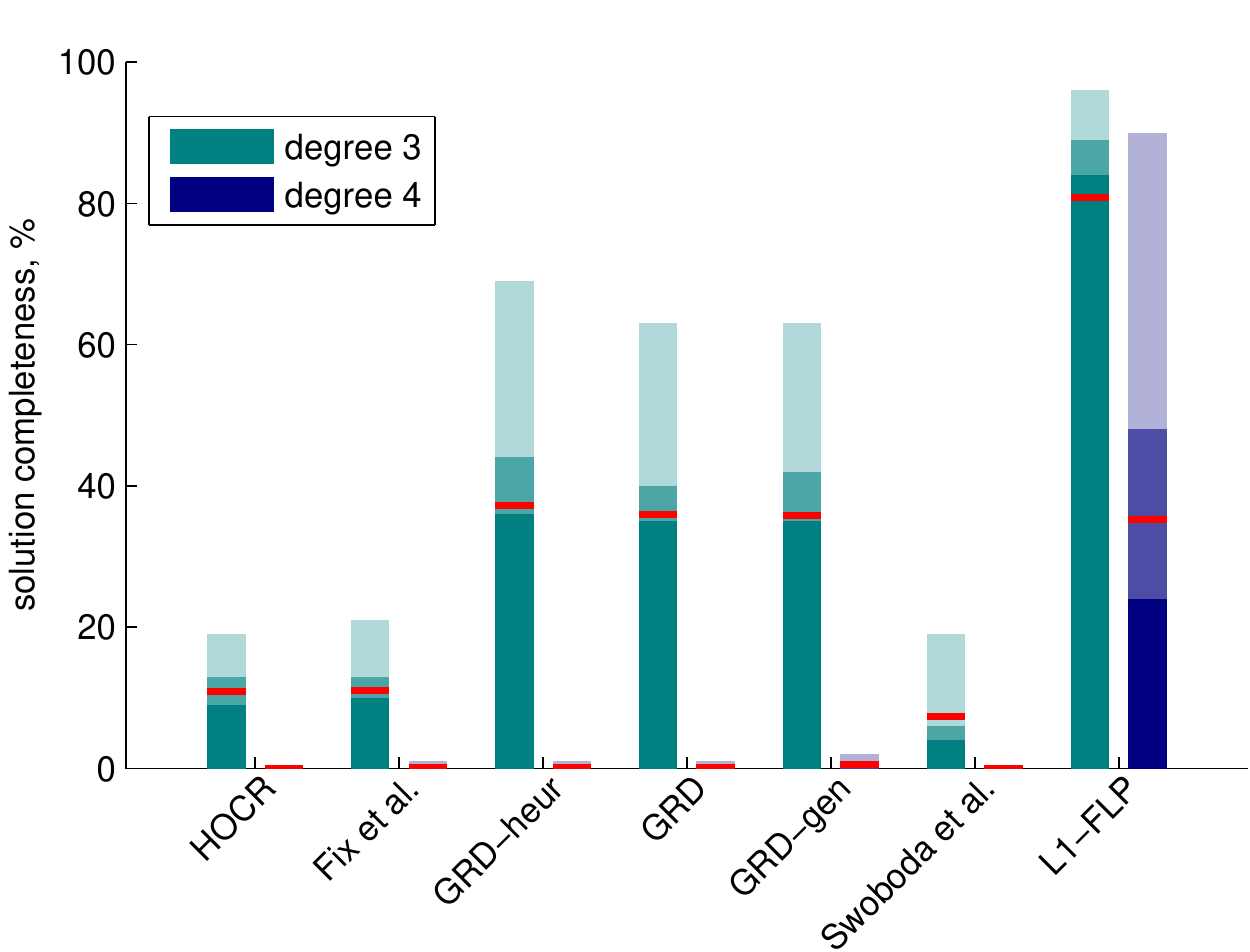}
\fi
\end{tabular}&
\begin{tabular}{c}
\includegraphics[width=0.5\linewidth]{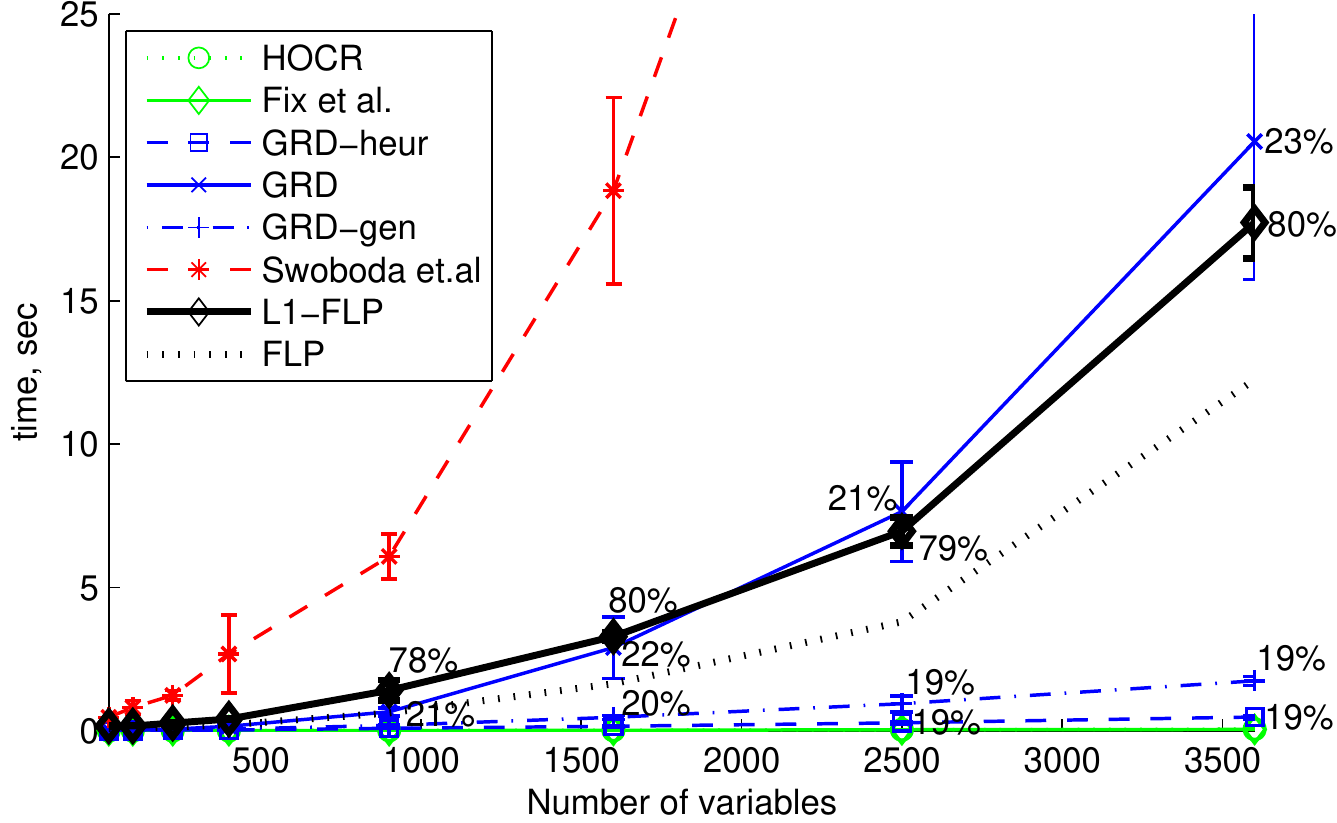}
\end{tabular}
\end{tabular}
\caption{
Evaluation of higher-order methods on non-FLP-tight problems on grid.
Left: Problems of degree 3 and 4 on grid size 10$\times$10. Note, degree 4 appears difficult for existing methods.
Right: Running time for problems of degree 3 and varied size up to 60$\times$60. Percentage indicates solution completeness for selected points. 
}
\label{f:exp-rand-h2}
\end{figure*}

In \Figure{f:exp-rand-h2}, we generated grid problems of degree $3$ with hyperedges {\small $\{(i,j),(i+1,j),(i,j+1)\}$} and of degree $4$ with hyperedges {\small $\{(i,j),(i+1,j),(i,j+1),(i+1,j+1)\}$} at every grid location $(i,j) \in\{1,\dots N-1\}^2$. 
For each such hyperedge $\c$ we sampled the term $f_\c$ as a random posiform ($2^{|\c|}$ uniformly distributed numbers, one per configuration, as opposed to sampling coefficients of multilinear polynomials in \Figure{f:exp-rand-h}). This results in somewhat more difficult problems to solve as there is no bias from an unsymmetrical treatment. We further selected only non-FLP-tight instances. It turns out that for the class of the problems of degree $4$ none of the baseline methods identified more than $1-2\%$ of the optimal solution, in contrast to the proposed method.
%
%

\paragraph{Running Time} 
\Figure{f:exp-rand-h2}(b) gives a rough idea of running times when using CPLEX to solve linear programs. The running time for L1-FLP and Swoboda \etal includes only the time to solve linear programs and excludes all data preparation in matlab.
The method of Swoboda \etal is the slowest one because it needs to solve several LP relaxations in the inner loop (but see~\cite{Swoboda-14,Swoboda-ArXiv2014} for applicability with suboptimal solvers and incremental computation). The proposed two-phase methods (L1-FLP) solves two linear programs. Somewhat unexpectedly, the initialization phase (FLP) takes more than a half of the total time. The optimal version of GRD performs similarly to the proposed method but determines less variables. GRD-heur is much faster while the result is comparable to GRD. It can be concluded that for practical applicability of the proposed method a feasible but possibly only approximately maximal solution should be found.
\section{Conclusions}\label{sec:conclusion}

Techniques for partial optimality avoid the NP-hardness of the energy minimization problem by exploiting different sufficient conditions by which a part of optimal solution can be found. 
We proposed a new sufficient condition corresponding to a given polyhedral relaxation and verifiable in polynomial time. 
The condition generalizes the mechanism of improving mapping which is present in many works (although often in a hidden form) and allows to explain them from this perspective.
We 
can explain variety of methods originating in different fields and relate these methods to linear relaxations. In particular, it follows that all covered methods cannot be used to tighten FLP relaxation. Applying them as a preprocessing in solving the FLP relaxation may only speed it up but cannot change the set of optimal relaxed solutions. 
We formally posed and studied the problem of determining the largest set of persistent variables subject to the general sufficient condition. It appeared that there are reasonably large classes of this problem (restricted by the set of allowed mappings) which can be solved in polynomial time. While the proposed solution might not be the most efficient, its generality allows to subsume multiple problem reformulations, reductions, equivalent transformations and choices that other existing techniques depend on. 
In bisubmodular relaxations, this is the choice of a bisubmodular lower bound function, in method~\cite{Kovtun-10} the choice of the order of labels and the test labeling, in methods~\cite{Ishikawa-11,Fix-11} choice of the sequence of the reductions and flips. While optimizing these methods \wrt to all such choices does not seem tractable, it is tractable to find a persistent assignment (by the proposed method) which is at least as good as if these choices were optimized over.
\par
In the experimental evaluation we verified that our theoretical comparisons hold true, \ie that all evaluated methods (except DEE2 for which we do not claim anything) have output FLP-improving maps in all test cases. Our linear program~\eqref{L1} had always integer optimal solution\footnote{With exception of few cases when CPLEX experienced a numerical error.} $\zeta$. The persistent assignment found by our method with FLP-relaxation was larger per instance and significantly larger on average. 
%
%
\subsection{Discussion}\label{sec:discussion}
\paragraph{Iterative Application}
Do we get more persistencies if the algorithm is run iteratively?
\par
If we consider FLP relaxation in the cases when maximality is guaranteed, a subsequent application of the method cannot give an improvement (it would contradict maximality). Maximality is achieved in pseudo-Boolean or multi-label class $\P^1$ under strong persistency. It is also achieved if we keep the test labeling $y$ fixed and consider the class $\P^{2,y}$ (for both weak and strong persistency). In the other cases it would be possible to improve by iterating the method. Because for BLP relaxation excluding persistent variables may lead to a tighter relaxation (see \Figure{fig:BLP}), the method can be iterated similarly to generalized roof duality, but the result is still dominated by L1-FLP. In the multi-label case we can iterate while varying the test labeling $y$, however this is computationally expensive and does not seem practical. 
%
\paragraph{Efficiency}
The present work focused on theoretical aspects. Practical applicability of the method requires some further research and development of efficient specialized methods that use approximate solutions of the relaxation as~\cite{Swoboda-14} or a windowing technique~\cite{shekhovtsov-14} or alike. Method of~\citet{Swoboda-14} performs not the best in \Figure{f:exp-rand-h2} and is also the slowest when implemented with CPLEX. However, it can be made optimal for class $\P^1$ as proposed in~\cite{Swoboda-ArXiv2014} and fast in practice using scalable dual solvers. Our most recent work in this direction~\cite{SSS-15-IRI} proposes an algorithm of this type for the pairwise multilabel case and $\P^{2,y}$ class of maps. It can be viewed as an alternative (combinatorial \wrt the mapping) algorithm for the problem (L1) and achieves the necessary efficiency.
%

\paragraph{Open Questions} 
It was shown that strict persistency leads to a tractable problem for a larger set of maps. It guarantees not to remove ambiguous solutions. By increasing $\varepsilon$ in the perturbation method one gets a potentially stronger guarantee \wrt the uncertainty of the data, which may be explored.
The general approach holds for an arbitrary bounded polytope, allowing one to incorporate also global linear constraints. This suggests a generalization to linearly constrained discrete optimization problems or mixed integer linear programs. 
%
Another interesting direction is how to combine the proposed persistency method with cutting plane techniques.
Finally, among the questions that remained open is polynomiality of \maxsi problem for the following cases:
(i) For all node-wise maps in a problem with 3 labels. (ii) For maps of the form $x\mapsto (x \vee x^{\rm \min}) \wedge x^{\rm max}$ with $x^{\rm min}$ and $x^{\rm max}$ being free variables, which is relevant when labels have a natural linear ordering. The optimal solution to this case would improve over the iterative method of Kovtun~\cite{Kovtun-10}, MQPBO and the method~\cite{Windheuser-et-al-eccv12}.


%
\appendices
\section{Proofs}\label{sec:A}
\LconiLambda*
\begin{proof}
We will show that the defining set \eqref{coni-Lambda-def} is contained in \eqref{coni-Lambda} and vice versa.
Let $\mu\in \Lambda$, let $\alpha \geq 0$. Then $A \alpha \mu = \alpha A \mu \geq 0$ and  $\alpha \mu \geq 0$. Therefore $\alpha \mu$ is contained in the set~\eqref{coni-Lambda}. Now let $\mu$ belong to the set~\eqref{coni-Lambda}. If $\mu_\emptyset > 0$, we can select $\alpha = \mu_\emptyset$ and vector $\mu' = \mu/\mu_\emptyset \in \Lambda$ and conclude that $\mu  = \alpha \mu'$ belongs to~\eqref{coni-Lambda-def}.
Let $\mu_\emptyset = 0$. Assume for contradiction that $\mu\neq 0$. Set $\Lambda$ is non-empty by assumption, let $\mu'\in \Lambda$. Then for any $\alpha \geq 0$ there holds $\mu'+\alpha\mu \in \Lambda$. But $||\mu'+\alpha\mu|| \geq \alpha||\mu||$, which is unbounded and contradicts boundedness of $\Lambda$. Therefore $\mu=0$, which belongs to the set~\eqref{coni-Lambda-def}. \qqed 
\end{proof}
%
%
\SIO*
\begin{proof}
By idempotency we have that $P(\Lambda) = \{\mu\in\Lambda,\ P\mu = \mu\}$. Let $\mu'\in P(\Lambda)$. Since $\<(I-P\T)f,\mu'\> = 0$ it is clear that 
the value of verification LP, $\min_{\mu\in\Lambda}\<(I-P\T)f,\mu\>$ is not positive.
\par
Direction ``$\Leftarrow$''. Let~\eqref{Def:SI-b} hold. For $\mu\in P(\Lambda) = \O$ condition~\eqref{Def:SI-eq} is trivially satisfied and we find that the value of verification LP is zero. For $\mu \in \Lambda\backslash P(\Lambda)$ from~\eqref{Def:SI-b} follows that $\mu$ is not a minimizer, therefore the objective must be strictly larger than zero and the strict inequality in~\eqref{Def:SI-eq} is satisfied.
\par
Direction ``$\Rightarrow$''.
Let $P$ satisfy~\eqref{Def:SI-eq}. Then the value of verification LP is zero. Since any $\mu \in P(\Lambda)$ satisfies $P\mu =\mu$ and achieves zero objective we have $P(\Lambda) \subset \O$. Now let $\mu \notin P(\Lambda)$. In this case $P\mu \neq \mu$ and inequality~\eqref{Def:SI-eq} is strict. Therefore $\mu\notin \O$.
\qqed  
\end{proof}
The condition~\eqref{Def:SI-b} can be further expressed in the components of mapping $p$ as follows.
%
%
\TSIcomponents*
\begin{proof}
Direction ``$\Rightarrow$'': We prove the negation of the implication. Assume $(\exists \c,\ \exists x_\c \in \O_\c)\ p_\c(x_\c) \neq x_\c$. Then $\exists \mu\in \O$ such that $\mu_\c(x_\c) > 0$. By idempotency, there is no $x'_\c$ such that $p_\c(x'_\c) = x_\c$. By~\eqref{pixel-ext} $([p]\mu)_\c(x_\c) =0 \neq \mu_\c(x_\c)$ and therefore $[p]\mu \neq \mu$. From~\eqref{Def:SI-b} follows that $[p]\notin\SI$.
\par
Direction ``$\Leftarrow$'': Let~\eqref{SI-components} hold. Let $\mu\in \O$. Then for all $\c$, $x_\c$ such that $\mu_{\c}(x_\c) > 0$ there holds $p_\c(x_\c) = x_\c$. It follows from~\eqref{pixel-ext} that $([p]\mu)_\c = \mu_\c$ and thus $[p]\mu = \mu$. Therefore $[p](\O) = \O$ and thus $\O \subset P(\Lambda)$ and the value of the verification LP is zero. Because $P(\Lambda)=\{\mu\in\Lambda \mid P \mu = \mu \}$ the value of the objective on $P(\Lambda)$ is zero and thus $ P(\Lambda) \subset \O$. From~\eqref{Def:SI-b} follows $[p]\in\SI$.
\qqed 
\end{proof}
\subsection{Properties}\label{sec:properties-proofs}
%
\necessaryLI*
\begin{proof}{\bf (i)\ }Assume $(\exists\mu\in\O)$ $P\mu \in\Lambda\backslash\O$. Then $\<f,P\mu\> > \<f,\mu\>$, therefore $p\notin\WI$. This proves equation~\eqref{necessary-wi-gen}.
\par
Assume for contradiction that~\eqref{necessary-wi-comp} does not hold, \ie\, $(\exists \c\in \E,\ \exists x_\c\in \O_\c)$ $x'_\c := p_\c(x_\c) \neq x_\c$. 
Because $x_\c\in \O_\c$ there exists $\mu\in \O$ such that $\mu_{\c}(x_\c) > 0$. 
From expression~\eqref{pixel-ext-comp} it follows that $([p]\mu)_\c(x'_\c) > 0$ but $x'_\c \not\in \O_\c$ and therefore $[p]\mu \notin \O$, which contradicts to~\eqref{necessary-wi-gen}.

{\bf (ii)\ }Assume $(\exists\mu\in\O)$ $P\mu \neq \mu$. Then $\<f,P\mu\> \geq \<f,\mu\>$ and therefore $p\notin\SI$. This proves equation~\eqref{necessary-si-gen}.
\par
Assume for contradiction that~\eqref{necessary-si-comp} does not hold, \ie\, $(\exists \c \in \V,\ \exists x_\c \in \O_\c)$ $p_\c(x_\c) \neq x_\c$. Because $x_\c\in \O_\c$ there exists $\mu\in \O$ such that $\mu_{\c}(x_\c) > 0$. Since $p_\c(x_\c) \neq x_\c$, from idempotency and~\eqref{pixel-ext} follows that $([p] \mu)_\c(x_\c) = 0 \neq \mu_\c(x_\c)$ and thus $[p]\mu \neq \mu$, which contradicts~\eqref{necessary-si-gen}.
\qqed
\end{proof}
%

%
\SSIdual*
\begin{proof} (i) The "if" part. 
Condition~\eqref{SI-dual} implies a weaker condition
\begin{align}
f^{{\varphi}} - [p]\T f \geq 0,
\end{align}
\ie it satisfies dual representation of $\WI$~\eqref{WI-dual} and therefore $p$ is relaxed-improving. It remains to prove strictness. The value of the verification LP in~\eqref{L-LP} is zero. The value of its dual problem
\begin{align}\label{L-LP-dual}
\max \{ \psi\in\Real \mid f^{{\varphi}} - [p]\T f - e_\emptyset \psi;\ \varphi \in\Real^m_+ \}
\end{align}
is thus also zero. It follows that $\varphi$ is optimal to~\eqref{L-LP-dual}. We need to show that for $\mu\in\O$ there holds $[p] \mu = \mu$. By multiplying~\eqref{SI-dual} with $\mu_\c(x_\c)$ and summing over $\c$ and $x_\c$ we obtain
\begin{align}
\<(I-[p]\T)f, \mu\>-\<\varphi,A \mu \> \geq \epsilon \<h, \mu\>.
\end{align}
Because $\mu, \varphi$ are optimal primal and dual solutions, by complementary slackness $\<\varphi,A \mu \> = 0$. 
Assume for contradiction that $[p] \mu \neq \mu$. Then $(\exists \c \in \E,\ \exists \x_\c \in \X_{\c})$ $([p] \mu)_{\c}(x_\c) \neq \mu_\c(x_\c)$. We consider now two cases
\newline\noindent
Case 1: if $p_\c(x_\c) \neq x_\c$, then by idempotency for all $x'_\c$ holds $p_\c(x'_\c) \neq x_\c$ and therefore from~\eqref{pixel-ext} we calculate that $([p]\mu)_\c(x_\c) = 0$. In this case from the assumption it must be $\mu_\c(x_\c)>0$ and 
\begin{align}
\<\mu,h\> \geq \mu_\c(x_\c)h_\c(x_\c) > 0.
\end{align}
Case 2: if $p_\c(x_\c) = x_\c$ then from~\eqref{pixel-ext} and the assumption follows $(\exists x'_\c \neq x_\c)$ such that $p_\c(x_\c') = x_\c$ and $\mu_\c(x'_\c) > 0$. In this case
\begin{align}
\<\mu,h\> \geq \mu_\c(x'_\c)h_\c(x'_\c) > 0. 
\end{align}
In both cases 1 and 2 we have $\<(I-[p]\T)f, \mu\> > 0$, which contradicts optimality of $\mu$.
\par
We now prove the ``only if'' part of (i).
Let $g = (I-P\T)f$ and let $(\mu,(\varphi,\psi))$ be a primal-dual strictly complementary pair of solutions to
\begin{equation} 
\min_{\mu\in \Lambda} \<g, \mu\>.
\end{equation}
Let $\O_\c$ be the $\c$-support set of primal solutions: $\O_\c = \{x_\c\in\X_\c \mid \mu_\c(x_\c) > 0 \}$. 
By \Statement{T:SI-components} and idempotency, there holds $p_\c(x_\c)=x_\c$ for $x_\c\in\O_\c$. 
%
By strict complementarity, for $x_\c\in \O_\c$ there holds $g^\varphi_\c(x_\c) = 0$ and for $x_\c \notin \O_\c$ there holds $g^\varphi_\c(x_\c) > 0$.
We let
\begin{equation}\label{varepsilon-choice}
\varepsilon = \min\csub{\c\in\E,\\ x_\c \in \X_\c\backslash \O_\c} \frac{g^\varphi_\c(x_\c)}{h_\c(x_\c)} > 0.
\end{equation}
Since for $x_\c\in\O_\c$ $h_\c(x_\c)=0$, we can bound now components of $g^\varphi$ as follows
\begin{align}
(\forall \c\in\E,\,\forall x_\c)\tab& g^{\varphi}_\c(x_\c) \geq \epsilon h_\c(x_\c).
\end{align}
Expanding components of $g^\varphi$ as $g^\varphi_{\c}(x_\c) = f_\c(x_\c)-f_\c(p_\c(x_\c))-(A\T \varphi)_{\c}(x_\c)$, we obtain relations~\eqref{SI-dual}.
\par
The statement of part (ii) of the theorem is proved as follows. The bit length of the rational dual solution $\varphi$ is polynomially bounded as well as the bit length of rational numbers $h$. It follows that $\varepsilon$ calculated by~\eqref{varepsilon-choice} is a rational number of polynomially bounded bit length.
\qqed
\end{proof}
%
%

\TChar*
\begin{proof} 
Let $g = (I-P\T)f$. The steps of the proof are given by the following chain:
\begin{equation}\label{proj2-chain}
\begin{array}{ll}
& \inf\limits_{\begin{subarray}{l} A \mu \geq 0 \\ \mu\geq 0\end{subarray}} \<f-P\T f,\mu \>
\stackrel{\rm (b)}{\leq} 
\inf\limits_{\begin{subarray}{l} A P \mu \geq 0 \\ A(I-P)\mu \geq 0\\ \mu\geq 0 \\ \ \end{subarray}} \<f-P\T f,\mu \> \\
& \stackrel{\rm (c)}{=} 
\inf\limits_{\begin{subarray}{l} \\ A(I-P)\mu \geq 0\\ \mu\geq 0\\ \ \end{subarray}} \<f-P\T f,\mu\>
\stackrel{\rm (d)}{=} \sup\csub{\varphi\in\Real^m_+ \\ (I-P\T)(f-A\T \varphi) \geq 0 } 0\,.
\end{array}
\end{equation}
On the LHS we have problem~\eqref{L-improving-coni}. If $P\in\WI$, this problem is bounded and the value of the problem is zero. 
Equalities (b), (c) essentially claims boundedness of the other two minimization problems in the chain. 
\par
Inequality~(b) is verified as follows. Inequality $\leq$ holds because by summing two inequalities
\begin{subequations}
\begin{align}
&A P \mu \geq 0,\\
&A (I-P) \mu \geq 0
\end{align}
\end{subequations}
we get $A \mu \geq 0$.
\par
Equality~(c) is the key step. We removed one constraint, therefore $\geq$ trivially holds. Let us prove $\leq$. Let $\mu$ be feasible to RHS of equality (c). Let $\mu=\mu_1 + \mu_2$, where
\begin{equation}
\begin{aligned}
\mu_1 = P \mu\,;\ \ \ \ \mu_2 = (I-P) \mu\,.
\end{aligned}
\end{equation}
There holds
\begin{equation}
\begin{aligned}
(I-P)\mu_1 & = (I-P)P \mu  = (P-P^2) \mu = 0\,,\\
P \mu_2 & = P (I-P) \mu = 0\,,
\end{aligned}
\end{equation}
\ie, $\mu_1 \in \Null(I-P)$ and $\mu_2 \in \Null(P)$. Let us chose $\mu_1'$ such that
\begin{equation}
\mu_1' \geq \mu_1 \tab\tab \AND \tab\tab A \mu_1' \geq 0\,.
\end{equation}
For example, the relaxed labeling
\begin{align}
\mu_1' = \gamma \frac{1}{|\X|}\sum_{x\in\X}\delta(x)
\end{align}
will satisfy these constraints for sufficiently large $\gamma>0$. Indeed, all components of $\mu_1'$ are strictly positive, it belongs to $\M$ as a convex combination of integer labelings and therefore satisfies constraints of the relaxation $A \mu_1' \geq 0$ for any $\gamma>0$. It remains to chose $\gamma$ large enough so as to have $(\mu_1') \geq \mu_1$ satisfied.
\par
Notice, that $(\mu_1')_\emptyset = \gamma$.
Let $\mu_1'' = P \mu_1'$. Because $P \geq 0$, we have
\begin{equation}
\mu_1'' = P \mu_1' \geq P \mu_1 = \mu_1\,.
\end{equation}
Because $\mu_1'\in \coni(\Lambda)$, there holds
$$
A P \mu_1'' = A P P \mu_1' = A P \mu_1' \geq 0.
$$
By idempotency, $(I-P) \mu_1'' = (I-P)P \mu_1' = 0$.
\par
Let $\mu^* = \mu_{1}''+\mu_{2}$. It preserves the objective,
\begin{align}
\<f-P\T f,\mu^*\> &= \<f,(I-P)(\mu_{1}''+\mu_{2}) \>  \\
\notag
&= \<f,(I-P)\mu_{2}\> = \<f,(I-P)\mu\>\,.
\end{align}
We also have that 
\begin{equation}
\begin{aligned}
\mu^* = \mu_{1}''+\mu_{2} \geq \mu_1 + \mu_{2} = \mu \geq 0\,,\\
A(I-P)\mu^* = A(I-P)\mu_2 = A(I-P)\mu \geq 0\,,\\
A P\mu^* = A P \mu_1'' \geq 0\,.
\end{aligned}
\end{equation}
Therefore, $\mu^*$ satisfies all constraints of the LHS of equality (c).
Equality~(d) is the duality relation that asserts that the maximization problem on the RHS is feasible. 
\par
\qqed
\end{proof}
\subsection{Relaxation of Sherali and Adams}\label{sec:SA-proof}
\SAequality*
\begin{proof}
The correspondence between $g$ and $G$ is through coefficients $\alpha$:
\begin{subequations}
\begin{align}
& g(z) = \sum_{\d\subset \c} \alpha_\d \prod_{s\in\d} z_s;\\
& G(\zeta) = \sum_{\d\subset \c} \alpha_\d \zeta_\d.
\end{align}
\end{subequations}
We have $g \equiv 0$ iff all coefficients of the (unique) multilinear polynomial representation $\alpha$ are zero and it is the case iff $G\equiv 0$. 
\end{proof}

For subsequent proofs let us introduce the correspondence between binary variables $z$ and their lifted representation $\zeta$ as the mapping $\zeta(z)$ from $\{0,1\}^\c$ to $\Real^{2^\c}$ with components
\begin{align}
\zeta(z)_\d = \prod_{s\in\d}z_s.
\end{align}
%
\SAinequality*
\begin{proof}
($\Leftarrow$) This part follows from the fact that $\zeta(z)$ satisfies all constraints of $\Zeta_\c$ and thus $g(z) = G(\zeta(z)) \geq 0$ for all $z\in\{0,1\}^\c$.
\par
($\Rightarrow$) Note that a special case when $g(z) = \prod_{s\in\b} z_s \prod_{s\in \a} (1-z_s)$ and $\a\cap \b = \emptyset$, $\a,\ \b\subset \c$ is proven in~\cite[Lemma 2]{SheraliA90}. Here is a different general proof.
\par
Since $g$ is non-negative, it can be represented as a posiform~\cite[Proposition 1]{BorosHammer02}:
\begin{align}
g(z) = \sum_{\b\subset \c} \alpha_{\b} \prod_{s\in\b} z_s \prod_{s\in \c\backslash \b} (1-z_s),
\end{align}
where $\alpha_{\b} = g(\bbbone_\b) \geq 0$ and $\bbbone_\b \in \{0,1\}^\c$ is the indicator of the set $\b$ defined as: $(\bbbone_\b)_s := \leftbb s\,{\in}\,\b \rightbb$. By linear combination property \Sthree, $G(\zeta)$ can be written as
\begin{align}
G(\zeta) = \sum_{\b\subset \c} \alpha_{\b} \sum_{\d\subset \c\backslash \b}(-1)^{|D|}\zeta_{\d\cup \b},
\end{align}
which is a non-negative combination of non-negative summands, as ensured by constraints of $\Zeta_\c$.
\end{proof}

\SAconvexhull*
\begin{proof}
Let $\H$ denote the convex hull~\eqref{zeta-set}.
Clearly, any vertex of $\H$ is in $\Zeta_\c$ and therefore $\H \subset \Zeta_\c$. 
\par
Let $G(\zeta) = \sum_{\d\subset \c}\alpha_\d \zeta_\d \geq 0$ be a facet-defining inequality of $\H$. Let us show it holds for all $\zeta\in\Zeta_\c$. Let $g(z) = \sum_{\d\subset \c}\alpha_\d \prod_{s\in\d}z_s$, $z\in\{0,1\}^\c$, a multilinear polynomial corresponding to $G$. 
For all vertices $\zeta(z)$ of $\H$ there holds $G(\zeta(z)) = g(z)$ and at the same time $G(\zeta)\geq 0$. It follows that $g(z) \geq 0$ for all $z\in\{0,1\}^\c$ and, by \autoref{L:SA-inequality}, $G(\zeta) \geq 0$ for all $\zeta\in\Zeta_\c$. We have proven that $\Zeta_\c \subset \H$.
\end{proof}

\SAproduct*
\begin{proof}
Using the convex hull property we can represent $\zeta\in\Zeta_\c$ as convex combination of vertices, \ie, $\zeta_\d = \sum_{k=1}^{n}\alpha_k \prod_{s\in\d} z^k_s$, $\d\subset\c$, where $z^k\in \{0,1\}^\c$ for $k = 1,\dots,n$ and $\alpha_k \geq 0$ and $\sum_{k=1}^n\alpha_k = 1$. Then
\begin{align}\label{product-proof-eq1}
&\zeta^2_\d = \sum_{k_1 = 1}^{n} \sum_{k_2 = 1}^{n} \alpha_{k_1} \alpha_{k_2} \prod_{s\in\d} z^{k_1}_s \prod_{s\in\d} z^{k_2}_s\\
&= \sum_{k_{1},\,k_2 = 1}^{n} \alpha_{k_1} \alpha_{k_2} \prod_{s\in\d} z^{k_1,k_k}_s,
\end{align}
where $z^{k_1,k_2} \in\{0,1\}^\c$ is the coordinate-wise product of $z^{k_1}$ and $z^{k_2}$. Note that $\alpha_{k_1} \alpha_{k_2} \geq 0$ and $\sum_{k_1,\,k_2 = 1}^{n} \alpha_{k_1} \alpha_{k_2} = 1$. Expression~\eqref{product-proof-eq1} proves that $\zeta^2$ is representable as a convex combination of vertices and thus belongs to $\Zeta_\c$. One could similarly show that for $\zeta,\,\eta\in\Zeta_\c$ their product $\zeta\eta$ also belongs to $\Zeta_\c$.
\end{proof}
%


\subsection{L1 construction}\label{L1-proofs}
In \Section{sec:LP-solution} we used representation of coefficients $P_{\c,x_\c,x'_\c}$ of the linear extension $[p_\zeta]$ in the form of a polynomial~\eqref{P_c-poly-expression}. This representation is obtained as follows. Starting from definition~\eqref{P-zeta-components}, we express:
\begin{align}
\notag
P_{\c,x_\c x'_\c} = & \prod_{s\in \c}\Big((\leftbb x'_s{=}x_s \rightbb-\leftbb y_s{=}x_s \rightbb) \zeta_{s, x'_s} + \leftbb y_s{=}x_s \rightbb \Big)\\
\label{extended-map-expr}
 = & \sum_{\d \subset \c} c_{\c,\d}(x_\c,x'_\d) \prod_{s\in \d}\zeta_{s,x'_s},
\end{align}
where 
\begin{equation}
\begin{aligned}\label{extended-map-coeffs}
& c_{\d}(x_s,x'_s) = \begin{cases}
\leftbb x_s'{=}x_s \rightbb-\leftbb y_s{=}x_s \rightbb, & s\in \d\\
\leftbb y_s{=}x_s \rightbb, & s\in \c \backslash \d
\end{cases}\\
& c_{\c,\d}(x_\c,x'_\d) = \prod_{s\in \c}c_{\d}(x_s,x'_s).
\end{aligned}
\end{equation}
\TSAequiv*
\begin{proof}
The fact that inequalities~\eqref{Zeta-set-c} imply $P_\zeta \geq 0$, assuming equality constraints~\eqref{Zeta-set-a}-\eqref{Zeta-set-b}, was shown in \Section{sec:LP-solution}.
%
We show now that $P_\zeta \geq 0$ implies inequalities~\eqref{Zeta-set-c}. 
\par
The inequality $P_\zeta \geq 0$ for the linear mapping $P_\zeta$ means that all its matrix elements are non-negative, \ie, the defining coefficients $P_{\c,x_\c,x'_\c}$ for $\c\in\E$, $x_\c\in \X_c$, $x'_c \in\X_\c$ are non-negative. 
Let us detail the constraint 
\begin{align}\label{P_C constraint}
P_{\c,x_\c,x'_\c} \geq 0.
\end{align}
Let 
\begin{equation}\label{set a(x)}
\a = \{s\in \c \mid x_s \neq x'_s\},\ \ \ 
\b = \{s\in \c \mid x_s \neq y_s\}.
\end{equation}
and let $\bar \a$ denote the complement in $\c$. 
From~\eqref{extended-map-coeffs} we have
\begin{align}\label{c_x_s1}
c_\d(x_s,x'_s) = \begin{cases}
1, &  s\in \bar \a \cap \b \cap \d,\\
-1,& s\in \a \cap \bar \b \cap \d,\\
0, & s\in \overline{(\a \triangle \b)} \cap \d,\\
1  &  s\in \bar \b \cap \bar \d,\\
0, &  s\in \b \cap \bar \d.\\
\end{cases}
\end{align}
Coefficient $c_\d(x_\c,x'_\c)$ in~\eqref{extended-map-coeffs}, which is the product of~\eqref{c_x_s1} over $s\in\c$, expresses as
\begin{align}\label{coeffs-c_d-sets}
c_\d(x_\c,x'_\c) = &(-1)^{|\a \cap \bar \b \cap \d|} 0^{|\d \backslash (\a \triangle \b)|} 0^{|\b \backslash \d|}.
\end{align}
It is non-zero only when $\d \backslash (\a \triangle \b) = \emptyset$ and $\b \backslash \d = \emptyset$ or equivalently
\begin{subequations}
\begin{align}
& \b \subset \d, \tab \d \subset \a \triangle \b;\\
\label{set-rels-a-b-d}
\Leftrightarrow\ \  & \b \subset \d, \tab \a \subset \bar \b, \tab \d \subset \a \cup \b.
\end{align}
\end{subequations}
Using sets $\a,\b$ and coefficients~\eqref{coeffs-c_d-sets} we obtain 
\begin{equation}\label{SA-type-constraint1}
P_{\c,x_\c,x'_\c} = \sum\csub{\d \\ \b \subset \d \subset \a \cup \b} (-1)^{|\d\backslash \b|} \zeta_{\d,x'_{\d}} \geq 0
\end{equation} which is equivalent to
\begin{equation}\label{SA-type-constraint2}
\sum\csub{\d \subset \a} (-1)^{|\d|} \zeta_{\d\cup\b,x'_{\d\cup\b}} \geq 0.
\end{equation}
In order to obtain inequalities~\eqref{Zeta-set-c} 
we need to show that for all $\x'_\c \in\tilde\X_\c$, by varying $x_\c \in\X_\c$, the set $\b$ ranges over all subsets in $\c$ while at the same time $\a$ equals to $\bar \b$. Since $\x'_\c\in\tilde \X_\c$ we have $x'_s \neq y'_s$ for all $s\in\c$. An arbitrary given set $\tilde \b$ can be realized by the choice $x_s = x'_s$ if $s\in \tilde \b$ and $x_s = y_s$ otherwise. At the same time for this choice of $\x$ there holds $\a = \bar \b$. 
\end{proof}
\Lsquaretozeta*
\begin{proof}
Let us calculate the expression of $(P^2)_{\c,x_\c x''_\c}$. It is equal to
\begin{subequations}\label{expr-both-factors}
\begin{align}
&\sum_{x_\c'\in \X_\c} P_{\c, x_\c x'_\c} P_{\c, x_\c' x''_\c} 
= \sum_{\d_1 \subset \c} \sum_{\d_2 \subset \c}\\
\label{expr-to-factor}
& \sum_{x'_\c \in \LL_\c} \prod_{s\in\c} c_{\d_1}(x_s,x'_s)c_{\d_2}(x'_s,x''_s) \zeta_{\d_1,x'_{\d_1}}\zeta_{\d_2,x''_{\d_2}}
\end{align}
\end{subequations}
The expression in line~\eqref{expr-to-factor} factors as 
\begin{subequations}
\begin{align}\label{first factor}
&\Big(\sum_{x'_{\d_1}\in \X_{\d_1}} \prod_{s\in \d_1} c_{\d_1}(x_s,x'_s)c_{\d_2}(x'_s,x''_s) \zeta_{\d_1,x'_{\d_1}} \Big)\\
\label{second factor}
&\Big(
\prod_{s\in \c \backslash \d_1} \sum_{x'_s\in\X_s} c_{\d_1}(x_s,x'_s)c_{\d_2}(x'_s,x''_s) \Big)
\Big(\zeta_{\d_2,x''_{\d_2}} \Big).
\end{align}
\end{subequations}
The term of the second factor for $s\in\c\backslash \d_1$ equals 
\begin{itemize}
\item Case $s\notin \d_1$, $s\in \d_2$:
\begin{align}\label{not SD1 sD2}
\sum_{x_s'}\leftbb y_s{=}x_s\rightbb (-\leftbb y_s{=}x'_s \rightbb + \leftbb x''_s{=}x'_s \rightbb) = 0.
\end{align}
\item Case $s\notin \d_1$, $s\notin \d_2$:
\begin{align}
\sum_{x_s'} \leftbb y_s{=}x_s\rightbb \leftbb y_s{=}x_s'\rightbb = \leftbb y_s{=}x_s\rightbb.
\end{align}
\end{itemize}
It follows from~\eqref{not SD1 sD2} that for $\d_2 \not \subset \d_1$ the factors vanishes and hence expression~\eqref{expr-to-factor} vanishes.
In the first factor the coefficient $c_{\d_1}(x_s,x'_s)c_{\d_2}(x'_s,x''_s)$ expresses as:
\begin{itemize}
\item Case $s\in \d_1$, $s\in \d_2$:
\begin{align}
\notag
&(-\leftbb y_s{=}x_s \rightbb + \leftbb x_s'{=}x_s \rightbb) (-\leftbb y_s{=}x'_s \rightbb + \leftbb x''_s{=}x'_s \rightbb)\\
& = (-\leftbb y_s{=}x_s\rightbb + \leftbb x_s'' {=}x_s\rightbb)\leftbb x_s'{=}x_s''\rightbb.
\end{align}
\item Case $s\in \d_1$, $s\notin \d_2$:
\begin{align}\label{SD1 not sD2}
\notag
&(-\leftbb y_s{=}x_s \rightbb + \leftbb x'_s{=}x_s \rightbb) \leftbb y_s{=}x_s'\rightbb\\
& = -\leftbb y_s{=}x_s{=}x_s'\rightbb + \leftbb x_s'{=}x_s{=}y_s\rightbb = 0.
\end{align}
\end{itemize}
Therefore, if $\d_1 \not \subset \d_2$, for each value of $x'_{\d_1}$ the product $\prod_{s\in\d_1} c_{\d_1}(x_s,x'_s)c_{\d_2}(x'_s,x''_s)$ vanishes and hence the sum~\eqref{first factor} and the expression~\eqref{expr-to-factor} vanish. 
It follows that we need to count expression~\eqref{expr-to-factor} only for the case $\d_1=\d_2 =:\d$. 
In this case, carrying the summation over $x'_{\d_1}$ in~\eqref{first factor} we obtain
\begin{equation}
\prod_{s\in \d}(-\leftbb y_s{=}x_s\rightbb + \leftbb x_s'' {=}x_s\rightbb)\zeta_{\d,x''_\d}.
\end{equation}
 From the full expression~\eqref{expr-both-factors} there remains
\begin{align}
\notag
&\sum_{\d \subset \c} \prod_{s\in \d}(-\leftbb y_s{=}x_s\rightbb + \leftbb x_s'' {=}x_s\rightbb)\prod_{s \in \c \backslash \d}\leftbb y_s=x_s\rightbb \zeta_{\d,x''_\d}\zeta_{\d,x''_\d}\\
& = \sum_{\d \subset \c} c_{\c,\d}(x'_\c,x''_\c) \zeta_{\c,x''_\d}^2 = (P_{\zeta^2})_{\c,x_\c,x''_\c}.
\end{align}
The claim $(P_\zeta)^2 = P_{\zeta^2}$ is proven.
\qqed 
\end{proof}
\subsection{Maximum Persistency for Local Relaxations}\label{sec:l1-s-proofs}
%
\ppreservesLambda*
\begin{proof}
Clearly, $[p]$ preserves non-negativity. Let $\mu$ satisfy marginalization constraint~\eqref{marg-d} for some $\c\in\E$, $\d\subsetneq \c$ and $x_\d \in \X_{\d}$. Then
\begin{subequations}
\begin{align}
\notag
& \sum\limits_{x_{\c \backslash \d }} ([p]\mu)_\c(x_\c) = 
\sum\limits_{x_{\c \backslash \d }} \sum_{x'_\c \in\X_\c} \leftbb p_\c(x'_\c){=}x_\c \rightbb \mu_{\c}(x'_\c)\\
& = \sum_{x'_\c} \Big( \sum\limits_{x_{\c \backslash \d }} \leftbb p_\c(x'_\c){=}x_\c \rightbb \Big) \mu_{\c}(x'_\c) \\
& = \sum_{x'_\c} \leftbb p_\d(x'_\d){=}x_\d \rightbb \mu_{\c}(x'_\c) \\
& = \sum_{x'_\d \in\X_\d} \leftbb p_\d(x'_\d){=}x_\d \rightbb \mu_\d(x'_\d) = ([p]\mu)_\d(x_\d).
\end{align}
\end{subequations}
\qqed
\end{proof}

\Lisclosed*
\begin{proof}
We need to prove that $(\forall \mu\in\Lambda)$
\begin{subequations}\label{P preserves local}
\begin{align}
\label{P_zeta c0 local}
& (P_\zeta \mu)_\emptyset = 1;\\
\label{P_zeta c1 local}
& P_\zeta \mu \geq 0; \\
\label{P_zeta c2 local}
& A P_\zeta \mu = 0.
\end{align}
\end{subequations}
Constraints~\eqref{P_zeta c0 local},~\eqref{P_zeta c1 local} are satisfied for any relaxation polytope (same as~\eqref{P preserve Lambda}). Constraint~\eqref{P_zeta c2 local} is local: for each $\c\subset\E$ it is given by equalities~\eqref{marg-d} for $(P \mu)_\c$, and hence involves only $(\zeta_\d \mid \d \subset\c)$. 
This constraint holds for all integer $\zeta$ by \Lemma{[p] preserves Lambda} and thus also for all $\zeta\in\Zeta$.
\end{proof}

\Tcharacterization*
\begin{proof}
We already have (a) $\Leftrightarrow$ (b) by \Theorem{S:WI-dual} and (a) $\Rightarrow$ (d) $\Leftrightarrow$ (c) by \Theorem{T:Char} (equivalence is the duality relation discussed in the proof).
Let us prove (c) $\Rightarrow$ (a). Let $\mu\in\Lambda$. We multiply component inequalities~\ref{char-c} 
with non-negative numbers $\mu_\c(x_\c)$ and sum over $x_\c$ and $\c$. We get inequality
$$
\<P\T f,\mu\> -\<P\T A\T\varphi,\mu\> \leq \<f,\mu\> -\<A\T\varphi,\mu\>.
$$
The sum $\<A\T \varphi, \mu\> = \<\varphi, A \mu\>$ vanishes because $A\mu=0$ and similarly $\<P\T A\T \varphi, \mu\> = \<\varphi, A P \mu\> = 0$ since by \Lemma{[p] preserves Lambda} $P\mu\in\Lambda$ and thus $A P \mu = 0$. The remaining inequality proves that $P\in\WI$. 
\qqed
\end{proof}

\subsection{NP Hardness Results}\label{sec:NP}
\TmaxwiNPhard*
\begin{proof}
In the quadratic pseudo-Boolean case the solution is given by the optimal relaxed labeling with the maximum number of integer components. For a more special case of vertex packing problem it was proven polynomial by~\citet{Picard-77}, whose proof we extended in~\cite[Statement 5]{shekhovtsov-14-TR}. For the general quadratic pseudo-Boolean case, the solution can be found efficiently by analyzing connected components in the network flow model~\cite{Boros:TR91-maxflow},~\cite[\S 2.3]{Kolmogorov-Rother-07-QBPO-pami}.
\par Next we prove that \maxwi is NP-hard if either: there are more than two labels or the order of the problem is 2 (cubic terms) or higher. We use reduction from pairwise constraint satisfaction problem (CSP) which is NP-complete when variables can take $3$ or more values or when constraints can couple 3 or more variables at a time (this case includes 3-SAT). This problem can be represented as energy minimization with constraints $f_\c \colon \X_\c \to \{0,1\}$. The CSP is satisfiable iff the minimum of the energy is zero. Let $V$ be the value of the LP-relaxation. Then either it is larger than zero and in this case the CSP is not satisfiable or it equals zero. In the latter case, the CSP is satisfiable iff there exist an integer solution with cost zero, \ie, the relaxation is tight. This is the case iff \maxwi determines all variables as persistent. Thus if \maxwi was in P we could solve CSP, which is a contradictions.
\qqed
\end{proof}

\TmaxsiNPhard*
\begin{proof}
We again use a reduction from CSP with 3 labels, $\X_s = \{1,2,3\}$. Let the constraints of CSP be defined by $g_\c \colon \X_\c \to \{0,1\}$. We construct an energy minimization problem with 4 labels as follows: 
\begin{equation}
f_\c(x_\c) = \begin{cases}
g_\c(x_\c), & x_\c \in\{1,2,3\}^\c;\\
\varepsilon & \x_\c = 4_\c;\\
B, & \OTHERWISE,
\end{cases}
\end{equation}
where $B > |\E|$ and $\varepsilon < 1/|\E|$. Let $V$ be the value of the BLP relaxation. If it is larger than zero, then the CSP is not satisfiable. Otherwise, the relaxation is tight iff the CSP is satisfiable. It is clear that if an integer solution of zero cost $y^*$ exists, it must take values in $\{1,2,3\}$. If it exists, then mapping $p_s \colon (1,2,3,4) \mapsto (1,2,3,y^*_s)$ is strictly relaxed improving as it replaces components of cost $\varepsilon > 0$ with components of cost $0$. Let $q$ be the maximum strict relaxed improving mapping in class $\P^{2}$. Let $x_s = 4$ for all $s\in\V$. If $(\exists s\in\V)$ $q(x)_s = 4$, then the CSP is not satisfiable, as $q$ is not larger than $p$. Otherwise $(\forall s\in\V)$ $q(x)_s \neq 4$ and $\f(q(x)) < |\E|\varepsilon < 1$. It follows that $g(q(x)) = 0$ and hence $q(x)$ is a solution to CSP. We showed that CSP was reduced in polynomial time to \maxsi.
\qqed
\end{proof}

\subsection{Method of Swoboda et al.}\label{sec:swoboda-proof}
\Tswobodacomp*
\begin{proof}
The method constructs a subset $\A \subset \V$, a labeling $y$ on $\A$ and an auxiliary energy $\g$ defined by:
\begin{align}
(\forall s\in \A) \tab & g_{s} = f_{s},\\
\notag
(\forall st\in \E,\ s\in \A, t\in \A) \tab & g_{st} = f_{st},
\notag
\end{align}
\begin{align}
\notag
& (\forall st\in \E,\ s\in \A, t \notin \A,\,\forall ij) \tab & \\
& g_{st}(i,j) = \begin{cases}
\max_{j'\in\X_t} f_{st}(i,j'), & i = y_s, \\
\min_{j'\in\X_t} f_{st}(i,j'), & i \neq y_s, \\
\end{cases}
\end{align}
with remaining terms set to zero. It can be seen that energy $\g$ depends on the assignment of $y$ only on the {\em boundary} $\partial\A = \{s\in \A \mid (\exists t\in \V\backslash \A)\ \{s,t\} \in \E\}$.
Let us extend $y$ to $\V$ in an arbitrary way, \eg, by $y_{\V\backslash \A} = 0$. 
The sufficient condition of~\cite[Corollary 1]{Swoboda-14} implies that $\delta(y) \in \argmin_{\mu\in\Lambda}\<g,\mu\>$ (the relaxation is tight). 
We construct mapping $p$ as
\begin{equation}
p_s(i) = \begin{cases}
y_{s} & \IF s \in \A,\\
i & \IF s \notin \A,
\end{cases}
\end{equation}
\ie, $p$ replaces part of a labeling $x$ on $\A$ with the labeling $y$. This mapping is illustrated in \Figure{fig:Swoboda}.
We claim that mapping $p$ is relaxed-improving.
\par
We first show that $g$ is {\em auxiliary} for $f$, \ie, that 
\begin{equation}\label{Kovtun-L-aux}
(\forall \mu \in \Lambda) \ \ \<(I-P)\T f,\mu\> \geq \<(I-P)\T g, \mu\>.
\end{equation}
We trivially have $f_s(i) - f_s(p_s(i)) = g_s(i) - g_s(p_s(i))$. We also have equality of pairwise terms 
$$
f_{st}(i,j) - f_{st}(p_s(i),p_t(j)) = g_{st}(i,j) - g_{st}(p_s(i),p_t(j))
$$ for $st\in \E$ in all of the following cases:
\begin{enumerate}
\item $s\in \A$ and $t \in \A$;
\item $s\notin \A$ and $t \notin \A$;
\item $s\in \A$ and $t \notin \A$, $i = y_s$.
\end{enumerate}
It remains to verify the inequality for boundary pairs $s\in\A$, $t\notin \A$ in the case $i\neq y_s$. We have
\begin{equation}
\begin{aligned}
& f_{st}(i,j) - f_{st}(p_s(i),p_t(j))\\
& \geq \min_{j'\in \LL_{t}}\big(f_{st}(i,j') - f_{st}(p_s(i),p_t(j')) \big)\\
\geq & \min_{j'}f_{st}(i,j') - \max_{j'}f_{st}(y_s,p_{t}(j'))\\
& = g_{st}(i,j) - g_{st}(p_s(i),p_t(j)).
\end{aligned}
\end{equation}
Because component-wise inequalities hold it follows that~\eqref{Kovtun-L-aux} holds. 
\par
The second step is to show that $p$ is relaxed-improving for $g$. By assumption, we have $\delta(y) \in \argmin_{\mu\in\Lambda}\<g,\mu\>$. 
Given a labeling $x$, mapping $p$ replaces part over $\A$ to the optimal labeling $y$.
It follows that $(\forall \mu\in\Lambda)$\ \ $\<g,P\mu\> = \<g,\delta(y)\> \leq \<g,\mu\>$. 
Combining this inequality with~\eqref{Kovtun-L-aux}, we obtain that $[p] \in \WI$.
\qqed
\end{proof}
\subsection{Quadratization Techniques}\label{sec:hocr-proof}
In this section we define a sufficient set of atomic reductions to represent methods~\cite{Ishikawa-11,Fix-11}. 
We first define a rather general reduction. 
\par
\begin{definition}\label{Def:reduction}
An {\em injective reduction} is a procedure that for a given energy minimization problem described by $\V,\E,\X,f$ specifies:
\begin{itemize}
\item The reduced energy minimization problem described by $\V',\E',\X',f'$; 
\item An injective mapping $\pi \colon \X \to \X'$;
\item A left inverse  of $\pi$, mapping $\sigma \colon \dom(\sigma) \to \X$, where $\X'\supset \dom(\sigma) \supset \pi(\X)$.
\end{itemize}
The energies are related by $\energy{f'}(\pi(x)) = \f(x)$ for all $x\in\X$.
\end{definition}
Mapping $\pi$ establishes a correspondence between labelings that preserves distinctness. Its left inverse always exists but may be non-unique (when $\pi(\X) \neq \X'$). For some labelings in $\X'$ there may be no meaningful correspondence in $\X$. For this reason the domain of $\sigma$ is allowed to be specified.
We introduce the following atomic injective reductions. 
First, the ones not changing the hypergraph:
\begin{itemize}
\item \codefunction[reparametrize]{Reparametrize}: Let $f' := f-A\T \varphi$, where $A$ is the matrix of a local relaxation. Mappings $\pi$ and $\sigma$ are identity. 
\item \codefunction[permute\char`_labels]{Permute}:
Apply a permutation (a bijection) of labels: for each $s\in\V$, mapping $\pi_s \colon \X_s \to \X_s$ is a bijection, $\X'_s = \X_s$ and $\sigma_s = \pi_s^{-1}$. The reduced energy is $f'_\c(\pi(x_\c)) = f_\c(x_\c)$.
\item \codefunction[add\char`_labels]{AddLabels}:
Expand variable domains as $\X_s \subset \X'_s$ and extend the objective component-wise: let $f'_\c(x'_\c) = f_\c(x'_\c)$ for $x'_\c \in \X_\c$ and arbitrary for $\x'_\c \in\X'_\c \backslash \X_\c$. Mapping $\pi \colon \X \to \X'$ is the injection $\x \mapsto x$. Mapping $\sigma \colon \X \to \X$ is the identity ($\dom(\sigma) = \X$).
\end{itemize}
Now, atomic reductions changing the hypergraph:
\begin{itemize}
\item \codefunction[remove\char`_zero]{RemoveZero}:
If the term $f_\c(x_\c) = 0$ for all $x_\c \in\X_\c$, exclude $\c$ from $\E$: let $\E' = \E\backslash \{\c\}$. 
Mapping $\pi$ is the identity. This operation is the converse of adding zero interactions in~\cite{Werner-PAMI07}.
\item \codefunction[clique\char`_aux]{CliqueReduce}:
For a given $\c \in\E$, introduce a new node $\omega$, an auxiliary variable $y\in\X_\omega$ and represent the term $f_\c$ as
\begin{equation}\label{clique_aux_min_def}
f_\c(x_\c) = \min_{y} f'_{\c \cup \omega}(x_\c,y).
\end{equation}
Define $\V' = \V \cup \{\omega\}$ and 
\begin{align}
\E' = \E \cup \{ \d \cup \{\omega\}  \mid \d\in \E,\ \d\subset \c \}.
\end{align}
For hyperedges $\d\in\E\backslash \{\c\}$ the energy terms are copied: $f'_\d(x_\d) =  f_\d(x_\d)$ and the remaining terms of $f'$ (other than in $(\E \backslash \{\c\}) \cup\{\c\cup\{\omega\}\}$) are zero.
The new labeling domain is $\X' = \X \times \X_\omega$. Mapping $\pi\colon \X \to \X'$ is defined as $\pi \colon x \mapsto (x,y(x))$ where 
\begin{align}
y(x)  = y(x_\c) \in \argmin_{y} f'_{\c\cup\{\omega\}}(x_\c,y).
\end{align}
Mapping $\sigma \colon \X' \to \X$ is the restriction $(x,y) \mapsto x$.
\item \codefunction[group\char`_aux]{GroupReduce}:
Let $\H \subset \E$ and intersection $\c = \bigcap_{\h\in\H}\h \in\E$.
Introduce a new node $\omega$ and a variable $y\in\X_\omega$. Let function $y \colon \X_\c \to \X_\omega$ be such that $(\forall \h \in \H)$
\begin{align}
\notag
f_\h(x_\h) = f'_{\h\cup\{\omega\}}(x_\h,y(x_\c)) = \min_{y\in\X_\omega} f'_{\h\cup\{\omega\}}(x_\h,y).
\end{align}
Define $\V' = \V \cup \{\omega\}$ and 
\begin{align}
\E' = \E \cup \bigcup_{\h\in\H} \{ \d \cup \{\omega\}  \mid \d\in\E,\ \d\subset\h\}.
\end{align}
Define $\pi\colon \X \to \X'\colon x \mapsto (x,y(x_\c))$ and $\sigma \colon \X' \to \X\colon (x,y) \mapsto x$.
\end{itemize}

In order to relate relaxed-improving maps before and after the reduction we need a correspondence between them. However, not all maps $p\colon \X'\to \X'$ of the reduced problem make sense for the initial problem.
\begin{definition}
A node-wise mapping $p'\colon \X' \to \X'$ is {\em admissible} for a reduction if 
$p'(\pi(\X)) \subset \dom(\sigma)$. In this case $p = \sigma \circ p' \circ \pi$ is the {\em corresponding} mapping for $p'$.
\end{definition}
All the above atomic reductions fulfill the requirements of~\autoref{Def:reduction} and, with the exception of \AddLabels reduction, all mappings $p'\colon \X' \to \X'$ are admissible.
\begin{definition}\label{def:maps-inclusion}
A reduction {\em has inclusion} of relaxed-improving maps for relaxations $\Lambda$ and $\Lambda'$ if for every admissible $\Lambda'$-improving mapping $p'$ for $f'$ its corresponding mapping $p$
is $\Lambda$-improving for $f$.
\end{definition}
\begin{theorem}\label{T:quad-preserve} Atomic reductions \Reparametrize and \Permute have inclusion of relaxed-improving maps for local relaxations.
\end{theorem}
\begin{proof} Let $\Lambda$ be a local relaxation. Let $p'$ be $\Lambda$-improving for $f'$. In case of \Permute, marginalization constraints are invariant \wrt the order of labels. We trivially get that $\sigma \circ p'\circ\pi$ is $\Lambda$-improving for $f$. In case of \Reparametrize, there holds $\<f-A\T\varphi,\mu\> = \<f,\mu\>$ for all $\mu\in\Lambda$ as well as for all $\mu \in [p'](\Lambda)\subset\Lambda$ and so $p = p'$ is $\Lambda$-improving for $f$.
\qqed 
\end{proof}
%
%



To a hypergraph $(\V,\E)$ let us associate an {\em induced} local relaxation $\Lambda_\E$ as follows. For $\a,\b\subset \V$ let $\a \Esubset \b$ if $\a,\b\in\E$ and $\a\subsetneq \b$. Let $\Lambda_\E$ be the local relaxation for the coupling $\Esubset$.

\begin{theorem}\label{T:quad-weaken1} Atomic reductions \RemoveZero, \AddLabels have inclusion of admissible relaxed-improving maps for local relaxations $\Lambda_{\E}$ and $\Lambda' = \Lambda_{\E'}$.
\end{theorem}
\begin{proof}[Proof \rm (\protect\RemoveZero)]
Because $\E'$ does not contain hyperedge $\c$, all variables $\mu_\c$, as well as the associated marginalization constraints, are absent in $\Lambda'$. Polytope $\Lambda'$ can be still represented in $\Real^\I$ as having unconstrained variables $\mu'_\c(x_\c)$. In this representation $\Lambda \subset \Lambda'$ and \Theorem{T:poly-inclusion} applies. \qqed
\end{proof}
\begin{proof}[Proof \rm (\protect\AddLabels)]
Let $p'$ be relaxed-improving for $f'$:
\begin{equation}
(\forall \mu'\in\Lambda')\ \<f', [p'] \mu' \> \leq  \<f', \mu' \>.
\end{equation}
By the dual representation~\Theorem{T:characterization}(b), there exists dual multipliers $\psi$ for $\Lambda'$ such that
\begin{equation}\label{component-inequalities-x}
(\forall \c\in\E,\ \forall x'_\c\in\X'_\c)\ f'_\c(p'(x'_\c)) \leq  f'^\psi_\c(x'_\c).
\end{equation}
It must be that $p'_s(\X_s) \subset \X_s$, otherwise $p'$ is not admissible. The corresponding mapping $p_s\colon \X_s\to \X_s$ is the restriction of $p'_s$ to $\X_s$. Restricting~\eqref{component-inequalities-x} to $\X_s$ we obtain $(\forall \c\in\E,\ \forall x_\c\in\X_\c)$
\begin{align}\label{add-labels-comp}
f_\c(p(x_\c)) \leq  f_\c(x_\c) - ((A')\T \psi)_\c(x_\c).
\end{align}
The RHS expression is of the form~\eqref{reparam-expand} and thus is a valid reparametrization of $f$ in $\Lambda$. 
Therefore~\eqref{add-labels-comp} satisfies component-wise inequalities and by \Theorem{T:characterization}(b), $p$ is relaxed-improving for $f$.
\qqed
\end{proof}
%
%
\begin{theorem}
\label{T:quad-weaken2}Atomic reductions \CliqueReduce and \GroupReduce have inclusion of relaxed-improving maps for local relaxations $\Lambda_\E$ and $\Lambda' = \Lambda_{\E'}$.
%
\end{theorem}
\begin{proof}[Proof \rm (\protect\CliqueReduce)]
First, let us show that a relaxed solution $\mu\in \Lambda$ can be mapped to a relaxed solution of the reduced problem $\mu'\in\Lambda'$. 
Let 
\begin{subequations}\label{clique-Pi}
\begin{align}
&(\forall \d\in\E)\ (\forall x_\d\in\X_\d)\  \ \mu'_{\d}(x_\d) := \mu_{\d}(x_\d), \\
& (\forall \d\subset \c,\ \d\in\E)\ (\forall x_\d\in\X_\d, \forall y\in\X_\omega)\\
\notag
&\mu'_{\d\cup\{\omega\}}(x_\d,y) := \sum_{x_{\c \backslash \d} }\mu_{\c}(x_\c) \leftbb y(x_\c){=}y \rightbb.
\end{align}
\end{subequations}
In particular, $\mu'_{\c\cup\{\omega\}}(x_\c,y) = \mu_{\c}(x_\c) \leftbb y(x_\c){=}y \rightbb$ and
$\mu'_{\{\omega\}}(y) = \sum_{x_\c}\mu_\c(x_\c) \leftbb y(x_\c){=}y \rightbb$.
Clearly, solution $\mu'$ satisfies all those marginalization constraints that $\mu$ does. The additional marginalization constraints of $\Lambda_{\E'}$ are given for $\b\subset\c$, $\a\subsetneq\b$, $\a,\b\in\E$ by couplings
\begin{align}
\a,\ \b,\ \a\cup\{\omega\}  \subsetneq \b\cup\{\omega\}.
\end{align}
%
Marginalization for $\b$ and $\b\cup\{\omega\}$:
\begin{align}
\notag
& \sum_{y} \mu'_{\b\cup\{\omega\}}(x_\b,y) = \sum_y\sum_{x_{\c \backslash \b} }\mu_{\c}(x_\c) \leftbb y(x_\c){=}y \rightbb\\
& = \sum_{x_{\c\backslash\b}}\mu_{\c}(x_\c)  = \mu_{\b}(x_\b) = \mu'_{\b}(x_\b).
\end{align}
Marginalization for $\a$ and $\b\cup\{\omega\}$:
\begin{align}
\notag
& \sum_{x_{\b\backslash \a}, y} \mu'_{\b\cup\{\omega\}}(x_\b,y) = \sum_{x_{\b\backslash \a},y}\sum_{x_{\c \backslash \b}}\mu_{\c}(x_\c) \leftbb y(x_\c){=}y \rightbb
\end{align}
\begin{align}
&= \sum_{x_{\c \backslash \a}}\mu_{\c}(x_\c) = \mu_{\a}(x_\a) = \mu'_{\a}(x_\a).
\end{align}
Marginalization for $\a\cup\{\omega\}$ and $\b\cup\{\omega\}$:
\begin{align}
\notag
& \sum_{x_{\b\backslash \a}} \mu'_{\b\cup\{\omega\}}(x_\b,y) = \sum_{x_{\b\backslash \a}}\sum_{x_{\c \backslash \b}}\mu_{\c}(x_\c) \leftbb y(x_\c){=}y \rightbb \\
&= \sum_{x_{\c\backslash \a}}\mu_{\c}(x_\c) \leftbb y(x_\c){=}y \rightbb = \mu'_{\a\cup\{\omega\}}(x_\a,y).
\end{align}
Therefore $\mu'\in\Lambda'$.
%
Since we have the equality $f_\c(x_\c) = f'_{\c\cup\{\omega\}}(x_\c,y(x_\c))$, there holds
\begin{align}\label{clique-marg1}
\notag
&\sum_{\x_\c}f_\c(x_\c) \mu_\c(x_\c) = \sum_{\x_\c}f'_{\c\cup\{\omega\}}(x_\c,y(x_\c)) \mu_\c(x_\c)\\
\notag
&= \sum_{\x_\c,y}f'_{\c\cup\{\omega\}}(x_\c,y)\leftbb y{=}y(x_\c)\rightbb \mu_\c(x_\c)\\
&= \sum_{\x_\c,y}f'_{\c\cup\{\omega\}}(x_\c,y)\mu'_{\c\cup\{\omega\}}(x_\c,y).
\end{align}
For other components $\d\in\E\backslash\{\c\}$ we have 
\begin{align}\label{clique-marg2}
\sum_{\x_\d}f_\d(x_\d)\mu_\d(x_\d) = \sum_{\x_\d}f'_\d(x_\d)\mu'_\d(x_\d)
\end{align}
and for all $\d\subsetneq\c$, $\d\in\E$ we have $f'_{\d\cup\{\omega\}} = 0$.
Let us denote the linear mapping $\Real^\I \to \Real^{\I'}$~\eqref{clique-Pi} by $\Pi$. It follows that for all $\mu\in\Lambda$ there holds $\Pi\mu \in\Lambda'$ and from~\eqref{clique-marg1},\eqref{clique-marg2} that
\begin{align}
\<f,\mu\> = \<f',\Pi \mu\>.
\end{align}
\par
This proves that 
\begin{align}
\min_{\mu\in\Lambda}\<f,\mu\> \geq \min_{\mu'\in\Lambda'}\<f',\mu'\>,
\end{align}
\ie, relaxation $\Lambda'$ is not tighter than $\Lambda$. 
\par
Let $p'\colon \X'\to \X'$ be node-wise $\Lambda'$-improving for $f'$ and $p = \sigma \circ p' \circ \pi$, \ie, $p_s = p'_s$ for $s\in\V$. Let $P' = [p']$, the extension to $\Real^{\I'}$.
Similarly to~\eqref{clique-marg1} and~\eqref{clique-marg2} we can express parts of the scalar product 
\begin{align}
\<f,[p]\mu\>  = \sum_{\d\in\E} \sum_{x_\d \in\X_\d}f_\d(p'_\d(x_\d)) \mu_\d(x_\d)
\end{align}
for each $\d\in\E$ as follows. For $\d\in\E\backslash\{\c\}$ we have 
\begin{align}
&\sum_{\x_\d}f_\d(p'_\d(x_\d))\mu_\d(x_\d) \\
\notag
& = \sum_{\x_\d} \sum_{\tilde x_\d} f_\d(\tilde x_\d) \leftbb \tilde x_\d{=}p'(x_\d) \rightbb \mu'_\d(x_\d)\\
& = \sum_{\tilde \x_\d} f'_\d(\tilde x_\d) (P'\mu')_{\d}(\tilde x_\d).
\end{align}
For $\c$ let $\c'=\c\cup\{\omega\}$, we have 
\begin{subequations}
\begin{align}
&\sum_{\x_\c}f_\c(p'_\c(x_\c))\mu_\c(x_\c) \\
\notag
& = \sum_{\x_\c} \sum_{\tilde x_\c} f_\c(\tilde x_\c) \leftbb \tilde x_\c{=}p'(x_\c) \rightbb \mu_\c(x_\c)\\
& = \sum_{\x_\c} \sum_{\tilde x_\c} f'_{\c'}(\tilde x_\c, y(\tilde x_\c)) \leftbb \tilde x_\c{=}p'(x_\c) \rightbb \mu_\c(x_\c)\\
\notag
& \leq \sum_{\x_\c} \sum_{\tilde x_\c} f'_{\c'}(\tilde x_\c, p'_\omega(y(x_\c))) \leftbb \tilde x_\c{=}p'(x_\c) \rightbb \mu_\c(x_\c)\\
& 
\begin{aligned}
= \sum_{\x_\c,y} \sum_{\tilde x_\c, \tilde y} f'_{\c'}(\tilde x_\c, \tilde y) \leftbb \tilde x_\c{=}p'(x_\c) \rightbb \leftbb \tilde y{=}p'_\omega(y) \rightbb& \\
\leftbb y{=}y(x_\c) \rightbb  \mu_\c(x_\c)&
\end{aligned}\\
& = \sum_{\x_\c,y} \sum_{\tilde x_\c, \tilde y} f'_{\c'}(\tilde x_\c, \tilde y) P'_{\c',(\tilde x_\c, \tilde y),(x_\c,y)} \mu'_{\c'}(x_\c,y)\\
& = \sum_{\tilde x_\c, \tilde y} f'_{\c\cup\{\omega\}}(\tilde x_\c, \tilde y) (P' \mu')_{\c\cup\{\omega\}}(\tilde x, \tilde \y).
\end{align}
\end{subequations}
The inequality is due to $\y(\tilde x_\c) \in \argmin_y f'_{\c'}(\tilde x_\c, y)$. For all $\d\subsetneq\c$, $\d\in\E$ there holds $f'_{\d\cup\{\omega\}} = 0$. It follows that
\begin{align}\label{clique-aux-final}
\<f,[p]\mu \> & = \<f' ,[p'] \mu' \> \\
\notag
& \leq \<f' ,\mu' \>  = \<f' , \Pi \mu \> = \<f,\mu\>.
\end{align}
%
%
%
Therefore $[p]\in\WI$.
%
%
%
%
%
\end{proof}
\begin{proof}[Proof \rm (\protect\GroupReduce)]
%
Because $y(x_\c)$ minimizes all group terms simultaneously, arguments of the proof for \CliqueReduce apply to each $\h\in\H$. It follows that the final relation~\eqref{clique-aux-final} is satisfied. 
\qqed
\end{proof}
\THOCR*
\begin{proof}
Reductions used in HOCR are of the following form:
\begin{align}
x_1 x_2 \dots x_n = \min_{y_1,y_2,\dots y_k}g(x,y),
\end{align}
where function $g(x,y)$ is by design partially separable so that the reduction decreases the order of the problem. This reduction can be implemented as follows:
\begin{itemize}
\item Use \CliqueReduce to introduce new variables $y_1,\dots y_k$, one at a time, \eg, first let
$x_1 x_2 \dots x_n = \min_{y_1} f^1(x,y_1)$, where $f^1(x,y_1) = \min_{y_2,\dots y_k}g(x,y)$, and so on.
\item Use \Reparametrize to rewrite the term $g(x,y)$ that gets assigned to the hyperedge over all variables $x_1,\dots, x_n$ and $y_1,\dots, y_k$  as a sum of terms over smaller subsets of variables as guaranteed by design of $g$.
\item Use \RemoveZero to clean up the hypergraph so that all now zero higher order terms are excluded.
\end{itemize}
HOCR full reduction can be implemented by iterating the steps above. Starting from the FLP relaxation of the initial problem, the necessary \Reparametrize operations are feasible as we add all possible interactions during \CliqueReduce. 
Each atomic reformulation has inclusion of relaxed-improving maps. It follows that the whole reformulation has inclusion of relaxed-improving maps. The persistency in the final reformulation is obtained with QPBO, which by \Theorem{T:QPBO-W} is an FLP-improving mapping. It follows that there is a corresponding FLP-improving mapping in the initial formulation.
\end{proof}
\par
\TFix*
\begin{proof}
Method~\cite{Fix-11} uses reductions of HOCR and a new reduction for a group of cliques~\cite[Theorem 3.1]{Fix-11}. Let us show that the group reduction has the form that can be implemented using \GroupReduce, \Reparametrize and \RemoveZero. The optimal value of the auxiliary variable $y$ in \cite[Theorem 3.1]{Fix-11} depends on the assignment of $x_\c$ only: it is given by $y(x_\c) = 1 - \prod_{j\in\c}x_j$. Since all hyperedges $\h\in\H$ contain $\c$, the reduced expression~\cite[equation (2)]{Fix-11} equals
\begin{align}
\sum_{\h\in\H} \alpha_\h \Big( y(x_\c)\prod_{j\in\c}x_j +(1-y(x_\c))\prod_{j\in\h\backslash \c} x_j \Big),
\end{align}
where $\alpha_\h > 0$, \ie, it has the form required by \GroupReduce. The actual simplification of the problem is achieved by applying \Reparametrize and \RemoveZero similarly to HOCR.
\par
It follows that reduction~\cite{Fix-11} can be implemented using operations \CliqueReduce, \GroupReduce, \Reparametrize and \RemoveZero.
\end{proof}
\subsection{Bisubmodular Relaxations}\label{sec:bisubmodular-proofs}
The section is organized as follows. We review definitions and the persistency property of bisubmodular functions. Then we define the sum of bisubmodular functions relaxation via injection of label space $\{0,1\}$ into $\{0,\frac{1}{2},1\}$ (with operation \AddLabels). The comparison result is then achieved in two steps. 
We show that persistency for a bisubmodular function is a BLP-improving mapping on 3 labels ($\{0,\frac{1}{2},1\}$) admissible for \AddLabels and then apply the result that \AddLabels has inclusion of BLP-improving maps to transfer all persistencies by bisubmodular relaxations to the BLP-relaxation of the initial problem.
%
\par
We follow the notation of~\cite{Kolmogorov12-bisub}. A sum of bisubmodular functions (SoB) relaxation is constructed as follows. Let $\X_s = \{0,1\}$ and 
\begin{equation}
 \K^{1/2} = \{0,\frac{1}{2},1\}^\V.
\end{equation}
\par
Define binary operations $\sqcap$, $\sqcup\colon \K^{1/2} \to \K^{1/2}$ component-wise according to:
\begin{equation}
\begin{aligned}\label{sqcup-def}
\arraycolsep=6pt\def\arraystretch{1.4}
\begin{array}{c|ccc}
\sqcap & 0 & \frac{1}{2} & 1\\
\hline
0 & 0 & \frac{1}{2} & \frac{1}{2} \\
\frac{1}{2} & \frac{1}{2} & \frac{1}{2} & \frac{1}{2}\\
1 & \frac{1}{2} & \frac{1}{2} & 1
\end{array}
 \ \ \ \ \ \ & 
\arraycolsep=6pt\def\arraystretch{1.4}
\begin{array}{c|ccc}
\sqcup & 0 & \frac{1}{2} & 1\\
\hline
0 & 0 & 0 & \frac{1}{2} \\
\frac{1}{2} & 0 & \frac{1}{2} & 1\\
1 & \frac{1}{2} & 1 & 1
\end{array}\\
\end{aligned}
\end{equation}

\begin{definition}
Function $f \colon \K^{1/2}\to \Real$ is called {\em bisubmodular} if
\begin{equation}\label{def:bisubmodular}
(\forall x,y\in\K^{1/2})\ \ f(x \sqcap y) + f(x \sqcup y) \leq f(x) + f(y).
\end{equation}
\end{definition}

\begin{definition}
Function $f'\colon \K^{1/2}\to \Real$ is a {\em sum of bisubmodular functions relaxation} (SoB relaxation) for $f \colon \X \to \Real$ if for every $\c\in \E$ there holds:
\begin{itemize}
\item $f_\c' \colon \K^{1/2}_\c \to \Real$ is bisubmodular;
\item $(\forall e\in\E,\ \forall x_\c \in\X_\c)\ \ f'_\c(x) = f_\c(x)$;
\end{itemize}
\end{definition}
The next lemma reviews persistency according to \cite[Proposition 12]{Kolmogorov12-bisub}.
\begin{lemma}\label{bisubmodular-maps-lemma}
Let $x^*\in \K^{1/2}$ be a minimizer of $f'$. Define the following mappings 
\begin{subequations}
\begin{align}
& p'_s \colon \K^{1/2}_s \to \K^{1/2}_s \colon x_s \mapsto (x_s \sqcup x^*_s) \sqcup x^*_s;\\
& p_s\colon \X_s \to \X_s \colon
x_s \mapsto \begin{cases}
x^*_s, & \IF x^*_s \neq \frac{1}{2};\\
x_s, & \OTHERWISE.
\end{cases}
\end{align}
\end{subequations}
There holds
\begin{itemize}
\item[(a)]
\label{bisub-autarky}
Autarky: 
$
(\forall x\in \K^{1/2})\ \ f'(p'(x)) \leq f'(x);
$
\item[(b)] $p'(\X) \subset \X$ and $p$ is the restriction of $p'$ to $\X$.
\end{itemize}
\end{lemma}
\begin{proof}
Part (a) follows from bisubmodularity~\eqref{def:bisubmodular} using that $x^*$ is a minimizer: for all $y\in\K^{1/2}$ there holds
\begin{equation}\label{improving-bisub-1}
f'(y \sqcup x^*) \leq \big( f'(x^*)-f'(y \sqcap x^*) \big) + f'(y) \leq f'(y).
\end{equation}
It follows that $(\forall x\in\K^{1/2})$
\begin{equation}\label{improving-bisub-2}
f'((x \sqcup x^*) \sqcup x^*) \leq f'(x \sqcup x^*) \leq f'(x).
\end{equation}
\par
Part (b) can be verified explicitly by calculating the tables for mappings $p$ and $p'$. They are represented respectively as: 
\begin{equation}
\begin{array}{cc}
p_s(x_s) & p'_s(x_s) \\[5pt]
\arraycolsep=6pt\def\arraystretch{1.4}
\begin{array}[t]{c|ccc}
x_s \backslash x^*_s & 0 & \frac{1}{2} & 1\\
\hline
0 & 0 & 0 & 1 \\
1 & 0 & 1 & 1
\end{array}\ \ \ \ \
&\arraycolsep=6pt\def\arraystretch{1.4}
\begin{array}[t]{c|ccc}
x_s \backslash x_s^* & 0 & \frac{1}{2} & 1\\
\hline
0 & 0 & 0 & 1 \\
\frac{1}{2} & 0 & \frac{1}{2} & 1\\
1 & 0 & 1 & 1
\end{array}
\end{array}
\end{equation}
It is seen that for non-fractional $x_s$ the results match.
\qqed 
\end{proof}
\par
%
We now upgrade the autarky property by \Lemma{bisubmodular-maps-lemma}(a) to the statement that $p'$ is BLP-improvinf for $f'$. In order to show this, we construct a reparametrization in which the improving inequality holds component-wise. This reparametrization is given by an optimal dual point, provided that it preserves bisubmodularity of all components. The next Lemma shows such a dual solution exists.
\par
\begin{lemma}\label{BLP for $f'$ modular} Let $f\colon \K^{1/2} \to \Real$ be a sum of bisubmodular functions. BLP for $f$ admits an optimal dual solution $\varphi$ such that each component 
\begin{equation}
f^\varphi_\c (x_\c) = f_\c (x_\c) - (A\T \varphi)_\c(x_\c)
\end{equation}
is bisubmodular.
\end{lemma}
\begin{proof}
The plan of the proof is as follows:
\begin{itemize}
\item Reformulate bisubmodular function $f\colon \K^{1/2} \to \Real$ as a function $g\colon \Bool^{2 \V} \to \Real$.
\item Show there exist a symmetric optimal dual solution to this BLP. Such symmetric solution defines a reparametrization which preserves the sum of bisubmodular functions property for $g$ and consequently for $f$.
\end{itemize}
The following reformulation of bisubmodular function $f\colon \K^{1/2} \to \Real$ as a function $g\colon \Bool^{2 \V} \to \Real$ is according to~\cite{Kolmogorov12-bisub}. Node $s$ is represented by a pair of nodes $(s,s')$. Label $x_s \in \K^{1/2}$ is represented by a pair of labels $(u_s,u_s')$ as follows:
\begin{align}
0 \rightarrow (1,0);\ \ 1 \rightarrow (0,1); \ \ \frac{1}{2} \rightarrow (0,0).
\end{align}
To a hyperedge $\c$ there corresponds hyperedge $\c \cup \c'$, where $\c' = \{s' \mid s\in\c \}$. 
In the hypergraph $(\V \cup \V', \G)$ there are edges of the form $\c \cup \c'$, $\{s\}$ and $\{s'\}$ for $\c \in\E$ and $s\in\V$. We will denote components $g_{\c \cup \c'}$ simply by $g_\c$ for $|\c|\geq 1$.
The energy $g$ is defined by
\begin{equation}\label{busubmodular-representation}
\begin{aligned}
(\forall u\in\Bool^{\c \times \c'} )\ \ g_\c(u_\c,u_{\c'}) = f_\c\left(\frac{u_\c+\bar u_{\c'}}{2}\right),
\end{aligned}
\end{equation}
where $\bar u := 1 - u$.
By definition, $g$ has the symmetry property:
\begin{align}
(\forall u \in \Bool^{\c \times \c'})\ \  g_\c(u_\c, u_{\c'}) = g(\bar u_{\c'}, \bar u_\c).
\end{align}
%
Let
$\X^-_s = \{(u_s,u_{s'})\in \Bool^2 \mid (u_s,u_{s'}) \neq (1,1)\}$ and $\X^- = \prod_{s\in\V} \X_s^-$.
For $u,v\in \X^-$ operations $\sqcup$, $\sqcap$ are defined by 
\begin{subequations}
\begin{align}
u \sqcap v & = u \wedge v,\\
u \sqcup v & = {\rm \tt reduce}(u \vee v),
\end{align}
\end{subequations}
where ${\rm \tt reduce}(w)$ is the labeling obtaining from $w$ by changing labels $(w_s ,w_{s'})$ from $(1,1)$ to $(0,0)$ for all $s \in \V$. It can be seen that these definitions are consistent with equations~\eqref{sqcup-def}.
Components $g_\c$ satisfy
\begin{equation}\label{def:bisubmodular-comp}
\begin{aligned}
(\forall u_\c,v_\c\in\X^{-}_\c)\ \ g_\c(u_\c \sqcap v_\c) + g_\c(u_\c \sqcup v_\c)& \\
\leq g_\c(u_\c) + g_\c(v_\c)&.
\end{aligned}
\end{equation}
\par
With this reformulation we proceed as follows. Let $\varphi$ be dual optimal to $g$. According to the hypergraph $(\V \cup \V', \G)$, there are only components $\varphi_{s,\c\cup \c'}(i)$, $\varphi_{s',\c\cup \c'}(i)$ for $\c\in\E$, $|\c|>1$, $s\in\c$, and $i\in\Bool$ and components $\varphi_{\emptyset,\c\cup \c'}$ for $\c\in\E$. Similarly to function $g$, we can denote the index $\c\cup \c'$ as just $\c$.
\par
If $\varphi_{\emptyset,\c} \neq 0$ for some $\c\in\E$, we can apply the transformation $g_{\c} := g_{\c}-\varphi_{\emptyset,\c}$; $g_{\emptyset}:= g_{\emptyset}+\varphi_{\emptyset,\c}$. Clearly this constant transformation does not change bisubmodularity of $g_{\c}$. Without loss of generality let us assume now that $\varphi_{\emptyset,\c} = 0$ for all $\c \in\E$.
%

Dual feasibility of BLP relaxation for each component $\c \cup \c'$ where $\c\in\E$, $|\c|>1$ reads $(\forall u_{\c \cup \c'} )$
\begin{equation}\label{SoB-g-dual-feas}
g_\c(u_\c,u_{\c'}) -\sum_{s\in\c}\big( \varphi_{s,\c}(u_s) + \varphi_{s',\c}(u_{s'})\big) \geq 0.
\end{equation}
And for unary terms (\ie, $\c = \{s\}$), it is $(\forall u_{\{s,s'\}})$
\begin{equation}
g_s(u_s,u_{s'}) + \sum_{\d \Esupset \{s\} }\big( \varphi_{s,\d}(u_{s'}) +\varphi_{s',\d}(u_s) \big) \geq 0.
\end{equation}

By symmetry of $g$, condition~\eqref{SoB-g-dual-feas} is equivalent to $(\forall u_{\c \cup \c'} )$
\begin{equation}
g_\c(\bar u_{\c'},\bar u_{\c}) -\sum_{s\in\c}\big( \varphi_{s,\c}(u_s) +\varphi_{s',\c}(u_{s'})\big) \geq 0
\end{equation}
and, 
by flipping all bound variables $u_{\c \cup \c'}$, to $(\forall u_{\c \cup \c'} )$
\begin{equation}
g_\c(u_{\c},u_{\c'}) -\sum_{s\in\c}\big( \varphi_{s,\c}(\bar u_{s'}) + \varphi_{s',\c}(\bar u_s)\big) \geq 0.
\end{equation}
Similarly, for unary terms 
there holds $(\forall u_{\{s,s'\}})$
\begin{equation}
g_s(u_s,u_{s'}) + \sum_{\d \Esupset \{s\} }\big( \varphi_{s,\d}(\bar u_{s'}) +\varphi_{s',\d}(\bar u_s) \big) \geq 0.
\end{equation}
Therefore the dual point $\varphi'$ with the following components is feasible:
\begin{equation}\label{varphi-prime}
\begin{aligned}
\varphi' _{s,\c}(i) := \varphi_{s',\c}(\bar i),\\
\varphi' _{s',\c}(i) := \varphi_{s,\c}(\bar i)\\
\end{aligned}
\end{equation}
for $\c\in\E$, $|\c|>1$, $s\in\c$, $i\in\Bool$ (components $\varphi'_{\emptyset,\c} = 0$ for $\c\in\E$ are omitted in this and subsequent steps).
Solution $\varphi'$ is clearly optimal to BLP of $g$ since 
$$
g^\varphi_\emptyset  = g_\emptyset =  g^{\varphi'}_\emptyset.
$$
Therefore the following symmetrized solution is optimal:
\begin{equation}\label{varphi symmetrized}
\begin{aligned}
\hat \varphi_{s,\c}(u_s) := & \frac{1}{2}\Big(\varphi_{s,\c}(u_s) + \varphi_{s',\c}(\bar u_s)\Big),\\
\hat \varphi_{s',\c}(u_{s'}) := & \frac{1}{2}\Big(\varphi_{s',\c}(u_{s'}) + \varphi_{s,\c}(\bar u_{s'})\Big).
\end{aligned}
\end{equation}
Let us check it is bisubmodular. We map this dual solution back to BLP for $f$ by reversing~\eqref{busubmodular-representation} as
\begin{align}
\notag
& \tilde \varphi_{s,\c}(0) := \hat \varphi_{s,\c}(1)+\hat \varphi_{s',\c}(0) = \varphi_{s,\c}(0) + \varphi_{s',\c}(1),\\
\notag
& \tilde \varphi_{s,\c}(1) := \hat \varphi_{s,\c}(0)+\hat \varphi_{s',\c}(1) = \varphi_{s,\c}(1) + \varphi_{s',\c}(0),\\
\label{BLP-construction-hat-varphi}
& \tilde \varphi_{s,\c}(\frac{1}{2})  := \hat \varphi_{s,\c}(0)+\hat \varphi_{s',\c}(0) \\
\notag
& = \frac{1}{2}\Big(\varphi_{s,\c}(0)+\varphi_{s,\c}(1) \Big) + \frac{1}{2}\Big(\varphi_{s',\c}(0)+\varphi_{s',\c}(1) \Big)
\end{align}
It satisfies $\tilde \varphi_{s,\c}(\frac{1}{2}) = \frac{1}{2}(\tilde \varphi_{s,\c}(0)+\tilde \varphi_{s,\c}(1))$ and therefore both $\tilde \varphi_{s,\c}$ and $-\tilde \varphi_{s,\c}$ are bisubmodular. 
\par
It remains to show that $\tilde \varphi$ is optimal to BLP of $f'$. 
%
While it is well known that BLP relaxation is tight for SoB function $f'\colon \K^{1/2} \to \Real$, \eg~\cite{Thapper-12}, it is not obvious that BLP relaxation for the reformulation $g \colon \Bool^{2\V}\to \Real$ is tight as well. Let us show this is the case. It will follow then that $\tilde \varphi$, constructed from an optimal dual solution $\hat \varphi$ to BLP of $g$, is optimal to BLP of $f'$.

\begin{statement}\label{BLP-SoB-tight} The BLP relaxation for sum of bisubmodular functions $g \colon \Bool^{2\V}\to \Real$ is tight.
\end{statement}
\begin{proof} The schema of the proof is similar to \eg \cite[T.6.2]{Cooper-08} or \cite{Schlesinger00} who considered sum of submodular functions.
We construct a primal integer optimal solution from an arc-consistent optimal dual solution. In the construction we will need that the reparametrized problem $g^\varphi$ is a sum of bisubmodular functions. Therefore we need an arc consistent symmetric optimal dual solution.
\par
We start by taking a pair $(\mu,\varphi)$ that satisfies strict complementarity slackness for BLP of $g$. Since arc consistency is a necessary condition for strict complementarity, $\varphi$ is arc consistent. As was shown before, $\varphi'$ defined by equations~\eqref{varphi-prime} is dual optimal to BLP of $g$. It is arc consistent because $\varphi$ was arc consistent. By taking a symmetrized dual solution $\hat \varphi$ defined by equations~\eqref{varphi symmetrized} we obtain an optimal symmetric arc consistent dual solution. By symmetry, it preserves component-wise bisubmodularity.
\par
Let now $\varphi := \hat \varphi$. We construct an integer solution as:
\begin{equation}
u^*_s = \Wedge \{ i\in\Bool \mid g^{\varphi}_s(i) = 0 \}. 
\end{equation}
In order to show that $u^*$ is optimal we prove complementarity with $\varphi$. Let $\c \in \E$. By arc consistency, for every $s\in \c$ there exists $u_\c^{(s)}$ such that $u_s^{(s)} = u^*_s$ and
$g^\varphi_{\c}(u^{(s)}_\c) = 0$. It also follows that $(\forall t\in\c \mid t\neq s)$ $g^\varphi_{t}(u^{(s)}_t) = 0$ and hence $u^{(s)}_t \geq u^*_t$. It follows that $u^*_\c = \Wedge_{s\in \c}u^{(s)}_\c$ and by component-wise bisubmodularity we have that
$$
g^\varphi_\c(\Wedge_{s\in \c}u^{(s)}_\c) = 0.
$$
From feasibility and complementarity slackness follows $u^*$ is optimal. Therefore BLP relaxation for $g$ is tight.
\qqed 
\end{proof}
It follows that $\tilde \varphi$ constructed in~\eqref{BLP-construction-hat-varphi} not only feasible to BLP of $f'$ but achieves the optimal dual objective $f^{\tilde \varphi}_{\emptyset} = g^{\varphi}_{\emptyset} = \min_{u\in\Bool^{2\V}}\energy{g}(u)$.
\end{proof}

\begin{lemma}\label{SoS-BLP-improving}
Mapping $p'$ defined in \autoref{bisubmodular-maps-lemma} by an optimal solution $x^*\in\K^{1/2}$ is BLP-improving for $f'$.
\end{lemma}
\begin{proof}
Let $\varphi$ provide a component-wise bisubmodular optimal reparametrization $f'^{\varphi}$ for BLP of $f'$ which exists by \Lemma{BLP for $f'$ modular}. Since BLP for $f'$ is tight, $\varphi$ and $\delta(x^*)$ must satisfy complementary slackness.
For every hyperedge $\c\in\E$, $\c\neq \emptyset$ by dual feasibility $f'^{\varphi}_\c \geq 0$ and by complementary slackness 
\begin{equation}
f'^{\varphi}_\c(x^*_\c) = 0.
\end{equation}
Therefore, labeling $x^*_\c$ is a minimizer of $f'^{\varphi}_\c$. 
Since $f'^{\varphi}_\c$ is bisubmodular it follows by the same argument as in equations~\eqref{improving-bisub-1}, \eqref{improving-bisub-2} that for each hyperedge $\c$ we have the component-wise autarky property:
\begin{equation}
(\forall x_\c\in \K^{1/2}_\c)\ \ f_\c'^{\varphi}(p'_\c(x_\c)) \leq f_\c'^{\varphi}(x_\c);\\
\end{equation}
By the characterization \Theorem{T:characterization}(c), mapping $p'$ is BLP-improving for $f'$.
\qqed 
\end{proof}
\TbisubmodularBLP*
\begin{proof}
Let $p'$ and $p$ be mappings defined in \autoref{bisubmodular-maps-lemma}, corresponding to the persistency~\cite{Kolmogorov10-bisub}.
We need to show that mapping $p$ is BLP-improving for $f$. By \autoref{SoS-BLP-improving}, $p'$ is BLP-improving for $f'$.
The discrete relaxation $f'$ is obtained from $f$ with \AddLabels which has inclusion of admisible BLP-improving maps (\autoref{T:quad-weaken1}). By \autoref{bisubmodular-maps-lemma}(b), $p'$ is admissible and thus \autoref{T:quad-weaken1} applies. 
\qqed 
\end{proof}
%
%
%
%
\subsection{Persistency in 0-1 Polynomial Programming by Adams \etalb}\label{sec:adams-proofs}
%
The $0$-$1$ polynomial programming problem is the following optimization problem:
\begin{equation}\label{PP}
\tag{PP}
\min \sum_{J\subset \V} c_J \prod_{j\in J} x_j,
\end{equation}
where $x$ is binary and coefficients $c_J \in\Real$ are not necessarily all non-zero. The objective of~\eqref{PP} is a multilinear polynomial expression, which can be written uniquely for any polynomial in $0$-$1$ variables applying the identity $x_j^2 = x_j$. \citet{Adams:1998} considered a hierarchy of relaxations of~\citet{SheraliA90} for this problem. A relaxation of level $d$ can be constructed assuming that for all $|J|>d$ coefficients $c_J$ are zero.

To match their result we need the following:
\begin{itemize}
\item A hypergraph $(\V,\E)$, where $\E$ contains all subsets of $\V$ of cardinality up to $d$ (in case $d=2$ it is a fully connected graph):
$$
\E = \{J\subset \V \mid |J|\leq d\}
$$
\item Represent the problem~\eqref{PP} as energy minimization with terms
\begin{align}
f_{J}(x_J) = \begin{cases}
c_J & \IF  x_J = 1_J,\\
0 & \OTHERWISE.
\end{cases}
\end{align}
\item Consider the FLP relaxation.
\end{itemize}
The persistency result~\cite[Lemma 3.2]{Adams:1998} can be described as follows. Their lemma partitions the set of nodes as $\V = N^+ \cup N^- \cup N^f$. A sufficient condition is proposed implying that the partial assignment $x_{N^-} = 0$, $x_{N^+} = 1$ is globally optimal. 
\par
We interpret their result in the dual decomposition framework. The hypergraph $(\V,\E)$ is split into two parts:
\begin{itemize}
\item Nodes $\V_1 = N^f$, hyperedges $\E_1 = \{J\in\E \mid J \subset N^f \}$.
\item Nodes $\V_2 = N^+\cup N^- \cup \B$, hyperedges $\E_2 = \{J\in\E \mid J \not\subset N^f \}$, where
$\B$ is the following {\em boundary} set:
\begin{equation}
\B = \{v \in N^f \mid (\exists J\in\E)\ v\in J,\  J \not\subset N^f\}.
\end{equation} 
\end{itemize}
It can be seen that hyperedges $\E_1$, $\E_2$ form a partition of $\E$ whereas the sets of nodes $\V_1$, $\V_2$ overlap over $\B$. Accordingly to the two hypergraphs we define the decomposition:
\begin{equation}
f(x) = f^1(x_{\V_1}) + f^2(x_{\V_2}),
\end{equation}
where
\begin{subequations}
\begin{align}
&f^1\colon \V_1\to \Real \colon x\mapsto \sum_{J\in\E_1}c_J x_J,\\
&f^2\colon \V_2\to \Real \colon x\mapsto \sum_{J\in\E_2}c_J x_J.
\end{align}
\end{subequations}
If we found an optimal solution to $f^1$ and an optimal solution to $f^2$ and accidentally they were consistent over the overlap part $\B$ we would have obtained an optimal solution to $f$.
\par
We will show that the conditions of \citet{Adams:1998} imply that an arbitrary solution $x_{\V_1}$ to $f^1$ defines an optimal solution $x'$ to $f^2$ given by
\begin{equation}
x'_s = \begin{cases}
x_s, & s\in\B\\
0, & s\in N^-\\
1, & s\in N^+.
\end{cases}
\end{equation}
In other words, any extension of $x_{\V_1}$ to $\V_2$ defines an optimal solution to $f^2$.
In fact, under these conditions, what happens inside the problem $f^1$ is irrelevant.
\par
Recall that the relaxation~\cite{Adams:1998} is obtained by introducing a relaxed variable $w_J\in [0,\,1]$ in place of every product $\prod_{s\in J} x_J$. The constraints are represented by the following linear forms:
\begin{equation}
f_d(J_1,J_2) = \sum_{J'\subset J_2} (-1)^{|J'|}w_{J_1 \cup J'}
\end{equation}
for each $J_1\cap J_2 = \emptyset$ and $|J_1\cup J_2| = d$. The relaxation, denoted LP$(d/n)$ is given by
\begin{align}
& \min \sum_{J\subset \V} c_J w_J\\
\notag
& (\forall J_1,J_2\ |J_1\cup J_2|=d, J_1 \cap J_2 = \emptyset )\ f_d(J_1,J_2) \geq 0.
\end{align} 
This relaxation can be matched to FLP by the relation
\begin{equation}
\mu_{J}(x_J) = f_d(J_1,J_2),
\end{equation}
where $J_1\cap J_2 = \emptyset$, $J_1 \cup J_2 = J$, $d = |J|$, $x_{J_1} = 1$, $x_{J_2} = 0$. In particular,
\begin{equation}
\mu_{J}(1_J) = w_J,
\end{equation}
where $1_J$ is a $|J|$-vector of ones. For $w$ feasible to LP(d/n), the terms $f_d(J_1,J_2)$ are non-negative and satisfy marginalization constraints according to~\cite[equation 2.3]{Adams:1998}. It follows from these marginalization constraints that a feasible solution satisfies further inequality constraints:
$$
(\forall p\leq d,\forall (J_1,J_2)_d)\ f_p(J_1,J_2) \geq 0.
$$
This ensures that the corresponding solution $\mu$ is feasible to FLP and hence the relaxations are identical.
The persistency lemma formulated in \cite[Lemma 3.2]{Adams:1998} has the following form. 
\begin{lemma}[Lemma 3.2 of \citet{Adams:1998}]\label{L:Adams}
Let $N^+\cup N^- \cup N^f$ be a partition of $\V$. Let there exist dual multipliers satisfying certain sufficient condition (as defined in \cite{Adams:1998}). Then there exists an optimal solution to LP($d/n$) having $w_s = 1$ for $s \in N^+$ and $w_s = 0$ for $s \in N^-$.
\end{lemma}
For an interested reader we remark that the sufficient conditions of their lemma ensure a dual feasible solution for $f^2$ which is complementary to the solution defined by $N^+$, $N^-$ on $N^+\cup N^-$ and is zero on all boundary constraints $J\subset \B$. This ensures complementarity with any feasible primal solution consistent with $N^+$, $N^-$. We are not going to prove this claim formally, but use the existing lemma together with the observation that their sufficient condition does not depend on the coefficients $\{c_J \mid J\subset N^f\}$.
%
\begin{lemma}\label{L:Adams+}
Assume that the conditions of \Lemma{L:Adams} are satisfied. 
Let $w_s = 1$ for $s \in N^+$, $w_s = 0$ for $s \in N^-$ and let $w$ be feasible to LP($d/n$). 
Then $w$ is optimal to LP($d/|\V_2|$) for $f^2$.
\end{lemma}
\begin{proof}
It can be seen from \cite[equation 3.4d]{Adams:1998} that if the conditions of their Lemma are satisfied then they are also satisfied with coefficients $c_J = 0$ for all $J\subset N^f$.
Since $w$ is feasible, it is also feasible to their equation 3.5[d]. The objective of the latter is identically zero, therefore $w$ is optimal to their equation 3.5[d]. \Lemma{L:Adams} proves that $w$ is optimal to LP($d/n$). Since we made $f^1$ zero, $w$ is optimal to LP($d/|\V_2|$) for $f^2$. \qqed 
\end{proof}
\par
With this refined result we can easily prove the relaxed-improving property. Define the mapping
\begin{equation}
p_s(x_s) = \begin{cases}
0, & s\in N^-,\\
1, & s\in N^+,\\
x_s, & \OTHERWISE. 
\end{cases}
\end{equation}
\TAdamscondition*
\begin{proof}
We need to show that mapping $p$ is FLP-improving. Let $\mu\in\Lambda$. Then $\mu' := [p]\mu \in \Lambda$ is optimal to $f^2$ by~\Lemma{L:Adams+}. At the same time $(\forall J\in\E, J\cap N^f \neq \emptyset)$ we have $\mu'_J = \mu_J$ and hence $\<f^1,\mu_J\> = \<f^1,\mu_J'\>$. It follows that
\begin{equation}
\<f,[p]\mu\> = \<f^1,[p]\mu\>+\<f^2,[p]\mu\> \leq \<f,\mu\>.
\end{equation}
\qqed 
\end{proof}
We obtained that FLP maximum persistency dominates the persistency result of \citet{Adams:1998}. In case of strong persistency, the former is found by \Algorithm{Alg1}.
\par
In the pairwise case, conditions~\cite{Adams:1998} are always satisfied for the same persistency assignment as the roof dual. However, in the higher order case our conditions are less restrictive as can be seen from the dual representation~\eqref{WI-dual}: we require less inequalities than the complementarity slackness imposed on all solutions of the unassigned part.
\section*{Acknowledgment}
I would like to thank the anonymous reviewers and my colleagues Thomas Pock and Paul Swoboda for their comments which helped to improve clarity and correctness. \mythanks
\def\bib{.}
\section*{References}
\bibliographystyle{apa}
\bibliography{\bib/suppl,\bib/maxflow,\bib/max-plus-en,\bib/kiev-en,\bib/pseudo-Bool,\bib/shekhovt,\bib/vcsp,\bib/dee,\bib/books}
\end{document}